\documentclass[accepted]{uai2026} 
                        
\usepackage[american]{babel}

\usepackage{natbib} 
\usepackage{mathtools} 
\usepackage{booktabs} 
\usepackage{tikz} 

\usepackage[T1]{fontenc}
\usepackage[utf8]{inputenc}
\usepackage{graphicx}
\usepackage{amsmath}
\usepackage{amssymb}
\usepackage{amsthm}
\usepackage{bm}
\usepackage{subcaption}
\usepackage{hyperref}
\usepackage{xcolor}
\usepackage{array}
\usepackage{makecell}
\usepackage{longtable}
\usepackage{mathtools}
\usepackage{enumitem}
\usepackage{tcolorbox}
\usepackage{booktabs}
\usepackage{wrapfig}
\usepackage{multirow}
\usepackage{placeins}
\usepackage{siunitx}

\newcommand{\bE}{\mathbb{E}}
\newcommand{\Ree}{\mathbb{R}}
\newcommand{\bV}{\mathbb{V}}
\newcommand{\given}{\ |\ }
\newcommand{\vtheta}{\bm{\theta}}
\newcommand{\vvtheta}{\bm{\vartheta}}

\newcommand{\tu}{\operatorname{TU}}
\newcommand{\au}{\operatorname{AU}}
\newcommand{\eu}{\operatorname{EU}}
\newcommand{\bP}{\mathbb{P}}
\newcommand{\bQ}{\mathbb{Q}}
\newcommand{\cP}{\mathcal{P}}

\newcommand{\cY}{\mathcal{Y}}
\newcommand{\cX}{\mathcal{X}}
\newcommand{\tr}{\mathrm{tr}}
\newcommand{\bx}{\boldsymbol{x}}

\newcommand{\by}{\boldsymbol{y}}

\newtheorem{definition}{Definition}[section]

\newtheorem{proposition}{Proposition}[section]

\usepackage{pifont}
\DeclareUnicodeCharacter{03C7}{$\chi$}
\newrobustcmd\B{\DeclareFontSeriesDefault[rm]{bf}{b}\bfseries}

\title{Uncertainty Quantification for Regression:\\ A Unified Framework based on Kernel Scores}

\author[1,2]{Christopher Bülte}
\author[1,2]{Yusuf Sale}
\author[1,2,3,4]{Gitta Kutyniok}
\author[1,2,5]{Eyke Hüllermeier}

\affil[1]{
LMU Munich
}
\affil[2]{%
Munich Center for Machine Learning (MCML)
}
\affil[3]{%
DLR-German Aerospace Center
}
\affil[4]{%
University of Troms\o
}
\affil[5]{%
German Research Center for Artificial Intelligence (DFKI, DSA) 
}

\begin{document}
\maketitle

\begin{abstract}
Regression tasks, notably in safety-critical domains, require reliable uncertainty quantification, yet the literature remains largely classification-focused. To address this, we introduce a family of measures for total, aleatoric, and epistemic uncertainty in multivariate regression based on strictly proper kernel scores. The framework provides a principled recipe for designing new uncertainty measures whose behavior, such as tail sensitivity or out-of-distribution responsiveness, is governed by the choice of the underlying kernel, while also encompassing existing measures under a joint analysis.
We prove explicit correspondences between properties of the kernel and behavior of resulting uncertainty measures, yielding concrete design guidelines for practitioners. Extensive experiments across structured regression tasks, including spatial and functional domains, demonstrate effectiveness on downstream tasks such as out-of-distribution detection and active learning, and reveal that different kernel choices lead to distinct trade-offs, offering practitioners guidance for task-specific selection.
\end{abstract}

\section{Introduction} 
Predictive models now drive decision-making in safety-critical domains such as weather forecasting \citep{priceProbabilisticWeatherForecasting2025, alet2025skillfuljointprobabilisticweather}, autonomous driving \citep{michelmore2018evaluatinguncertaintyquantificationendtoend} or healthcare \citep{lohr2024towards,uq_mri}; tasks where careful analysis of the model predictions and accurate uncertainty quantification are indispensable. Many studies have analyzed different approaches to quantify predictive uncertainty, often distinguishing between different sources of uncertainty. In particular, one usually considers two sources of uncertainty: \emph{aleatoric uncertainty} and \emph{epistemic uncertainty} \citep{hullermeier2021aleatoric}. Broadly speaking, aleatoric uncertainty describes the inherent randomness in the data-generating process, for example, due to measurement errors and, as it describes variability that is independent of the amount of data, is often referred to as \emph{irreducible} uncertainty. Epistemic uncertainty, on the other hand, arises from a lack of knowledge about the data-generating process and can be reduced by improving the model or acquiring more data; therefore, it is also referred to as \emph{reducible} uncertainty.

While aleatoric uncertainty is well captured in predictive models, epistemic uncertainty is more difficult to represent and requires higher-order formalisms, such as second-order distributions (distributions of distributions), which is referred to as \emph{uncertainty representation} \citep{hullermeier2021aleatoric}.
Given such a representation, the key question is how to measure or quantify the total, aleatoric, and epistemic uncertainty (\emph{uncertainty quantification}). While the representation mainly determines predictive performance, the choice of uncertainty measure plays a vital role in decision making and can have an additional impact on the performance of downstream tasks, with numerous works developing and analyzing new measures for uncertainty quantification \citep{sale2023secondorderuncertaintyquantificationdistancebased, malinin2021uncertaintyestimationautoregressivestructured, kotelevskii2022nonparametric, berry2024efficientepistemicuncertaintyestimation}. In addition, recent work focuses on steps towards more unified approaches that incorporate many existing measures and give guidance on how to construct new ones \citep{hofman2024quantifyingaleatoricepistemicuncertainty, kotelevskii2025from}. However, research has focused either on uncertainty quantification in classification or on parametric univariate regression tasks, neglecting the increasing amount of structured domains where generative models and other nonparametric methods show great performance \citep{alet2025skillfuljointprobabilisticweather, ke2023repurposing}.

In \emph{regression} tasks, a practitioner is generally interested in predictive uncertainty, which describes the uncertainty of the target $\bm y \in \cY$ given some covariates $\bm x \in \cX$. While the notions of total, aleatoric, and epistemic uncertainty remain the same \citep{hullermeier2021aleatoric}, the corresponding uncertainty measures fundamentally differ from the classification case. Unlike classification, where the label space is discrete and bounded, regression targets lie in an (often) unbounded, continuous, and possibly high-dimensional domain, which renders existing measures unsuitable.
While many regression methods focus on uncertainty \emph{representation} \citep{aminiDeepEvidentialRegression2020, lakshminarayananSimpleScalablePredictive2017, kelen2025distributionfree}, only a few works study the underlying uncertainty measures from a theoretical standpoint \citep{berry2024efficientepistemicuncertaintyestimation, buelte2025axiomaticassessmententropyvariancebased}.
Score-divergence-based decompositions of uncertainty have recently been formalized for \emph{classification} \citep{kotelevskii2025from, hofman2024quantifyingaleatoricepistemicuncertainty}, yet no analogous, theoretically grounded measures exist for the multivariate regression setting. 
%
\begin{figure}[t]
    \centering
    \includegraphics[width=\linewidth]{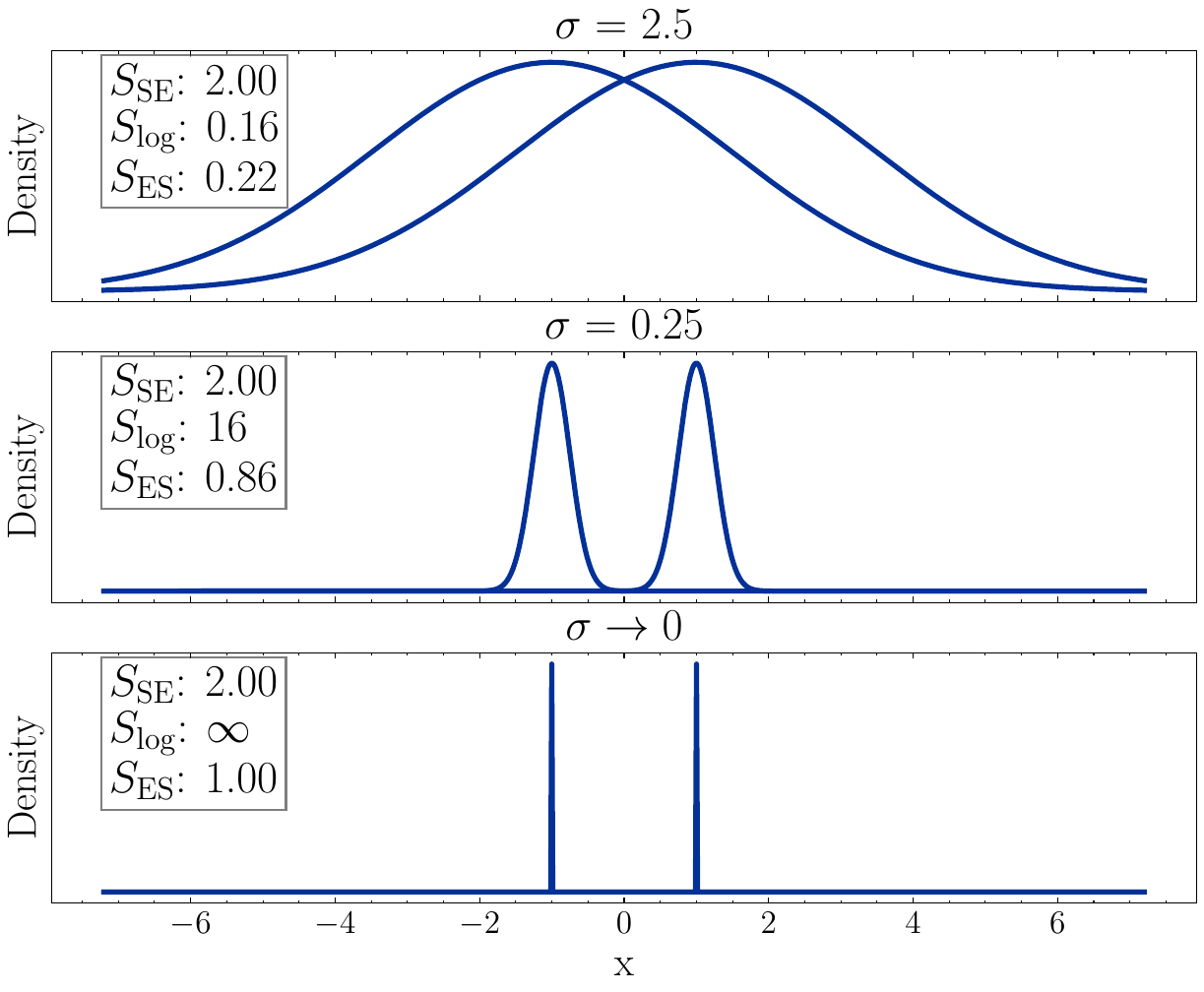}
    \caption{Illustration of epistemic uncertainty for a two-member Gaussian ensemble with shared variance. As $\sigma$ shrinks, the variance-based measure ($S_\mathrm{SE}$) stays constant, the entropy-based measure ($S_\mathrm{log}$) diverges, while our proposed energy-score-based measure ($S_\mathrm{ES}$) converges to half the Euclidean distance between component means.   
    }
    \label{fig:ensemble_eu}
\end{figure}
\paragraph{Contributions}
We address this gap in two steps.
\emph{(i)} We transfer the score-divergence formulation of total, aleatoric, and epistemic uncertainty \citep{kotelevskii2025from, hofman2024quantifyingaleatoricepistemicuncertainty} from classification to the multivariate regression setting, yielding a well-defined framework for designing uncertainty measures in continuous target spaces.
\emph{(ii)} Within this setting, we identify strictly proper \emph{kernel scores} \citep{Gneiting.2007} as a particularly well-suited family for instantiating these measures: they carry a metric structure, and come with an unbiased, sample-based estimator, which keeps them applicable to complex predictive distributions where density-based measures break down.
The choice of kernel then acts as a design lever: we prove explicit connections between properties of the kernel and desirable behavior of the associated uncertainty measure, with regards to the assessment of uncertainties, translation invariance, and robustness.
These properties target concrete failure modes of existing measures, as illustrated in \autoref{fig:ensemble_eu}.
Finally, we validate the proposed measures empirically, demonstrating the derived theoretical properties in practice and showcasing their strong performance across a wide range of complex structured regression tasks, including depth estimation and the prediction of dynamical systems.

\section{Uncertainty in regression}
In the following, we denote by $\mathcal{X} \subseteq \mathbb{R}^k$ and $\mathcal{Y} \subseteq \mathbb{R}^d$ the (real-valued) feature and target space, respectively. Furthermore, let $\mathcal{P}(\mathcal{Y})$ denote a convex set of probability measures on the measure space $(\mathcal{Y}, \sigma(\mathcal{Y}))$, where $\sigma(\mathcal{Y})$ is a suitable $\sigma$-algebra, and let $\overline{\Ree} = \Ree \cup \{- \infty, \infty \}$. In addition, we write $\mathcal{D} = \{\boldsymbol{x}_i, \boldsymbol{y}_i  \}_{i=1}^n \in (\mathcal{X} \times \mathcal{Y})^n$ for the training data. For $i \in \{1, \dots, n\}$, each pair $(\bm{x}_i, \bm y_i)$ is a realization of the random variables $(X_i, Y_i)$, which are assumed to be independent and identically distributed (i.i.d) according to some probability measure $\bP$. Therefore, each feature vector $\bm{x} \in \mathcal{X}$ induces a conditional probability distribution $\bP(\cdot \given \bm{x}) \in \mathcal{P}(\cY)$ over the outcome space $\cY$.

\paragraph{Uncertainty representation}
Regarding second-order uncertainty quantification, we similarly define by $\mathcal{P}( \mathcal{P}(\mathcal{Y}))$ the set of all probability measures on $(\mathcal{P}(\mathcal{Y}), \sigma(\mathcal{P}(\mathcal{Y})))$, with a suitable $\sigma$-algebra. We refer to $Q \in \mathcal{P}(\mathcal{P}(\mathcal{Y}))$ as a \emph{second-order distribution}. In contrast to the classification setting, the probability measures $\bP \in \cP(\mathcal{Y})$ are not necessarily defined on a bounded domain. While we keep the setup as general as possible and this article mainly revolves around uncertainty quantification rather than uncertainty representation, the following examples illustrate how a second-order distribution could be specified within our framework:

\emph{Parametric distributions:}
Given absolute continuity with respect to the Lebesgue measure and a (fixed) parametric distribution $p(\cdot \mid \vtheta(\boldsymbol{x}) )$ with $\vtheta \in \Theta \subseteq \mathbb{R}^p$, we can consider the second-order distribution to be on the (measurable) parameter space $(\Theta, \sigma(\Theta))$, e.g. $Q \in \mathcal{P}(\Theta)$. In particular, this includes many uncertainty quantification methods, such as deep ensembles \citep{lakshminarayananSimpleScalablePredictive2017}, deep evidential regression \citep{aminiDeepEvidentialRegression2020}, or distributional regression \citep{kneibRageMeanReview2023}.

\emph{Ensemble approaches: }Given an empirical measure, i.e. $Q = Q_m :=  \frac{1}{M}\sum_{m=1}^M \delta_{\bP_m}$ for first-order distributions $\bP_m \sim Q$, the setting includes ensembles of general first-order methods such as normalizing flows \citep{berryNormalizingFlowEnsembles2023}, mixture density networks \citep{bishop}, nonparametric ensembles \citep{kelen2025distributionfree} or diffusion models \citep{wolleb2021diffusionmodelsimplicitimage}.

Unless noted otherwise, we will consider arbitrary first- and second-order distributions, where we assume that we have a first-order distribution $\bP \sim Q$, distributed to some second-order distribution $Q$ and $Y \sim \bP$. In addition, we define the (first-order) predictive mixture distribution $\overline{\bP} \coloneq \bE_Q[\bP]$, which can be interpreted as the Bayesian model average (BMA) predictive distribution \citep{schweighofer2023introducingimprovedinformationtheoreticmeasure}.

\section{Uncertainty quantification based on proper scoring rules}
\label{sec:scoring_rule_entropies}
In this section, we recall how proper scoring rules can be established to define uncertainty measures.
A \emph{scoring rule} \citep{Gneiting.2007} is a function $S: \mathcal{P}(\cY) \times \mathcal{Y} \to \overline{\Ree}$, such that $S(\bP, \mathbb{Q}) \coloneq \int S(\bP,\bm y)\, d \mathbb{Q}(\bm y)$ is well-defined for all $\mathbb{P,Q} \in \mathcal{P}(\cY)$. $S$ is called \emph{proper}, if
$
    {S(\bQ,\bQ) \leq S(\bP,\bQ)}, \mathrm{for \ all\ } \bP,\bQ \in \mathcal{P}(\cY)
$
and \emph{strictly proper} if equality holds only when $\bP=\mathbb{Q}$. Intuitively, proper scoring rules quantify the discrepancy between a predictive distribution and an observed outcome. 
Following \cite{dawidGeometryProperScoring2007}, every scoring rule $S$ can be associated with a \emph{(generalized) entropy} $H: \mathcal{P}(\cY) \to \overline{\Ree}$ and a \emph{divergence} $D: \mathcal{P}(\cY) \times \mathcal{P}(\cY)  \to \overline{\Ree}$, via
\begin{alignat}{2}
    H:   \bP            & \mapsto H(\bP) \coloneq S(\bP,\bP)              \\
    D:   (\mathbb{P,Q}) & \mapsto D(\mathbb{P,Q}) \coloneq S(\mathbb{P,Q}) - H(\mathbb{Q}).
\end{alignat}
For (strictly) proper scoring rules, $H$ is (strictly) concave on $\mathcal{P}(\cY)$, while the divergence satisfies $D(\mathbb{P,Q}) \geq 0$ for $\mathbb{P,Q}\in \mathcal{P}(\cY)$ with equality if and only if $\bP = \mathbb{Q}$ \cite[compare][]{dawidGeometryProperScoring2007}. These quantities generalize the familiar notions of Shannon entropy and Kullback-Leibler divergence: $H$ captures the average surprisal under a distribution, and $D$ measures the discrepancy between two distributions. Under mild assumptions, proper scoring rules can be characterized in terms of their entropy function \citep{Gneiting.2007}, so either can be used to construct the other.

Scoring rules have been utilized to construct uncertainty measures \citep{kotelevskii2025from, hofman2024quantifyingaleatoricepistemicuncertainty}, which can be adapted to our second-order distribution $Q$ in the following way:
\begin{align}
 \label{eq:scoring_rule_uq_bma}
 \tu_{\mathrm{B}}(Q) &\coloneq \mathbb{E}_{\bP\sim Q}[S(\overline{\bP},\bP)], \nonumber \\ 
\eu_{\mathrm{B}}(Q) &\coloneq \bE_{\bP\sim Q}[D(\overline{\bP},\bP)], \\
\au_{\mathrm{B}}(Q) &\coloneq \bE_{\bP\sim Q}[H(\bP)], \nonumber
\end{align}
Here, epistemic uncertainty (EU) measures the spread of the predictive distributions around their mixture, while aleatoric uncertainty (AU) captures average irreducible noise. Total uncertainty (TU) is the sum thereof.

Alternatively, \citet{kotelevskii2025from, schweighofer2023introducingimprovedinformationtheoreticmeasure, berry2024efficientepistemicuncertaintyestimation} proposes \emph{pairwise estimators} that replace the mixture $\overline{\bP}$ with expectations over independent draws from $Q$: 
\begin{align}
 \label{eq:scoring_rule_uq_p}
\tu_{\mathrm{P}}(Q) &\coloneq \mathbb{E}_{\bP,\bP'\sim Q}[S(\bP',\bP)],  \\
\eu_{\mathrm{P}}(Q) &\coloneq \bE_{\bP,\bP'\sim Q}[D(\bP',\bP)], \nonumber
\end{align}
with AU unchanged. Both variants satisfy the additive decomposition $\tu = \eu + \au$.
The pairwise estimator avoids computing or sampling from the typically intractable mixture distribution, and\textemdash{}crucially\textemdash{}admits closed-form expressions for many parametric families. The BMA estimator is the less expensive alternative ($\mathcal{O}(M)$ vs. $\mathcal{O}(M^2)$ for an ensemble of size $M$), but usually requires approximation of $\overline{\bP}$.
When $S$ is convex in its first argument, Jensen's inequality gives $\tu_P \geq \tu_B$, so that the pairwise estimator provides an upper bound \citep{schweighofer2023introducingimprovedinformationtheoreticmeasure}.

While this decomposition is general, its application to regression has been limited: the log-score, which leads to the familiar entropy-based measure \citep{fishkov2025uncertaintyquantificationregressionusing}, requires absolute continuity and therefore density estimation, which is intractable for high-dimensional data. In the following section, we propose kernel scores as a principled and practically advantageous instantiation of this framework for general regression settings.

\section{Kernel scores}
We now introduce kernel scores as the central tool of our framework. The key insight is that kernel scores inherit all the structural properties required for the decomposition in \eqref{eq:scoring_rule_uq_bma}–\eqref{eq:scoring_rule_uq_p}, while additionally providing closed-form expressions for a broad class of distributions, unbiased nonparametric estimators, and applicability to structured domains such as graphs or functional data. Here, we draw mainly on the notation of \citet{waghmareProperScoringRules2025}.

\begin{definition}[Kernel score]
    Let $k:\mathcal{Y} \times \mathcal{Y} \to \Ree$ be a continuous, conditionally negative definite kernel\footnote{A kernel $k: \mathcal{Y} \times \mathcal{Y} \to  \Ree$ is conditionally negative definite if ${\sum_{i,j=1}^n a_i a_j k(\bx_i, \bx_j) \leq 0},\ {\forall n \in \mathbb{N}},\ {\bx_1 ,\ldots, \bx_n \in \mathcal{Y}}, \ \mathrm{and }\\ {a_1, \ldots, a_n \in \Ree}\ \mathrm{with}\ \sum_{j=1}^n a_j = 0.$ \citep{waghmareProperScoringRules2025}.}, and ${\mathcal{P}_k = \{\bP \in \mathcal{P}(\cY): \iint k(\bm x, \bm x') \, d\bP(\bm x) \, d\bP(\bm x') < \infty  \}}$.
    Then, the associated kernel score is
    \begin{align}
        \label{eq:kernel_score}
        S_k(\bP,\bm y) &= \int k(\bm x,\bm y) \, d\bP(x) \\
        &- \frac{1}{2} \iint k(\bm x,\bm x') \, d\bP(\bm x) \, d\bP(\bm x') - \frac{1}{2}k(\bm y,\bm y), \nonumber
    \end{align}
    for $\bP \in \cP_k, \bm y \in \cY$, with induced entropy and divergence
\begin{align}
    H_k(\mathbb{P})   & = \frac{1}{2} \iint  k(\bm x, \bm x') \, d\bP(\bm x) \, d\bP(\bm x')\\
    &- \frac{1}{2} \int k(\bm x,\bm x) \, d\mathbb{P}(\bm x),   \nonumber        \\
    D_k(\mathbb{P,Q}) & = - \frac{1}{2} \iint  k(\bm y,\bm y') \, d(\bP-\mathbb{Q})(\bm y) \, d(\mathbb{P-Q})(\bm y'),
    \label{eq:kernel_divergence}
\end{align}
for $\bP, \bQ \in \cP_k$.
\end{definition}
$S_k$ is nonnegative and (strictly) proper for a (strongly) conditionally negative definite kernel \citep{waghmareProperScoringRules2025}.

Instantiating the pairwise estimator \eqref{eq:scoring_rule_uq_p} with $S_k$ directly yields tractable uncertainty measures, whose closed-form expressions for Gaussian and mixture distributions are derived in \autoref{app:derivation_scoring_rules}. Kernel scores have been increasingly applied in forecast evaluation and machine learning \citep{Gneiting.2007, doi:10.1137/22M1532184, 10.1093/jrsssb/qkae108}, including complex regression settings such as weather forecasting \citep{chen_generative, alet2025skillfuljointprobabilisticweather} or solving PDEs \citep{bultepno}.

Crucially, $D_k$ is essentially the squared distance between the kernel mean embeddings of the probability distributions into some Hilbert space \citep{steinwartStrictlyProperKernel2021} and is closely related to the Maximum Mean Discrepancy (MMD\textsuperscript{2}), a well-studied divergence in statistics and machine learning \citep{grettonKernelTwosampleTest2012, sejdinovicEquivalenceDistancebasedRKHSbased2013}. This connection provides both a theoretical basis and practical advantages that distinguish our framework from alternatives such as the log- or quadratic score.

\emph{Metric structure:} Under mild conditions, kernel scores are the only scoring rules that induce a valid metric on $\cP_k$ \citep[Theorem 19,][]{waghmareProperScoringRules2025}. Furthermore, existence only requires $H_k(\bP) < \infty$, which allows for measuring the divergence between continuous, discrete, or degenerate distributions, as opposed to other scoring rules that require absolute continuity with respect to the Lebesgue measure (compare \autoref{fig:ensemble_eu}).

\emph{Sample-based estimation:} The MMD\textsuperscript{2} (and therefore also $S_k$ and $H_k$) admits an unbiased empirical estimator \citep{grettonKernelTwosampleTest2012}; therefore, the uncertainty measures can be estimated consistently from samples alone. This makes the framework applicable to implicit or sample-based models, such as diffusion, or flow-based models, where likelihood evaluation is intractable in high dimensions. 

\emph{Structured domains:} The kernel $k$ can be adapted to the underlying output domain: stationary kernels for Euclidean regression, variogram-based kernels for spatial outputs \citep{VariogramBasedProperScoringRulesforProbabilisticForecastsofMultivariateQuantities}, graph kernels for molecular data \citep{JMLR:v11:vishwanathan10a}, or functional kernels for PDE solution spaces \citep{JMLR:v23:20-1180}. This flexibility is unique among common scoring rules and is central to providing domain-independent uncertainty measures.

\emph{Translation invariance and homogeneity:} When $k(\bx, \by) \equiv \kappa(\bx - \by), \ \bx, \by \in \cY$, for some conditionally negative definite function $\kappa: \mathcal{Y} \to \Ree$, the score is \emph{translation invariant}, i.e., $S_k(\bP,\boldsymbol{y}) = S_k(\bP_{\boldsymbol{h}}, \boldsymbol{y +  h})$ for $\boldsymbol{y,h} \in \mathcal{Y}$. A scoring rule is \emph{homogeneous of degree $\alpha$} if $S(\bP_c, c\by) = c^\alpha S(\bP, \by)$ for every $c>0, \by \in \mathcal{Y},\bP \in \mathcal{P}$ \citep{waghmareProperScoringRules2025}. This ensures that affine rescalings of the data do not change the relative performance assessment\textemdash{}a desirable invariance for regression tasks spanning different output scales.

\section{Properties of kernel scores as an uncertainty measure}
The properties of kernel scores described above carry over directly to the induced uncertainty measures. For instance, the ability to compare arbitrary distributions\textemdash{}including degenerate ones\textemdash{}via a sample-based estimator is particularly relevant when first-order distributions are combined via a linear pool \citep{combining_forecasts}, as is common in forecast ensembles \citep{repec:eee:jbfina:v:72:y:2016:i:s:p:s172-s186}. Beyond these inherited characteristics, we now show that principled choices of $k$ lead to uncertainty measures satisfying additional desirable properties, extending previous studies on axiomatic frameworks \citep{pmlr-v216-wimmer23a, hullermeier2022quantification, buelte2025axiomaticassessmententropyvariancebased}.
One trivial aspect of the corresponding measures is that they are all nonnegative, which follows directly from the kernel score being nonnegative.

Let $\bP \sim Q, \bP' \sim Q'$ be random first-order distributions with $Q, Q' \in \mathcal{P}(\mathcal{P}(\mathcal{Y}))$ and let $\delta_{\bP} \in \mathcal{P}(\cP(\mathcal{Y}))$ denote the Dirac measure at $\bP \in \cP(\mathcal{Y})$. For $\bP_1, \bP_2 \in \cP(\mathcal{Y})$ let $\leq_{\text{cx}}$ denote the convex order \citep{book}, meaning that $\bP_1 \leq_{\text{cx}} \bP_2 \iff \bE_{X\sim \bP_1}[\phi(X)] \leq \bE_{Y \sim \bP_2}[\phi(Y)]$ for all convex $\phi: \mathcal{Y} \to \Ree$. Similarly, for $Q_1, Q_2 \in \cP(\cP(\mathcal{Y}))$, let $\leq_{\text{cx}}^2$ denote the convex order with respect to all convex functionals $\Phi: \cP(\cY) \to \Ree$. In particular for $\bP_1 \leq_{\text{cx}} \bP_2$ it holds that $\bE_{X\sim \bP_1}[X] = \bE_{Y\sim \bP_2}[Y]$ and $\bV_{X\sim \bP_1}[X] \leq \bV_{Y\sim \bP_2}[Y]$, since the stochastic order is a measure of variability \citep{book}. We propose the following properties of the corresponding uncertainty measures, which are proved for both types of estimators in \autoref{app:proofs}.

First, we formalize the intuition that a fully concentrated second-order distribution, i.e., no model disagreement, should yield zero epistemic uncertainty. In addition, epistemic uncertainty should be monotone with respect to the convex order: a second-order distribution with greater spread over models implies greater epistemic uncertainty. This leads to the following proposition:

\begin{proposition}[Epistemic uncertainty]
\label{prop:eu}
    For any proper scoring rule $S$, for which the map $\bP \mapsto S(\bP, \bQ)$ is convex for fixed $\bQ$, it holds that
    \begin{enumerate}
        \item $~{Q=\delta_\bP \implies \eu(Q)=0}$, while for a strictly proper scoring rule the converse holds as well,
        \item $\eu(\delta_\bP) \leq \eu(Q_1) \leq \eu(Q_2), \quad \forall Q_1 \leq_{\mathrm{cx}}^2 Q_2$.
    \end{enumerate}
\end{proposition}

Similarly, if a \emph{first-order} predictive distribution has more variability, it should be assigned a higher value of \emph{aleatoric} uncertainty, as formalized in the following proposition:

\begin{proposition}[Aleatoric uncertainty]
\label{prop:au}
    Any kernel score $S_k$ with a translation invariant kernel $k(\bx,\bx')$ that is convex in one of its arguments fulfills $\au(\delta_{\bP_1}) \leq \au(\delta_{\bP_2}), \ \forall \bP_1 \leq_{\mathrm{cx}} \bP_2$.
\end{proposition}

Finally, we want to analyze how robust an uncertainty measure is to deviations in the second-order distribution. We consider robustness in terms of the influence function \citep[][Chapter 2]{hampel}, which analyzes the limiting behavior if the underlying (second-order) distribution is perturbed by a single point diverging to infinity. If the influence function is bounded, any outlier in $Q$ can only have a finite impact on the estimation of the uncertainty measure $M$, making it robust against such outliers. This is formalized in the following proposition:

\begin{proposition}[Robustness]
\label{prop:robustness}
Consider a parametric first-order distribution $\bP_{\vtheta} \in \mathcal{P}(\mathcal{Y})$ with $\vtheta \in \Theta \subseteq \Ree^p$, a second-order distribution $Q \in \mathcal{P}(\Theta)$, and $\vvtheta \sim Q$. Let $S_k$ be a
kernel score with bounded kernel, i.e., $\|k\|_\infty \coloneq \sup_{\bx,\by \in \mathcal{Y}} |k(\bx, \by)| < \infty$
and let
$Q_\varepsilon \coloneq (1-\varepsilon)Q + \varepsilon \delta_{\vtheta_0}$,
$\vtheta_0 \in \Theta$, with influence function
\begin{align*}
    \mathrm{IF}(\vtheta_0; M, Q)
    \coloneq \lim_{\varepsilon \to 0} \frac{M(Q_\varepsilon) - M(Q)}{\varepsilon}.
\end{align*}
Then, for each uncertainty measure $M \in \{\au, \eu\}$, we have $M(Q) < \infty$, and
\begin{align*}
    \sup_{\vtheta_0 \in \Theta} \left| \mathrm{IF}(\vtheta_0; M, Q) \right|
    \;\le\; C_M \, \|k\|_\infty \;<\; \infty,
\end{align*}
for some constant $C_M$ so $M$ is robust in terms of the influence function.
\end{proposition}

Together, Propositions~\ref{prop:eu}–\ref{prop:robustness} characterize certain desirable behavior of uncertainty measures: epistemic uncertainty vanishes if and only if all models agree, aleatoric uncertainty increases with the variability of the predictive distribution, and neither measure can be destabilized by outlying ensemble members when $k$ is bounded. Crucially, these properties are not guaranteed by properness alone\textemdash{}they depend on the specific choice of kernel. The propositions, therefore, serve as a principled guide for selecting $k$ in dependence on the underlying task and corresponding requirements.
In particular, we propose the following instantiations, with closed-form expressions derived in \autoref{app:derivation_scoring_rules}.

\emph{Energy score: }The energy score ($S_\mathrm{ES}$), with kernel ${k(\bx,\bx')=\|\bx - \bx' \|}$ is strictly proper, translation invariant, and homogeneous, satisfying both Propositions~\ref{prop:eu} and~\ref{prop:au}. In fact, it is the unique homogeneous translation invariant kernel score on $\Ree^d$. Its univariate special case $d=1$ recovers the continuous ranked probability score, arguably one of the most widely used proper scoring rules in regression settings \citep{Gneiting.2007}. Any univariate strictly proper score, such as the CRPS, can be extended to a multivariate strictly proper rule via marginal averaging (see Appendix~\ref{app:derivation_scoring_rules}), which we denote by $S_\mathrm{CRPS}$. However, in that case, the dependence structure is not accounted for.

\emph{Gaussian kernel score: } The (negative) Gaussian kernel score ($S_{k_\gamma}$) corresponding to the kernel ${k(\bx,\bx') = - \exp \left( - \|\bx-\bx' \|^2 / \gamma^2 \right)}$ and bandwidth $\gamma>0$ is strictly proper, satisfies Propositions~\ref{prop:eu}, but fails \ref{prop:au} due to the kernel not being convex. As the only bounded kernel among our proposals, it is however, the only one satisfying the robustness condition of Proposition~\ref{prop:robustness}, making it the most conservative choice when outlying ensemble members are anticipated. 

\emph{Squared-error:} Finally, the squared-error ($S_\mathrm{SE}$) with kernel $k(\bx,\bx') = \|\bx-\bx'\|^2$ falls within the kernel score framework and recovers the commonly-used variance-based uncertainty measure in the univariate case \citep{aminiDeepEvidentialRegression2020}, providing a natural link to already existing measures.
However, since it is not strictly proper, it fails the converse in Proposition~\ref{prop:eu} and has stronger assumptions (existence of second moments) than the other scoring rules. We therefore include it primarily as a baseline for comparison rather than as a recommended instantiation.

For completeness, while the strictly proper \emph{log-score} ($S_\mathrm{log}$) is not a kernel score, it satisfies Propositions~\ref{prop:eu} and~\ref{prop:au} under standard regularity conditions (see \autoref{app:proofs}) and leads to the well-known entropy-based uncertainty measure \citep{kendall2017uncertainties}. However, it requires absolute continuity with respect to the Lebesgue measure, density evaluation and can assign negative uncertainty values, limiting its applicability as a general uncertainty measure. In summary, we propose the kernel-based uncertainty decomposition \eqref{eq:scoring_rule_uq_p} instantiated with $S_\mathrm{ES}, S_\mathrm{CRPS}, S_{k_\gamma}$ as principled uncertainty measures for regression, supported by the theoretical properties of the measures itself, as well as the guarantees established in Propositions~\ref{prop:eu}–\ref{prop:robustness}. The log-score and squared-error remain valid instantiations within the scoring rule framework, serving as a natural ground for comparison.

\section{Numerical experiments}
We evaluate our kernel-based uncertainty measures across four experimental protocols: robustness evaluation, selective prediction, out-of-distribution detection, and active learning, probing complementary aspects of uncertainty quality from calibration under shift to data-efficient acquisition. 

\begin{table}[h]
    \centering
    \caption{Predictive uncertainty representations used across the experiments.}
    \label{tab:methods_main}
    \begin{tabular}{lll}
        \toprule
        \textbf{Method}  & \textbf{Predictive form} & \textbf{Second-order} \\
        \midrule
        NG              
            & $\mathcal{N}(\mu, \sigma^2)$ 
            & Ensemble \\ DER
          
            & $t_{2\alpha}(\cdot; \gamma, \frac{\beta(1+\upsilon)}{\upsilon \alpha}, 2\alpha)$   
            & Conjugate prior \\
            LoRa       
      
            & $\mathcal{N}(\bm\mu,\, \bm U\bm U^\top + \bm D)$                 
            & Ensemble \\
        MDN        
        
            & $\sum_k w_k\,\mathcal{N}(\mu_k, \sigma_k^2)$                 
            & Ensemble \\
          SB
 
            & $\frac{1}{N}\sum_n\delta_{\by_n}$             
            & Ensemble\\
        \bottomrule
    \end{tabular}
\end{table}

To demonstrate that the proposed measures are agnostic to the choice of predictive model, we evaluate them across five uncertainty representation methods, summarized in \autoref{tab:methods_main}. The natural Gaussian (NG) method \citep{immer2023effective} uses a predictive univariate normal distribution with a second-order ensemble, while the deep evidential regression (DER) approach \citep{aminiDeepEvidentialRegression2020} uses the corresponding conjugate prior. Further, we consider a univariate mixture density network (MDN) \citep{bishop} and a multivariate Gaussian with a low-rank covariance matrix (LoRa) \citep{rezendeStochasticBackpropagationApproximate2014}, both using second-order ensembling. Finally, we utilize a nonparametric sampling-based generative model (SB) \citep{JMLR:v25:23-0038}.
These models cover a variety of predictive representations, including closed-form, sampling-based, multimodal, or multivariate.

\begin{figure}[ht]
    \centering
    \includegraphics[width=\linewidth]{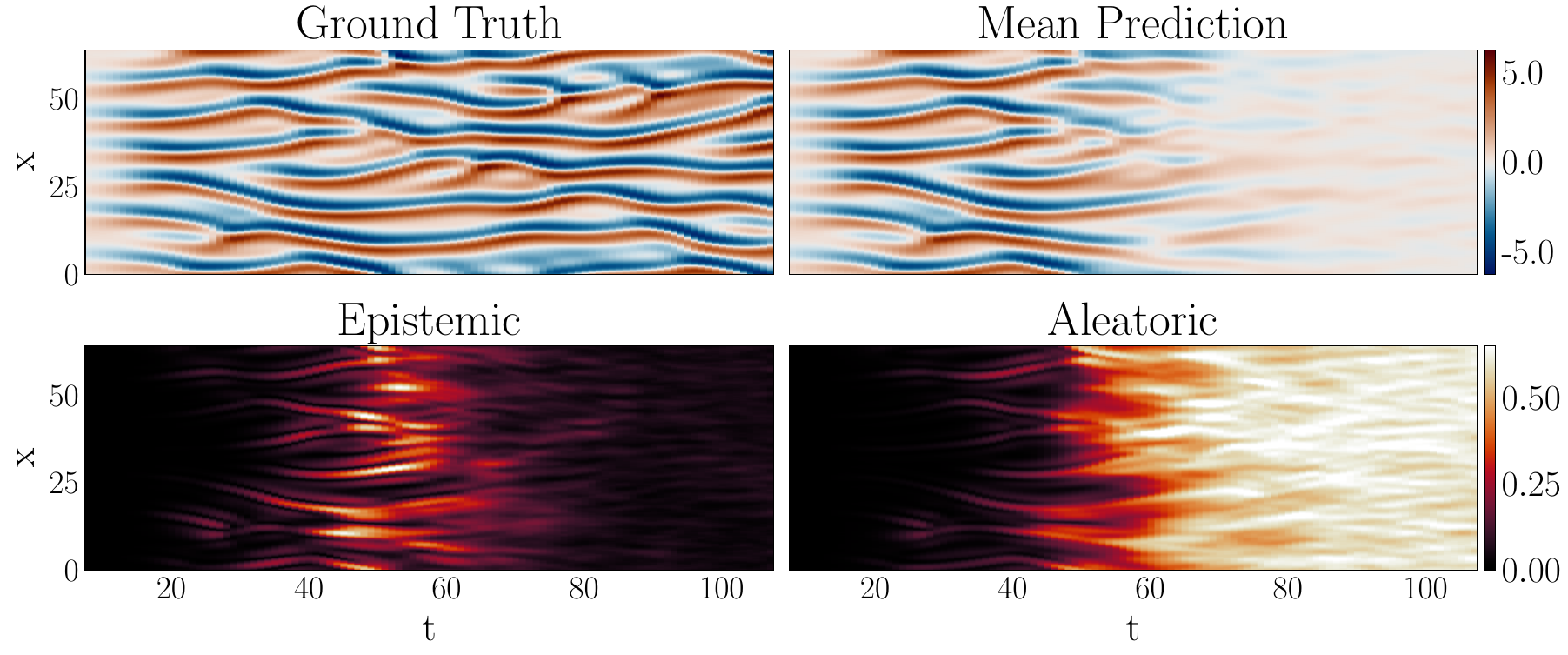}
    \caption{Predictions for the (chaotic) Kuramoto-Sivashinsky equation using the sampling-based method and corresponding estimates of epistemic and aleatoric uncertainty using the $S_\mathrm{CRPS}$ measure. Both components show structural consistency with the underlying task; aleatoric uncertainty increases with time, as physical predictability decreases, while epistemic uncertainty peaks around the predictability limit of the system.}
    \label{fig:ks_main}
\end{figure}

Further, we consider a variety of benchmark datasets, covering different predictive tasks, dimensions, and data modalities. In particular, we consider the UCI dataset for univariate regression \citep{hernándezlobato2015probabilisticbackpropagationscalablelearning}, two one-dimensional PDE prediction tasks \citep{NEURIPS2022_0a974713}, and two two-dimensional vision tasks, namely depth regression \citep{aminiDeepEvidentialRegression2020} and surface temperature prediction \citep{rasp2024weatherbench2benchmarkgeneration}.
Note that for the PDEs, we use the probabilistic neural operator \citep{bultepno} as the sample-based method, which generates solution samples in the corresponding function space. However, the properties of our selected kernels also hold in the corresponding Hilbert spaces \citep{ziegelCharacteristicKernelsHilbert2024}, highlighting the broad applicability of our framework. A sample prediction and corresponding uncertainty estimates are shown in \autoref{fig:ks_main}.

As uncertainty quantification baselines, we include the log-score $S_\mathrm{log}$ and the squared-error $S_\mathrm{SE}$, which are commonly used in practice but either lack the theoretical guarantees established in the previous section or do not fit into the kernel framework at all. The bandwidth for the Gaussian kernel score $S_{k_\gamma}$ is chosen via the median heuristic \citep{garreau2018largesampleanalysismedian} on each dataset, which we found to perform reliably across tasks.

In the following, we use the pairwise estimator throughout, as it admits closed-form expressions for all considered first-order distributions (compare \autoref{app:derivation_scoring_rules}). Detailed descriptions of the experimental setup, as well as additional results and visualizations are provided in \autoref{app:experiment_details}. For completeness, we also provide an analysis of computational complexity and approximation error of the different instantiations, as well as a corresponding empirical runtime analysis in \autoref{app:computational_complexity}.

\subsection{Robustness analysis}
\begin{table}[ht]
    \centering
    \caption{MAPE ($\downarrow$) of AU and EU estimates on the concrete dataset, comparing a base ensemble of $M=25$ members against an augmented ensemble with one additional member trained on targets distorted by $\Tilde{y} = y + \mathcal{N}(0, \delta^2)$.}
    \label{tab:robustness_concrete}
\begin{tabular}{lllll}
\toprule
& \multicolumn{4}{c}{Aleatoric} \\
\midrule
$S/ \delta$ & 0.0& 1.0 & 2.5 & 5.0  \\
\midrule
$S_\mathrm{log}$ & 0.2 & 3.2 & 4.5 & 4.8  \\
$S_\mathrm{SE}$ & 1.1 & \num{4.6e3} & \num{6.8e4}& \num{4.8e5} \\
$S_\mathrm{ES}$ & 0.6 & \num{6.7e1} &  \num{2.2e2} & \num{5.0e2}   \\
$S_{k_\gamma}$ & 0.0 & 0.2 & 0.2 & 0.2  \\
\midrule
& \multicolumn{4}{c}{Epistemic} \\
\midrule
$S_\mathrm{log}$ &  3.5 & \num{1.5e4} & \num{5.3e4} & \num{2.4e5}\\
$S_\mathrm{SE}$  & 3.7 & \num{3.0e4} & \num{6.2e4} & \num{4.7e5} \\
$S_\mathrm{ES}$ & 2.7 & \num{8.1e2} & \num{1.2e3} & \num{4.1e3} \\
$S_{k_\gamma}$ & 1.6 & \num{1.3e1} & \num{1.1e1} & \num{1.3e1} \\
\bottomrule
\end{tabular}
\end{table}
To empirically validate the robustness (in terms of the influence function) of different measures, we use three datasets from the UCI benchmark \citep{hernándezlobato2015probabilisticbackpropagationscalablelearning} and train a deep ensemble \citep{lakshminarayananSimpleScalablePredictive2017} on each task. Then, we train one additional ensemble member using a target variable with added noise, i.e. $\Tilde{y} = y + \mathcal{N}(0,\delta^2)$ with gradually increasing noise. While this is a synthetic outlier creation, it allows for comparing the robustness of each uncertainty measure and the corresponding scoring rule to a single corrupted ensemble member.
To measure the deviation, we use the mean absolute percentage error (MAPE) with respect to the uncertainty in the base ensemble, i.e.,
$$
\text{MAPE} \coloneq \frac{100}{n} \sum_{i=1}^n \left| \frac{M_i(Q^\delta)-M_i(Q)}{M_i(Q)}\right|,
$$
where $M_i \in \{ \mathrm{AU, EU}\}$ denotes the uncertainty estimate at input $\bx_i, \ i = 1,\ldots, n$, $Q$ denotes the base ensemble and $Q^\delta$ denotes the corrupted ensemble.
\autoref{tab:robustness_concrete} shows the results for the concrete dataset. The Gaussian kernel score $S_{k_\gamma}$ remains stable across all distortion levels for both aleatoric and epistemic uncertainty. In contrast, the other measures, most notably the squared-error degrade by several orders of magnitude even at moderate $\delta$, consistent with their unbounded influence functions.

\subsection{Selective Prediction}
In selective prediction, the model is evaluated only on parts of the (test-) dataset, typically a specific subset with low uncertainty. Therefore, this task assesses the ability of the uncertainty measure to indicate whether a prediction is correct or not. Here, one typically uses total uncertainty \citep{kotelevskii2025from, hofman2025uncertaintyquantificationmachinelearning}, as neither component, aleatoric or epistemic, determines the prediction correctness alone.
The performance for selective prediction is measured using prediction-reject-ratios \citep[PRRs,][]{malinin2021uncertaintyestimationautoregressivestructured}, which are negatively oriented (lower is better) of inaccurate predictions using the corresponding uncertainty measure.
We evaluate this experiment for all datasets and uncertainty representation methods; \autoref{tab:selective_prediction_ranks} shows the corresponding average ranks, while the full results table and selected visualizations are available in Appendix~\ref{app:selective_prediction}.

\begin{table}[ht]
    \centering
    \caption{Average rank ($\downarrow$) of the PRR aggregated over all datasets, with the best measure in bold. The total average is calculated per task, i.e. Univariate, 1D, and 2D.}
    \label{tab:selective_prediction_ranks}
    \begin{tabular}{lccccc}
    \toprule
       Method  & $S_\mathrm{log}$  & $S_\mathrm{SE}$  & $S_\mathrm{CRPS}$  & $S_\mathrm{ES}$  & $S_{k_\gamma}$\\
       \midrule
        NG & \textbf{2.56} & 2.61 & 2.61 & 2.61 & 2.67\\
       DER  & 3.39 & \textbf{1.94} & 3.00 & 2.56 & 2.44\\
       MDN  & - & 1.78 & \textbf{1.67} & 1.89 & 2.67\\
        LoRa & 5.00  & 2.00 & 4.00 & 3.00 & \textbf{1.00}\\
       SB  & - & 2.45 & 2.41 & 2.05 & \textbf{1.77}\\
       \midrule
      Task-weighted & 3.51 & 2.34 & 2.69 & 2.43 & \textbf{2.29}\\
       \bottomrule
    \end{tabular}
\end{table}

Overall, $S_{k_\gamma}$ achieves the best aggregated performance across all tasks, with $S_\mathrm{ES}$ and $S_\mathrm{CRPS}$ obtaining slightly worse but comparable ranks. For more complex multimodal or multivariate representation methods these strictly proper scoring rules consistently outperform alternatives, which follows directly from Proposition~\ref{prop:eu}: scoring rules sensitive to distributional shape beyond the first two moments are better equipped to reflect uncertainty in more expressive predictive distributions. The comparatively weaker performance of $S_\mathrm{SE}$ for the multimodal methods is therefore expected: reducing uncertainty to a point estimate of the predictive mean cannot distinguish between, e.g., a peaked unimodal and a diffuse multimodal distribution. Nevertheless, $S_\mathrm{SE}$ remains competitive, however, only for methods assuming a unimodal or multivariate Gaussian output, where the mean is a sufficient summary.

\subsection{Out-of-distribution detection}
Out-of-distribution (OOD) detection is a commonly used task to assess and compare the quality of uncertainty measures and uncertainty quantification methods. In essence, a model is trained on in-distribution (ID) data and its predictions, as well as corresponding uncertainty estimates, are compared between ID and OOD data. Since the model has not seen the OOD data before, it should assign higher epistemic uncertainty to those inputs.
\begin{figure}[t]
    \centering
    \includegraphics[width=\linewidth]{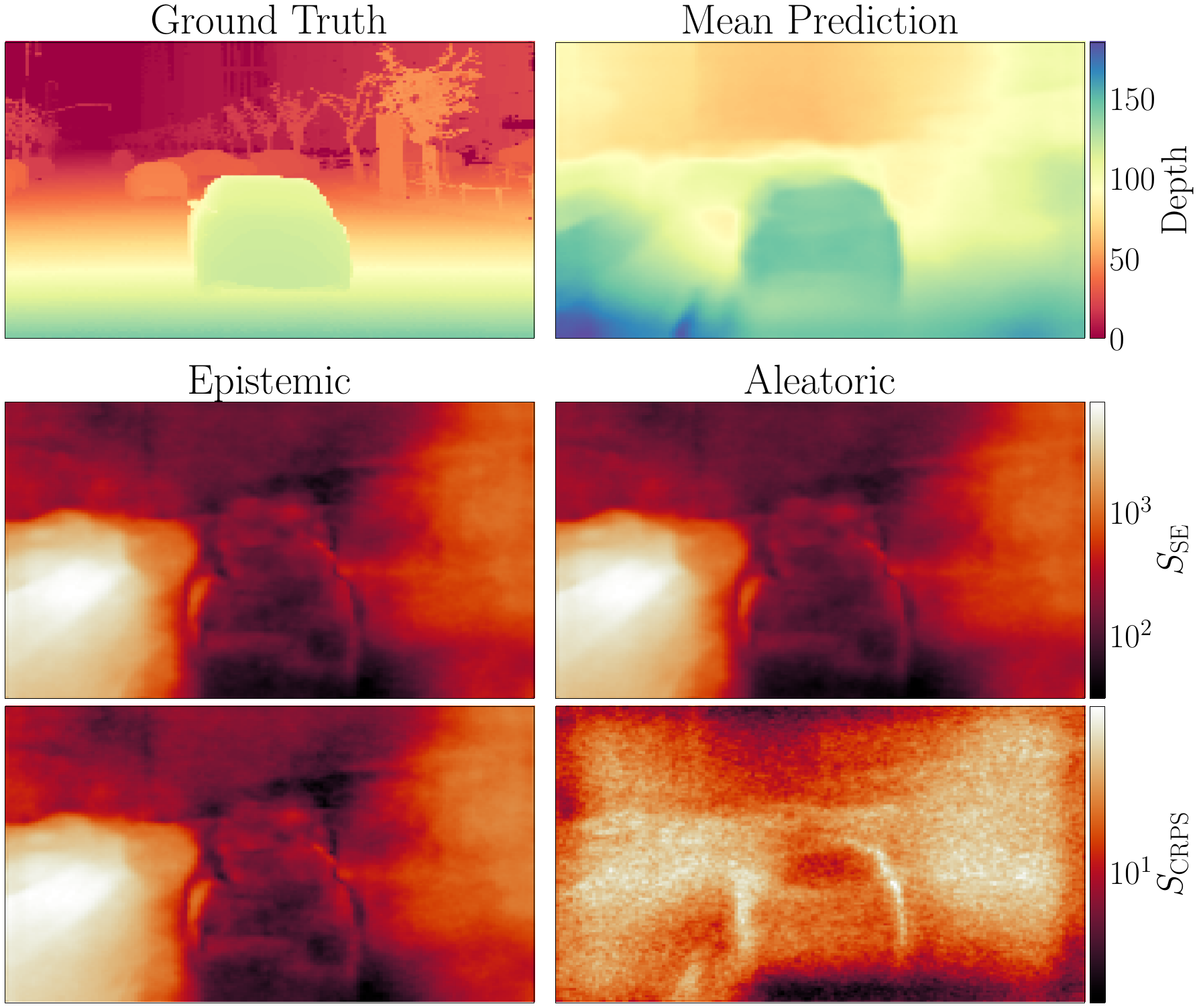}
    \caption{Predictions of the sampling-based method for the depth regression OOD dataset, as well as uncertainty estimates of the (pointwise) measures $S_\mathrm{SE}$ and $S_\mathrm{CRPS}$.}
    \label{fig:ood_depth_main}
\end{figure}
\begin{table}[ht]
    \centering
    \caption{Average rank ($\downarrow$) of the AUROC for out-of-distribution detection for each method aggregated over all datasets used. The best measure is highlighted in bold.}
    \label{tab:ood_rank}
    \begin{tabular}{lccccc}
    \toprule
       Method  & $S_\mathrm{log}$  & $S_\mathrm{SE}$  & $S_\mathrm{CRPS}$  & $S_\mathrm{ES}$  & $S_{k_\gamma}$\\
       \midrule
       NG &\textbf{ 2.00} & 4.00 & 2.50& 3.00 & 3.50\\
       DER & \textbf{1.00} & 3.50 & 3.00 &4.50 & 3.00\\
       MDN & - & 3.25 & 2.25 & \textbf{1.75} & 2.75 \\
       LoRa & 5.00 & 3.25 & \textbf{1.75} & 2.25 & 2.75 \\
       SB & - & 3.25 & \textbf{2.00} & \textbf{2.00} &2.75\\
       \midrule
       Average & 2.67 & 3.42 & \textbf{2.25} & 2.58 & 2.92 \\
       \bottomrule
    \end{tabular}
\end{table}

We generate OOD data for the 1D PDE tasks by changing the underlying coefficients of the PDE (i.e., viscosity and length scale) and for the 2D tasks using a domain shift (different scene for the depth regression and a different geographical domain for the weather prediction task).
To evaluate the performance of the different measures, we evaluate the AUROC of the uncertainty scores between OOD and ID samples. \autoref{tab:ood_rank} shows the corresponding rank of the uncertainty measures, averaged across the uncertainty representation methods.

Overall, the marginal score $S_\mathrm{CRPS}$ shows the best performance, followed by its multivariate version $S_\mathrm{ES}$. Here, the log-score $S_\mathrm{log}$ performs quite well for a first-order univariate Gaussian, but does not lead to a good OOD recognition for the multivariate Gaussian. In this experiment, $S_\mathrm{SE}$ performs worst across all measures. \autoref{fig:ood_depth_main} provides an exemplary visualization of the out-of-distribution prediction for the ApolloScape dataset and the sampling-based method. It is evident that the $S_\mathrm{SE}$ measure leads to almost identical predictions for AU and EU, while the $S_\mathrm{CRPS}$ measure shows better disentangled uncertainty estimates.

It is worth noting that OOD detection benchmarks are inherently difficult to construct in a way that is both realistic and discriminative \citep{hofman2025uncertaintyquantificationmachinelearning, li2025outofdistributiondetectionmethodsanswer}, as detection difficulty is inseparable from how the distribution shift is defined.
This is reflected in \autoref{tab:ood_full}, where AUROC values are near one across almost all datasets and methods\textemdash{}a result that should be interpreted carefully, since not all distribution shifts meaningfully increase predictive difficulty. 
Crucially, this near-ceiling performance is nonetheless consistent across diverse datasets, shift types, and uncertainty representation methods, providing broader evidence for the reliability of kernel scoring rules as uncertainty measures beyond what any single dataset could establish.

\subsection{Active learning}
\begin{table}[ht]
    \centering
    \caption{Average rank ($\downarrow$) of the final test loss in the active learning setting aggregated over all datasets, with the best measure in bold. Here, $\mathcal{U}$ denotes the random baseline.}
    \label{tab:active_learning_rank}
    \begin{tabular}{lcccccc}
    \toprule
       Method  & $S_\mathrm{log}$  & $S_\mathrm{SE}$  & $S_\mathrm{CRPS}$  & $S_\mathrm{ES}$  & $S_{k_\gamma}$ & $\mathcal{U}$ \\
       \midrule
        NG & 3.57 & 4.00 & 4.00 & 4.00 & \textbf{1.43} & 2.00 \\
       DER  & 3.57 & 4.00 & \textbf{1.57} & \textbf{1.57} & 2.86 & 3.00\\
       MDN  & - & 2.57 & 3.29 & 3.29 & \textbf{1.57} & 2.57 \\
       SB  & - & 2.22 & 2.00 & \textbf{1.67} & 2.44 & 4.11\\
       \midrule
      Average & 3.57 & 3.13 & 2.67 & 2.57 & \textbf{2.10} & 3.00\\
       \bottomrule
    \end{tabular}
\end{table}

As a final task, we consider an active learning experiment, which is frequently used to evaluate probabilistic predictions and uncertainty measures. Here, the objective is to select new training instances under a computational budget, using epistemic uncertainty as the selection criterion \citep{nguyen, NEURIPS2019_95323660}. Starting from a small training set, the learner iteratively selects new datapoints based on the corresponding (predictive) uncertainty to minimize the corresponding loss function with as few labels as possible. We use this task as a comparison ground for our different estimators of epistemic uncertainty.

Due to the high computational load of the 2D tasks, we focus on the univariate and 1D PDE datasets, which still offer a diverse ground for comparison. We split the data into train, validation, and test datasets and use 5\% of the training data size for a random initialization. In each of 20 rounds, the learner acquires 1\% of the dataset size as new datapoints. In each round, $M=5$ ensemble members are trained for 50 epochs from scratch, and the whole experiment is repeated across three independent seeds.
\autoref{tab:active_learning_rank} shows the performance ranks of the uncertainty measures per representation method, while \autoref{tab:active_learning_full} provides the full results. Here, the three strictly proper scoring rules ($S_\mathrm{CRPS}$, $S_\mathrm{ES}$, $S_{k_\gamma}$) lead to the best performance. In particular, $S_{k_\gamma}$ obtains a significantly lower rank than the comparison methods. As opposed to the other experiments, the methods $S_\mathrm{log}$ and $S_\mathrm{SE}$ do not perform well, even worse than the random baseline $\mathcal{U}$, also for the first-order Gaussian predictions.

\paragraph{Findings \& Insights}
Across four evaluation protocols spanning various data domains, the kernel-based (strictly proper) scoring rules demonstrate consistently strong performance. $S_{k_\gamma}$ achieves the best overall rank in selective prediction and active learning, while $S_\mathrm{CRPS}$ and $S_\mathrm{ES}$ lead in out-of-distribution detection. The strictly proper kernel scores particularly outperform $S_\mathrm{log}$ and $S_\mathrm{SE}$ for expressive predictive distributions beyond unimodal Gaussians, where sensitivity to distributional shape beyond the first two moments is critical. Additionally, $S_{k_\gamma}$ remains stable under ensemble corruption, whereas other measures degrade by orders of magnitude.
Overall, our proposed measures consistently outperform the baselines: on every task, at least one kernel-based measure outperforms both baselines, and each kernel-based measure outperforms the baselines on the majority of tasks. Although no single uncertainty measure dominates uniformly across all tasks and representation methods, this is expected given the broad framework and the fact that different kernels lead to different uncertainty assessments. However, in general, the kernel scores that fulfill the posed theoretical properties collectively offer the most reliable uncertainty quantification.

\section{Related work}
\emph{Novel uncertainty measures.}
Many studies focus on quantifying uncertainty for predictive models, especially for classification. While the most commonly used measures are based on the Shannon entropy \citep{houlsby2011bayesian}, those have been criticized for having undesirable properties \citep{pmlr-v216-wimmer23a}. Several generalizations have been proposed, such as variance-based \citep{sale2023secondorderuncertaintyquantificationvariancebased}, distance-based \citep{sale2023secondorderuncertaintyquantificationdistancebased} or pairwise \citep{schweighofer2023introducingimprovedinformationtheoreticmeasure, berry2024efficientepistemicuncertaintyestimation} estimators. Closest to our work are recent developments in deriving uncertainty measures based on proper scoring rules and divergences. \cite{gruber2023uncertaintyestimatespredictionsgeneral, adlam2022understandingbiasvariancetradeoffbregman} derive a bias-variance decomposition based on Bregman divergences, which was extended to kernel scores by \cite{gruber2024a}, where the corresponding uncertainty measures are conceptually similar to our proposed ones. However, their study focuses on generative models and on assessing predictive performance. Recently, \citet{kotelevskii2025from, hofman2024quantifying} introduced a framework for decomposing and quantifying uncertainty based on proper scoring rules, which was extended to the univariate (Gaussian) regression case \citep{fishkov2025uncertaintyquantificationregressionusing}. While similar in nature, our work specifically considers kernel scores with advantageous properties and works in more general regression domains, moving away from the univariate Gaussian assumptions to more complex uncertainty representation.

\emph{Uncertainty quantification in regression.}
While many works focus on uncertainty representation in regression, for example, via second-order distributions \citep{aminiDeepEvidentialRegression2020, meinert2022multivariatedeepevidentialregression, malinin2020regressionpriornetworks} or ensembles \citep{berryNormalizingFlowEnsembles2023, lakshminarayananSimpleScalablePredictive2017, kelen2025distributionfree}, little is usually done in the direction of analyzing the underlying uncertainty measures. The studies usually employ either the variance-based measure \citep{aminiDeepEvidentialRegression2020, meinert2022multivariatedeepevidentialregression, 9857056} or (a variant of) the entropy-based measure \citep{malinin2020regressionpriornetworks, berry2024efficientepistemicuncertaintyestimation, postels2021hiddenuncertaintyneuralnetworks}. While \cite{buelte2025axiomaticassessmententropyvariancebased} compare both measures with respect to a given set of preferable properties, they do not consider other measures or the pairwise variants thereof. In contrast, our work proposes a general way to construct uncertainty measures that can be used with many different instantiations, leading to different properties.

\section{Conclusion}
We propose a general framework for uncertainty quantification in regression, based on strictly proper kernel scores, which encompasses different uncertainty measures in a single principled construction.
In particular, it turns the design of uncertainty measures into the selection of an underlying kernel $k$, providing a systematic way for domain or task-specific constructions.
Our analysis shows how structural properties of kernels, such as strictly properness or boundedness, directly translate into distinct characteristics of the induced uncertainty measures, allowing practitioners to design measures aligned with specific requirements. 
Beyond theoretical contributions, our empirical results demonstrate the validity of the proposed measures, yielding consistently strong performance across diverse data modalities, uncertainty representations, and benchmark tasks.

\textbf{Limitations and future work}
While our framework provides a principled foundation, it is not unique; alternative measures may satisfy the same properties. 
Developing principled selection procedures for choosing or learning suitable kernels remains an important direction for future research.
In particular, exploring alternative score constructions, such as weighted kernel scores \citep{doi:10.1137/22M1532184}, different kernel families (e.g., Laplace or inverse-multiquadratic), or learnable kernels could enable more targeted sensitivity to specific phenomena, such as extreme events. Similarly, improving the bandwidth selection procedure within the kernel, for example, by mixing different scales, could further improve empirical performance.
In general, our work focused on a specific set of theoretical properties, and extending the analysis to other aspects such as computational efficiency, scalability, or interpretability could broaden the framework and extend its applicability.
Exciting opportunities also lie in exploring other data domains, such as graph-structured data, where the usage of kernel scores could open up new possibilities of uncertainty quantification in, for example, molecule design.
Empirically, extending the evaluation to a broader range of generative models, including diffusion and flow-based architectures, may provide further insight into the practical behavior of kernel-based scores. Finally, further theoretical investigation of the relationship between kernel scores and maximum mean discrepancy could yield new insights into their geometric and statistical properties, potentially guiding the principled design of uncertainty measures that are optimal for a specific downstream task. 

\begin{acknowledgements}
The authors acknowledge support by the DAAD programme Konrad Zuse Schools of Excellence in Artificial Intelligence, sponsored by the Federal Ministry of Research, Technology and Space.

C. Bülte and G. Kutyniok acknowledge support by the German Research Foundation under the grant DFG-SPP-2298.

E. H\"ullermeier acknowledges support by the German Research Foundation under the grant GRK 3081 (project number 534429653).

G. Kutyniok also acknowledges support by the gAIn project, which is funded by the Bavarian Ministry of Science and the Arts (StMWK Bayern) and the Saxon Ministry for Science, Culture and Tourism (SMWK Sachsen). Furthermore, G. Kutyniok is supported by LMUexcellent, funded by the Federal Ministry of Education and Research (BMBF) and the Free State of Bavaria under the Excellence Strategy of the Federal Government and the Länder as well as by the Hightech Agenda Bavaria.

\end{acknowledgements}

\bibliography{ref}

\newpage

\onecolumn

\title{Uncertainty Quantification for Regression:\\ A Unified Framework based on kernel scores\\(Supplementary Material)}
\maketitle
\vspace{0.5cm}
\appendix

\section{Proofs}
\label{app:proofs}
\subsection{Proofs of Propositions \ref{prop:eu} - \ref{prop:robustness}}

\begin{proof}[Proof of Proposition \ref{prop:eu}]
Here we prove that for any proper scoring rule $S$, it holds that
    \begin{enumerate}
        \item $~{Q=\delta_\bP \implies \eu(Q)=0}$, while for a \emph{strictly} proper scoring rule the converse holds as well,
        \item $\eu(\delta_\bP) \leq \eu(Q_1) \leq \eu(Q_2)$.
    \end{enumerate}

1. Consider the BMA estimator. For $Q=\delta_\bP$ we have $\overline{\bP} = \bP$ and $\eu(Q) = \bE_{\bP \sim Q}[D(\overline{\bP}, \bP)] = D(\bP,\bP) = 0$, since $D$ is a divergence.
For a strictly proper scoring rule, we obtain
\begin{align*}
    \eu(Q) = \bE_{\bP \sim Q}[D(\overline{\bP}, \bP)] = 0 \implies \overline{\bP} = \bE_Q[\bP] = \bP \implies Q = \delta_\bP.
\end{align*}
For the pairwise estimator, the proof works in an analogous way.

2 (BMA). The lower bound follows immediately from the nonnegativity of the divergence $D$ and the first part of the proposition being fulfilled for a proper scoring rule. Furthermore, we are given $Q_1 \leq_{\mathrm{cx}}^2 Q_2$ and $\eu(Q) = \bE_{\bP \sim Q}[D(\overline{\bP}, \bP)]$. 
Recall that for a scoring rule with $\bP, \bQ \in \cP({\cY})$, the divergence is given as $D(\bP, \bQ) = S(\bP, \bQ) - S(\bQ, \bQ).$
    We want to show that
    \begin{align*}
        \eu(Q_1) = \bE_{\bP \sim Q_1}[D(\overline{\bP}, \bP)] \leq \bE_{\bP \sim Q_2}[D(\overline{\bP}, \bP)] = \eu(Q_2).
    \end{align*}
    We will show that $D(\overline{\bP},\bP)$ is a convex functional in $\bP$. Then, by definition of the convex order, it follows that $\eu(Q_1) \leq \eu(Q_2)$. 
    
    First, note that by definition of the convex order we have a fixed $\overline{\bP} = \bE_{\bP \sim Q_1}[\bP] =  \bE_{\bP \sim Q_2}[\bP]$. By definition of proper scoring rules, the term $S(\bP, \bQ)$ is affine in $\bQ$ \citep{dawidGeometryProperScoring2007} and therefore convex. Furthermore, we know that $H(\bQ) = S(\bQ, \bQ)$ is a concave function in $\bQ$ \citep{waghmareProperScoringRules2025} and therefore $-H(\bQ)$ is convex. In total, $D(\overline{\bP}, \bP)$ consists of an affine function plus a convex function in $\bP$ and is therefore also convex in $\bP$ \citep{Boyd_Vandenberghe_2004}.

2 (Pairwise). For the pairwise estimator, we require the additional assumption that for a fixed $\bQ$, the map $\bP \mapsto S(\bP, \bQ)$ is convex, which is fulfilled by kernel scores or scoring rules of Bregman type. First, write $F(\bP,\bP') := D(\bP , \bP') = S(\bP,\bP') - H(\bP')$.

By the convexity assumption, for fixed $\bP'$, the map $\bP \mapsto F(\bP,\bP')$ is convex.  
Furthermore, since $S(\bP, \bQ)$ is affine in $\bQ$ and $H(\bP)$ is concave \citep{dawidGeometryProperScoring2007}, for fixed $\bP$, the map $\bP' \mapsto F(\bP,\bP')$ is affine $+\,$convex, hence convex.

For every fixed $\bP'$, we obtain the following via the convex order
\[
\bE_{\bP \sim Q_1} F(\bP,\bP') \;\leq\; \bE_{\bP \sim Q_2} F(\bP,\bP').
\]
Integrating over $\bP' \sim Q_1$ gives
\[
\bE_{\bP,\bP' \sim Q_1} F(\bP,\bP') \;\leq\; \bE_{\bP' \sim Q_1} \bE_{\bP \sim Q_2} F(\bP,\bP').
\]

Similarly, for every fixed $\bP$, we obtain
\[
\bE_{\bP' \sim Q_1} F(\bP,\bP') \;\leq\; \bE_{\bP' \sim Q_2} F(\bP,\bP').
\]
Integrating over $\bP \sim Q_2$ gives
\[
\bE_{\bP \sim Q_2} \bE_{\bP' \sim Q_1} F(\bP,\bP') \;\leq\; \bE_{\bP,\bP' \sim Q_2} F(\bP,\bP').
\]

Since both sides coincide (by Fubini's theorem), we ultimately get
\[
\bE_{\bP,\bP' \sim Q_1} F(\bP,\bP') \;\leq\; \bE_{\bP,\bP' \sim Q_2} F(\bP,\bP'),
\]
i.e.
\[
\eu_{P}(Q_1) \;\leq\; \eu_{P}(Q_2).
\]
\end{proof}

\begin{proof}[Proof of Proposition \ref{prop:au}]
Here we prove that any kernel score $S_k$ with a translation invariant kernel $k(x,x')$ that is convex in one of its arguments fulfills $\au(\delta_{\bP_1}) \leq \au(\delta_{\bP_2})$.

We know by assumption that $\bP_1 \leq_{\mathrm{cx}} \bP_2$ and $\au(\delta_\bP) = H(\bP)$. Therefore, we need to show that $H(\bP_1) \leq H(\bP_2)$. Recall that for any translation invariant kernel score we have $k(x,x') = \psi(x-x')$ for some $\psi:\mathcal{Y} \to \Ree$ and the corresponding entropy is given as
\[
H(\bP) = \frac{1}{2} \bE_{X,X’\sim \bP}[k(X - X')] - \frac{1}{2} \bE_{X \sim \bP}[\underbrace{k(X-X)}_{\equiv k(0)}],
\]
where the last part is a constant, due to the translation invariance, and therefore does not affect the inequality. Now define $\phi_P(x) \coloneq \bE_{X'\sim \bP} [\psi(x - X')]$, which is convex in x, since $\psi$ is convex and linearity in expectation preserves convexity.

    Now, using convex order, we have
    \[
    \bE_{X \sim \bP_1}[\phi_{\bP_1}(X)] \leq \bE_{Y \sim \bP_2}[\phi_{\bP_1}(Y)].
    \]
    Similarly, we can also obtain an order for the convex function $\phi_{\bP_2}$ as
    \[
    \bE_{X \sim \bP_1}[\phi_{\bP_2}(X)] \leq \bE_{Y \sim \bP_2}[\phi_{\bP_2}(Y)].
    \]
    Now, note that using Fubini's theorem, we obtain
    \[
   \bE_{X \sim \bP_1}[\phi_{\bP_2}(X)] = \bE_{Y \sim \bP_2}[\phi_{\bP_1}(Y)] = \bE_{U\sim \bP_1, V \sim \bP_2}[\psi(U-V)].
    \]
    Therefore, we obtain
     \[
    \bE_{X \sim \bP_1}[\phi_{\bP_1}(X)] \leq \bE_{Y \sim \bP_2}[\phi_{\bP_2}(Y)],
    \]
    and therefore
    \begin{align*}
        \au(\delta_{\bP_1}) =  H(\bP_1) \leq H(\bP_2) = \au(\delta_{\bP_2}).
    \end{align*}
\end{proof}



\begin{proof}[Proof of Proposition \ref{prop:robustness}]
Write $\kappa\coloneq\|k\|_\infty$ and, for $\bP,\bP'\in\mathcal{P}(\mathcal{Y})$, set $a(\bP,\bP')\coloneq\bE_{X\sim\bP,\,X'\sim\bP'}[k(X,X')]$.
The form $a$ is symmetric, satisfies $|a(\bP,\bP')|\le\kappa$, and is linear in each argument under a mixture representation. Since sampling from a mixture means sampling its component first, we have $a\big(\int\bP_{\vtheta}\,dQ(\vtheta),\,\bP'\big)=\int a(\bP_{\vtheta},\bP')\,dQ(\vtheta)$. The entropy and divergence read
\begin{align*}
    H_k(\bP)=\tfrac12 a(\bP,\bP)-\tfrac12\bE_{X\sim\bP}[k(X,X)],
    \qquad
    D_k(\bP,\bQ)= a(\bP,\bQ) - \frac{1}{2} a(\bP,\bP)- \frac{1}{2}a(\bQ,\bQ),
\end{align*}
where $D_k\ge0$ as the divergence of a (strictly) proper score, and $D_k(\bP,\bP)=0$. From $|a|\le\kappa$ we get $|H_k(\bP)|\le\kappa$ and $0\le D_k(\bP,\bP')\le2\kappa$, hence $\au(Q),\eu(Q)<\infty$.

\emph{Aleatoric.} As $\au(Q)=\bE_{\vvtheta\sim Q}[H_k(\bP_{\vvtheta})]$ is linear in $Q$, we have $\au(Q_\varepsilon)=(1-\varepsilon)\au(Q)+\varepsilon H_k(\bP_{\vtheta_0})$, so
\begin{align*}
    \mathrm{IF}(\vtheta_0;\au,Q)=H_k(\bP_{\vtheta_0})-\bE_{\vvtheta\sim Q}[H_k(\bP_{\vvtheta})],
    \qquad
    \sup_{\vtheta_0\in\Theta}|\mathrm{IF}(\vtheta_0;\au,Q)|\le2\kappa.
\end{align*}

\emph{Epistemic, pairwise.} With $\phi(\vtheta,\vtheta')\coloneq D_k(\bP_{\vtheta'},\bP_{\vtheta})$ we have $\eu_{\mathrm{P}}(Q)=\bE_{Q\otimes Q}[\phi]$ and $\phi(\vtheta_0,\vtheta_0)=0$. Substituting $Q_\varepsilon\otimes Q_\varepsilon$,
\begin{align*}
    \eu_{\mathrm{P}}(Q_\varepsilon)=(1-\varepsilon)^2\eu_{\mathrm{P}}(Q)+2\varepsilon(1-\varepsilon)\bE_{\vvtheta\sim Q}[D_k(\bP_{\vtheta_0},\bP_{\vvtheta})],
\end{align*}
so using $0\le \bE_{\vvtheta\sim Q}[D_k(\bP_{\vtheta_0},\bP_{\vvtheta})],\eu_{\mathrm{P}}(Q)\le2\kappa$, we obtain
\begin{align*}
    \mathrm{IF}(\vtheta_0;\eu_{\mathrm{P}},Q)=2\big(\bE_{\vvtheta\sim Q}[D_k(\bP_{\vtheta_0},\bP_{\vvtheta})], -\eu_{\mathrm{P}}(Q)\big),
    \qquad
    \sup_{\vtheta_0\in\Theta}|\mathrm{IF}(\vtheta_0;\eu_{\mathrm{P}},Q)|\le4\kappa.
\end{align*}

\emph{Epistemic, BMA.} Set $A\coloneq\bE_{\vvtheta\sim Q}[a(\bP_{\vvtheta},\bP_{\vvtheta})]$ and $\bar a\coloneq\bE_{\vvtheta,\vvtheta'\sim Q}[a(\bP_{\vvtheta},\bP_{\vvtheta'})]$. Expanding the divergence form,
\begin{align*}
    \eu_{\mathrm{P}}(Q)
    =\bE_{\vvtheta,\vvtheta'}\!\big[a(\bP_{\vvtheta},\bP_{\vvtheta'}) - \frac{1}{2}(\bP_{\vvtheta},\bP_{\vvtheta})-\frac{1}{2}a(\bP_{\vvtheta'},\bP_{\vvtheta'})\big]
    =\bar a - A.
\end{align*}
For the mixture $\overline{\bP}=\int\bP_{\vtheta}\,dQ(\vtheta)$, linearity of $a$ gives $a(\overline{\bP},\bP_{\vvtheta})=\bE_{\vvtheta'}[a(\bP_{\vvtheta'},\bP_{\vvtheta})]$ and $a(\overline{\bP},\overline{\bP})=\bar a$, hence
\begin{align*}
    \eu_{\mathrm{B}}(Q)
    =\bE_{\vvtheta}\big[D_k(\overline{\bP},\bP_{\vvtheta})\big]
    =\bE_{\vvtheta}\big[\bE_{\vvtheta'}[a(\bP_{\vvtheta'},\bP_{\vvtheta})] - \frac{1}{2}\bar - \frac{1}{2}a(\bP_{\vvtheta},\bP_{\vvtheta})\big]
    =\bar a - \frac{1}{2}\bar a+\frac{1}{2}A
    =\frac{1}{2}(A-\bar a)=\tfrac12\eu_{\mathrm{P}}(Q).
\end{align*}
Thus $\mathrm{IF}(\vtheta_0;\eu_{\mathrm{B}},Q)=\tfrac12\,\mathrm{IF}(\vtheta_0;\eu_{\mathrm{P}},Q)$, bounded by $2\kappa$.

In all cases $\sup_{\vtheta_0\in\Theta}|\mathrm{IF}(\vtheta_0;M,Q)|\le C_M\,\kappa<\infty$ for a finite constant $C_M$, so $M$ is robust in terms of the influence function.
\end{proof}

\subsection{Additional propositions for existing measures}
Here, we introduce and prove two more propositions regarding the variance- and entropy-based measures.

\begin{proposition}
    The variance-based measure (squared error) does not fulfill point 1 of Proposition \ref{prop:eu}.
\end{proposition}
\begin{proof}
Consider the BMA estimator, two first-order Gaussian distribution, e.g. $\bP_1 = \mathcal{N}(0, \sigma_1^2), \bP_2 = \mathcal{N}(0, \sigma_2^2)$ with $\sigma_1^2 \neq \sigma_2^2$ and a second-order distribution, specified as a Dirac mixture, i.e. $Q = \frac{1}{2} \delta_{\bP_1} + \frac{1}{2} \delta_{\bP_2}$. Recall that for the variance-based measure, we have $D(\mathbb{P,Q}) = (\bE_{Y \sim \bP}[Y] - \bE_{Y' \sim \bQ}[Y'])^2$. In addition, we obtain $\overline{\bP} = \frac{1}{2} \bP_1 + \frac{1}{2} \bP_2$ and $\bE_{Y' \sim \overline{P}}[Y'] = 0$. Then we obtain
\begin{align*}
    \eu(Q) & = \bE_{\bP \sim Q}[D(\overline{\bP}, \bP)] = \bE_{\bP \sim Q}[ (\underbrace{\bE_{Y' \sim \overline{\bP}}[Y']}_{=0} - \bE_{Y \sim \bP}[Y])^2 ] =\bE_{\bP \sim Q}[( \bE_{Y \sim \bP}[Y])^2 ] \\
           & = \frac{1}{2}\bE_{\bP_1 \sim Q}[(\underbrace{\bE_{Y \sim \bP_1}[Y]}_{=0})^2 ] + \frac{1}{2} \bE_{\bP_2 \sim Q}[( \underbrace{\bE_{Y \sim \bP_2}[Y]}_{=0})^2 ] = 0.
\end{align*}
Therefore, we obtain $\eu(Q) = 0$ although $Q \neq \delta_\bP$. The same argument also works for the pairwise estimator.
\end{proof}

\begin{proposition}
    Assume that $\bP_1, \bP_2$ are absolutely continuous with respect to the Lebesgue measure and therefore admit a probability density. Further assume that both densities are log-concave
    Then, the entropy-based measure (log-score) fulfills $\au(\delta_{\bP_1}) \leq \au(\delta_{\bP_2})$.
\end{proposition}
\begin{proof}
    A probability distribution has log-concave density if the density can be expressed as $p(x) \equiv \exp(\varphi(x))$ for a concave function $\varphi(x)$. Recall that the log-score corresponds to the differential entropy, which can be expressed as
    \begin{align*}
        H(\bP) = - \int p(x) \log  p(x) d\mu(x) = \bE_\bP[-\log p(X)].
    \end{align*}
    Then, for a log-concave density, we have that $\phi(x) \coloneq -\log p_2(x)$ is a convex function in $x$. By convex order, we then have
    \[
    \bE_{X \sim\bP_1}[- \log p_2(X)] = \bE_{X \sim\bP_1}[\phi(X)] \leq \bE_{Y \sim\bP_2}[\phi(Y)] = H(\bP_2).
    \]
    The left-hand side is the cross-entropy of $\bP_1, \bP_2$, which, by definition, can be decomposed into 
    \[
  \bE_{X \sim\bP_1}[- \log p_2(X)] = H(\bP_1) + D_\mathrm{KL}(\bP_1 \| \bP_2) \geq H(\bP_1),
    \]
    where the inequality follows from the KL-divergence being nonnegative. Combining the above gives
    \[
     H(\bP_1) \leq \bE_{X \sim\bP_1}[- \log p_2(X)] = \bE_{X \sim\bP_1}[\phi(X)] \leq \bE_{Y \sim\bP_2}[\phi(Y)] = H(\bP_2),
    \]
    and therefore
    \begin{align*}
        \au(\delta_{\bP_1}) = H(\bP_1)\leq H(\bP_2) = \au(\delta_{\bP_2}).
    \end{align*}
\end{proof}

\section{Derivation of measures for specific choices of scoring rules}
\label{app:derivation_scoring_rules}
In this section, we derive expressions for the (generalized) entropy- and divergence term of the uncertainty measures introduced in this article. Recall that in order to assess EU, AU and TU, one requires expressions for the entropy, divergence and expected scoring rule. This is regardless whether one chooses the pairwise or the BMA estimator. Therefore, for $\bP, \bQ \in \cP(\cY)$ and $X,X' \sim \bP, \ Y,Y' \sim \bQ$, and $\bP, \bP' \sim Q, \ \overline{\bP} = \bE_Q[\bP]$, we will derive the quantities $H(\bP), D(\bP, \bQ)$, as well as the gap between the BMA and pairwise estimation $\Delta$, for different scoring rules.

\paragraph{Log-score}

    Let $\mathcal{P}$ be the set of distributions on $\mathcal{Y}$ that are absolutely continuous with respect to the Lebesgue measure $\mu$ and $\bP, \bQ \in \mathcal{P}$ with corresponding densities $p,q$. The \emph{logarithmic score} $S_{\mathrm{log}}: \mathcal{P} \times \mathcal{Y} \to \overline{\Ree}$, given by
    \begin{equation*}
        \label{eq:log_sr}
        S_{\mathrm{log}}(\bP,\boldsymbol{y}) = - \log p(\boldsymbol{y})
    \end{equation*}
    is a strictly proper scoring rule. The associated entropy and divergence are given as
    \begin{align*}
        H_{\mathrm{log}}(\bP)          & = - \int p(\boldsymbol{x}) \log p(\boldsymbol{x}) \, d\mu(\boldsymbol{x}),                                                                       \\
        D_{\mathrm{log}}(\mathbb{P,Q}) & = \int q(\boldsymbol{y}) \log \left(\frac{q(\boldsymbol{y})}{p(\boldsymbol{y})}\right) \, d\mu(\boldsymbol{y}) = D_\mathrm{KL}(\mathbb{Q}\|\bP),
    \end{align*}
    which are the Shannon entropy and Kullback-Leibler divergence, respectively. Utilizing the BMA estimator, we obtain the entropy-based measure, while for the pairwise estimator we obtain the pairwise KL-divergence, as shown by \cite{schweighofer2023introducingimprovedinformationtheoreticmeasure}. For their difference, we obtain the so-called reverse mutual information
    \[
    \Delta = \bE_{Q}\left[D_{\mathrm{KL}}\left(\overline{\bP} \| \bP \right)\right].
    \]

\paragraph{Kernel score}
Consider the kernel score $S_k: \mathcal{P}_k \times \cY $ associated with a negative definite kernel $k$. We obtain the following expressions for the pairwise estimator:
\begin{align*}
        H(\bP) &= \frac{1}{2} \bE_{\bP}\left[k(X,X') \right]- \frac{1}{2} \bE_\bP [k(X,X)], \\ 
        D(\mathbb{P,Q}) &= \bE_{\bP, \bQ}\left[k(X,Y)\right] -\frac{1}{2} \bE_{\bP}\left[k(X,X') \right] -\frac{1}{2} \bE_{\bQ}\left[k(Y,Y') \right].
\end{align*}
The corresponding uncertainty measures are obtained by plugging the selected kernel into the above quantities.

\paragraph{Squared error}
Let $\mathcal{P}$ be the set of distributions on $\mathcal{Y} \subseteq \Ree^p$ such that $\int \|\boldsymbol{x}\|^2 \, d\bP(\boldsymbol{x})<\infty$ and $ Y \sim \bP \in \cP(\cY)$. The squared error $S_{\mathrm{SE}}:\mathcal{P}\times \mathcal{Y} \to \overline{\Ree}$ given by
\[
    S_{\mathrm{SE}}(\bP,\boldsymbol{y}) = (\boldsymbol{y} - \bE_\bP[ Y])^2,
\]
is a proper (but not strictly proper) kernel rule, with $k(\bm x, \bm x') = \|\bm x- \bm x' \|^2$.
The associated entropy and divergence are given as
\begin{align*}
    H_{\mathrm{SE}}(\bP) = \tr(\mathrm{Cov}_{\bP}[Y]), \qquad  
    D_{\mathrm{SE}}(\mathbb{P,Q}) = \left\|\bm \mu_\bP  - \bm \mu_\bQ \right\|^2.
\end{align*}
In the case of the squared error, the corresponding uncertainty measures can be expressed in terms of moments of the first-order distribution, leading to the following measures for the BMA estimator:
\begin{align*}
    \au_B(Q) &= \bE_Q\left[ \tr(\mathrm{Cov}_{\bP}[Y])\right], \\
    \eu_B(Q) &= \bE_Q\left[\left\|\bm \mu_\bP  - \bm \mu_{\bP'} \right\|^2\right] = \tr \left(\mathrm{Cov}_Q[\bm \mu_\bP] \right), \\
    \tu_B(Q) &= \bE_Q \left[\| Y - \bE_Q[\bm \mu_\bP] \|^2 \right],
\end{align*}
which reduces to the variance-based decomposition in the univariate case $\cY \subseteq \Ree$. For the pairwise estimator, we obtain
\begin{align*}
    \au_P(Q) &= \bE_Q\left[ \tr(\mathrm{Cov}_{\bP}[Y])\right], \\
    \eu_P(Q) &= 2\bE_Q\left[\left\|\bm \mu_\bP  - \bm \mu_{\bP'} \right\|^2\right] = 2\tr \left(\mathrm{Cov}_Q[\bm \mu_\bP] \right), \\
    \tu_P(Q) &= \bE_Q \left[\| Y - \bE_Q[\bm \mu_\bP] \|^2 \right] + \tr \left(\mathrm{Cov}_Q[\bm \mu_\bP] \right),
\end{align*}
which shows that both estimators only differ by a factor of two for the epistemic uncertainty. The gap between both estimators is
\[
\Delta = \tr \left(\mathrm{Cov}_Q[\bm \mu_\bP] \right) = \bE_Q[ D_{\mathrm{SE}}(\overline{\bP},\bP)].
\]
This quantity measures the expected (score-) divergence between the BMA against all possible models.

\subsection{Closed-form expressions for Gaussians}
\label{app:closed_forms}
Here, we derive closed-form expressions for the entropy and divergence term of different scoring rules for first-order (univariate) Gaussian and mixture of Gaussian distributions. Recall that for kernel scores $S_k$ with a conditionally negative definite kernel $k$, the entropy and divergence of two probability measures $\bP, \bQ \in \cP(\cY)$ are given as
\begin{align}
    H_k(\bP)      & = \frac{1}{2}\bE_{X, X' \sim \bP}[k(X,X')] - \frac{1}{2}\bE_{X \sim \bP}[k(X,X)]                                               \\
    D_k(\bP, \bQ) & = \bE_{X \sim \bP, Y \sim \bQ}[k(X,Y)] - \frac{1}{2} \bE_{X, X' \sim \bP}[k(X,X')] - \frac{1}{2}\bE_{Y, Y' \sim \bQ}[k(Y,Y')].
\end{align}

Consider two first-order Gaussian distributions $X \sim \bP = \mathcal{N}(\mu, \sigma^2), \ Y \sim \bQ = \mathcal{N}(\nu, \tau^2)$. Then we obtain the following expressions:

\paragraph{Log-score}
\begin{align}
    H(\bP)     & = \frac{1}{2}\log(2\pi e \sigma^2),                                                             \\
    D(\bP,\bQ) & = \log \left(\frac{\sigma}{\tau} \right) + \frac{\tau^2 +(\mu - \mu)^2}{2\sigma^2}-\frac{1}{2}.
\end{align}
These expressions are obtained via well-known results from the differential entropy and KL-divergence for Gaussian distributions (compare for example \cite{10.7551/mitpress/3206.001.0001}).

\paragraph{Squared error}
\begin{align}
    H(\bP)     & = \sigma^2,      \\
    D(\bP,\bQ) & = (\mu - \nu)^2.
\end{align}
\begin{proof}
    For the entropy, we obtain
    \begin{align*}
        H(\bP) = \frac{1}{2}\bE_{X, X' \sim \bP}[(X-X')^2)] = \frac{1}{2}\left( \bE_\bP[X^2] - 2 \bE_\bP[X]\bE_\bP[X'] +\bE_\bP[X'^2] \right) = \bV_\bP[X] = \sigma^2.
    \end{align*}
    In addition, we have that $\bE_{X \sim \bP, Y \sim \bQ}[(X-Y)^2] = \bE_\bP[X^2] - 2 \bE_\bP[X]\bE_\bQ[Y] +\bE_\bQ[Y^2]$ such that for the divergence we obtain
    \begin{align*}
        D(\bP, \bQ) & = \bE_\bP[X^2] - 2 \bE_\bP[X]\bE_\bQ[Y] +\bE_\bQ[Y^2] -  \bV_\bP[X] - \bV_\bQ[Y]                                 \\
                    & = \bE_\bP[X^2] - 2 \bE_\bP[X]\bE_\bQ[Y] +\bE_\bQ[Y^2] - \bE_\bP[X^2] + \bE_\bP[X]^2- \bE_\bQ[Y^2] + \bE_\bQ[Y]^2 \\
                    & = \bE_\bP[X]^2 -2 \bE_\bP[X]\bE_\bQ[Y]+ \bE_\bQ[Y]^2= \left(\bE_\bP[X] - \bE_\bQ[Y] \right)^2                    \\
                    & =(\mu - \nu)^2.
    \end{align*}
\end{proof}

\paragraph{CRPS}
\begin{align}
    H(\bP)     & = \frac{\sigma}{\sqrt{\pi}},                                                                                                                                                                                   \\
    D(\bP,\bQ) & = \left(\sqrt{\sigma^2 + \tau^2}\right)\frac{\sqrt{2}}{\sqrt{\pi}} {}_{1}F_1 \left(-\frac{1}{2}, \frac{1}{2}; -\frac{1}{2} \frac{(\mu - \nu)^2}{\sigma^2 + \tau^2} \right) - \left(\frac{\sigma+\tau}{\sqrt{\pi}}\right).
\end{align}
\begin{proof}
    \cite{winkelbauer2014momentsabsolutemomentsnormal} show that for the raw absolute moment of a Gaussian we have
    \[
        \bE[|X|^p] = \sigma^p 2^{p/2} \frac{\Gamma(\frac{p+1}{2})}{\sqrt{\pi}}{}_{1}F_1 \left(-\frac{p}{2},\frac{1}{2}; - \frac{\mu^2}{2\sigma^2} \right),
    \]
    where ${}_{1}F_1$ denotes Kummer's confluent hypergeometric function. Furthermore, we know that ${X-Y \sim \mathcal{N}(\mu - \nu, \sigma^2+\tau^2)},\ X-X'\sim \mathcal{N}(0, 2\sigma^2)$ and $Y-Y' \sim \mathcal{N}(0, 2\tau^2)$. Therefore, we obtain
    \begin{align*}
        H(\bP) = \frac{1}{2} \bE_{X,X' \sim \bP}[|X-X'|] = \frac{1}{2} \sqrt{2\sigma^2} \sqrt{2} \frac{\Gamma(1)}{\sqrt{\pi}} {}_{1}F_1\left( -\frac{1}{2}, \frac{1}{2}; 0 \right) = \frac{\sigma}{\sqrt{\pi}}.
    \end{align*}
    With $\bE_{X \sim \bP, Y \sim \bQ}[|X-Y|] = \sqrt{\sigma^2 + \tau^2} \frac{\sqrt{2}}{\sqrt{\pi}} {}_{1}F_1 \left(-\frac{1}{2}, \frac{1}{2}; -\frac{1}{2} \frac{(\mu - \nu)^2}{\sigma^2 + \tau^2} \right)$ we obtain the divergence $D(\bP, \bQ)$ by plugging in the corresponding expectations.
\end{proof}

\paragraph{Gaussian kernel score}
Given the (negative) Gaussian kernel $k(x,y) = -\exp(-(x-y)^2/ \gamma^2)$ with scalar bandwidth $\gamma$, we obtain
\begin{align}
    H(\bP)     & = \frac{1}{2}\left( 1 - \frac{\gamma}{\sqrt{\gamma^2+4\sigma^2}}\right)\\
    D(\bP,\bQ) & = \frac{1}{2} \frac{\gamma}{\sqrt{\gamma^2+4\sigma^2}} + \frac{1}{2} \frac{\gamma}{\sqrt{\gamma^2+4\tau^2}} - \frac{\gamma}{\sqrt{\gamma^2+2(\sigma^2+\tau^2)}} \exp \left( -\frac{(\mu - \nu)^2}{\gamma^2 + 2(\sigma^2+\tau^2)} \right)
\end{align}
\begin{proof}
    Let $Z \coloneq X - Y \sim \mathbb{P}_Z \coloneq \mathcal{N}(\delta, \upsilon)$ with $\delta \coloneq \mu - \nu, \upsilon \coloneq \sigma^2 +  \tau^2$. Then $\frac{Z^2}{\delta}$ follows a noncentral chi-squared distribution, i.e. $\frac{Z^2}{\delta} \sim \chi^2(1, \lambda)$ with noncentrality parameter $\lambda = \frac{\delta^2}{\upsilon}$. Furthermore, we have
    \begin{align*}
        \bE_{X \sim \bP, Y \sim \bQ}[k(X,Y)] & = -\bE_{\bP_Z}\left[\exp\left(- \frac{\frac{Z^2}{\upsilon}\upsilon}{\gamma^2} \right)\right] = - M_{\chi^2(1,\lambda)}\left(-\frac{\upsilon}{\gamma^2}\right).
    \end{align*}

    Here, $M_{\chi^2(k,\lambda)}(t)$ is the moment-generating function of $\chi^2(k, \lambda)$, with $t = -\frac{\upsilon}{\gamma^2}$, which can be expressed analytically (compare, for example, \cite{patnaik}) as $M_{\chi^2(k,\lambda)}(t) = \frac{\exp \left( \frac{\lambda t}{1-2t} \right)}{(1-2t)^{k/2}}$. Therefore, we obtain
    \begin{align*}
        \bE_{X \sim \bP, Y \sim \bQ}[k(X,Y)] & = - \frac{\gamma}{\sqrt{\gamma^2 + 2 (\sigma^2+\tau^2)}} \exp \left(- \frac{(\mu - \nu)^2}{\gamma^2 + 2(\sigma^2 + \tau^2)} \right)
    \end{align*}
    and
    \begin{align*}
        H(\bP) & =  \frac{1}{2}\bE_{X, X' \sim \bP}[k(X,X')] - \frac{1}{2}\bE_{X \sim \bP}[k(X,X)] \\
               & = \frac{1}{2}\left( 1 - \frac{\gamma}{\sqrt{\gamma^2 + 4\sigma^2}}\right).
    \end{align*}
    By plugging these expressions into the definition of the divergence $D(\bP, \bQ)$, we obtain the corresponding closed form.
\end{proof}

\paragraph{Gaussian mixtures}
Here, we consider a mixture of Gaussians, i.e. $X \sim \bP = \sum_{i=1}^M w_i \mathcal{N}(\mu_i, \sigma_i^2), Y \sim \bQ = \sum_{j=1}^N v_j \mathcal{N}(\mu_j, \sigma_j^2)$ with nonnegative weights $w_i, v_j$ that sum to one. For a mixture of Gaussians, closed-form expressions are not necessarily available, as is the case for the log-score. However, for specific cases, closed-form expressions are available via the corresponding marginals. For a translation-invariant kernel score, the expressions for the mixture density network can be derived in terms of the kernel score of the individual components. By linearity of the expectation, we obtain
\begin{align*}
    \bE[k(X,Y)] = \sum_{i=1}^M \sum_{j=1}^N w_i v_j \bE_{X\sim \mathcal{N}(\mu_i,\sigma_i^2), Y\sim \mathcal{N}(\mu_j, \sigma_j^2)}[k(X,Y)].
\end{align*}
In the case of a translation invariant kernel, i.e. $k(X,Y) \equiv k(X-Y)$, this reduces to a weighted sum of the corresponding Gaussian score, as we have $X-Y \sim \mathcal{N}(\mu_i - \mu_j, \sigma_i^2 + \sigma_j^2)$. Therefore, we can use the results from the previous section to derive the scores for the Gaussian mixtures analytically.

\paragraph{Marginal scores}
In the multivariate setting $\mathcal{Y} \subseteq \mathbb{R}^d$ for $d>1$, closed-form expressions are more difficult to obtain than in the univariate setting. For instance, for a Gaussian distribution, the energy score admits an analytic solution for $d=1$ but not for $d>1$. However, one can always define a multivariate proper scoring rule from a univariate one. Let $\{ Y_j \}_{j=1}^d$ be a collection of marginal distributions from the multivariate random variable $\boldsymbol{Y}$. Then one can construct a \emph{marginal score} for $\boldsymbol{Y}$ as
\begin{equation*}
    S_M(\bP, \by) =  \sum_{j=1}^d S(\bP_j, y_j),
\end{equation*}
where $Y_j \sim \bP_j$ when $\boldsymbol{Y} \sim \bP$ and $S$ is a (strictly) proper scoring rule for the marginal $Y_j$ \citep{dombry}. Then, the scoring rule $S_M$ is proper, but not strictly proper. This is especially interesting if the main interest is in the marginals, for example, if the dependence structure across the marginals is of little interest.

\section{Computational complexity and approximation error}
\label{app:computational_complexity}

\paragraph{Computational complexity.}
The cost of the pairwise estimator separates into a second-order level over the
$M$ ensemble members and a first-order level for evaluating the divergence/entropy
per member. The second-order level requires $\mathcal{O}(M^2)$ divergence evaluations for
EU and $\mathcal{O}(M)$ entropy evaluations for AU. This is shared by \emph{all} pairwise measures, including $S_\mathrm{log}$ and $S_\mathrm{SE}$. The first-order cost
depends on the scoring rule and the uncertainty representation; if it is available in closed-form, the cost is negligible, otherwise the cost depends on the additional number of first-order samples $N$ (compare \autoref{tab:complexity}):
\begin{itemize}
  \item $S_\mathrm{SE}$ is cheapest at $\mathcal{O}(MNd)$ and scales linearly in $M$, $N$,
        and $d$: since $\mathrm{EU}_\mathrm{SE} = \tr(\mathrm{Cov}_Q[\bm{\mu}_\mathbb{P}])$,
        it suffices to compute $M$ empirical means and the trace of their covariance, with
        no pairwise sample comparisons. This efficiency comes at the price of discarding all
        distributional information beyond the first moment.
  \item With the closed-form expressions of NG, DER, MDN, LoRa, the kernel scores attain $\mathcal{O}(M^2 d)$, adding negligible overhead for typical
        ensemble sizes ($M \leq 10$).
  \item Without closed forms, sample-based kernel scores cost $\mathcal{O}(M^2 N^2 d)$, i.e.\
        $\mathcal{O}(N^2)$ per pair, but require no density evaluation.
  \item $S_\mathrm{log}$ requires a density: sample-based evaluation therefore falls back to kernel density estimation (KDE), and the BMA log-score admits no closed form when $\bar{\mathbb{P}}$ is a mixture, whereas kernel scores decompose linearly over mixture components and remain exact.
\end{itemize}

\begin{table}[h]
  \centering
  \caption{First-order computational cost and estimation error of the pairwise estimator
  per scoring rule. The second-order level adds a shared $\mathcal{O}(M^2)$ (EU) /
  $\mathcal{O}(M)$ (AU) term. $M$: ensemble size, $N$: first-order samples, $d$: output
  dimension. The error column refers to sample-based evaluation.}
  \label{tab:complexity}
  \begin{tabular}{lccc}
    \toprule
    Measure & Closed-form cost & Sample-based cost & First-order error \\
    \midrule
    $S_\mathrm{SE}$                 & $\mathcal{O}(MNd)$ & $\mathcal{O}(MNd)$    & $\mathcal{O}(1/\sqrt{N})$ \\
    $S_\mathrm{ES},\, S_{k_\gamma}$ & $\mathcal{O}(M^2 d)$ & $\mathcal{O}(M^2 N^2 d)$ & $\mathcal{O}(1/\sqrt{N})$ \\
    $S_\mathrm{CRPS}$               & $\mathcal{O}(M^2 d)$ & $\mathcal{O}(M^2 N^2 d)$ & $\mathcal{O}(1/\sqrt{N})$ \\
    $S_\mathrm{log}$                & --- (no closed form for mixtures) & KDE & $\mathcal{O}(N^{-4/(d+4)})$ \\
    \bottomrule
  \end{tabular}
\end{table}

\begin{figure}[h]
  \centering
  \includegraphics[width=\linewidth]{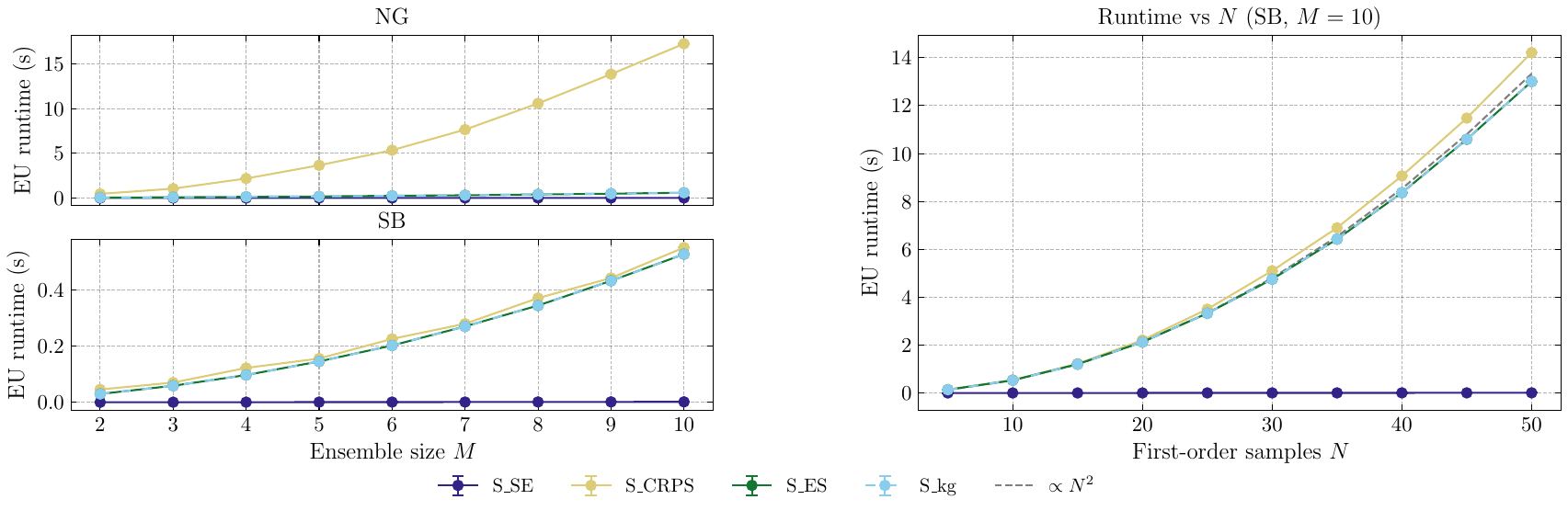}
  \caption{Wall-clock runtime of the uncertainty measures on NYU Depth v2 (100 images).
  Left: varying ensemble size $M$. Right: varying sample count $N$ (sample-based, $M = 10$).}
  \label{fig:runtime}
\end{figure}

\paragraph{Empirical runtime.}
We validate these costs on NYU Depth v2 (100 test images, $d = 55 \times 74$;
\autoref{fig:runtime}). Varying $M \in \{2,\dots,10\}$ with closed-form (NG) and sample-based
(SB, $N = 10$) representations, all kernel measures evaluate in well under one second at
$M = 10$, while $S_\mathrm{SE}$ is near-constant owing to its linear scaling. Varying
$N \in \{5,\dots,50\}$ at fixed $M = 10$, the sample-based kernel measures follow the
$\mathcal{O}(N^2)$ reference curve and $S_\mathrm{SE}$ stays flat. (The marginal
$S_\mathrm{CRPS}$ appears slower under the closed form only because its ${}_1F_1$
evaluation falls back to a CPU \texttt{scipy} routine.) At our 2D experimental setting ($N = 10$), all runtimes are well below one second
per 100 images.

\paragraph{Approximation error.}
Two error sources arise: finite ensemble size $M$ (second-order) and, for sample-based
evaluation, finite sample count $N$ (first-order). The second-order estimators both converge at $\mathcal{O}(1/\sqrt{M})$; AU, because it is estimated as a sample mean and EU, as it is estimated via a one-sample U-statistic of order two. Crucially, this rate is identical for every scoring rule (including $S_\mathrm{log}$ and $S_\mathrm{SE}$): the finite-ensemble error is a property of the uncertainty representation, not of the measure. For the first-order error, if a closed-form expression is available, the estimation error reduces to zero.
When sampling is required, the U-statistic estimator of the kernel divergence $D_k$ is unbiased and converges at $\mathcal{O}(1/\sqrt{N})$ independently of the output dimension $d$ \citep{grettonKernelTwosampleTest2012}, in contrast to KDE-based evaluation of $S_\mathrm{log}$, which converges at $\mathcal{O}(N^{-4/(d+4)})$ and is impractical for large $d$. The total estimation error of the sample-based estimator thus decomposes as
\begin{equation*}
  \mathcal{O}\!\left(1/\sqrt{M}\right) + \mathcal{O}\!\left(1/\sqrt{N}\right),
\end{equation*}
with the second term vanishing whenever closed-form first-order expressions are available.

\section{Experiment details}
\label{app:experiment_details}
Our model implementations and reproducible experiments are available at \url{https://github.com/cbuelt/kernel-uq}.

\subsection{Datasets}
\label{app:datasets}
In this section, we describe all the datasets used, as well as their generation procedure and out-of-distribution version, if applicable. All datasets and corresponding tasks are some sort of regression problem, where the quality of a prediction is evaluated using the mean-squared-error (MSE). An overview of the datasets is available in \autoref{tab:datasets}.

\begin{table}[h]
    \centering
    \caption{Overview of datasets and evaluation protocols used in experiments.}
    \label{tab:datasets}
    \begin{tabular}{llll}
        \toprule
        \textbf{Task} & \textbf{Domain} & \textbf{Datasets} & \textbf{OOD shift} \\
        \midrule
        Univariate & Tabular regression  & UCI   & - \\
        \midrule
        \multirow{2}{*}{1D PDEs} &   PDE& Burgers' & Change in viscosity $\nu$ \\
        &   PDE&Kuramoto-Sivashinsky& Change in length scale $L$\\
        \midrule
        \multirow{2}{*}{2D}&  Climate  & ERA5  & Geographic domain shift  \\
        & Computer vision & NYU Depth / ApolloScape & Scene shift \\
        \bottomrule
    \end{tabular}
\end{table}

\paragraph{Univariate regression}
For the univariate regression task, we utilize the UCI benchmark \citep{hernándezlobato2015probabilisticbackpropagationscalablelearning}, from which we use all datasets, except \emph{boston}, due to raised ethical concerns\footnote{\url{https://medium.com/@docintangible/racist-data-destruction-113e3eff54a8}}, and \emph{wine-quality} due to its categorical prediction objective. 

\begin{figure}[ht]
  \includegraphics[width = \linewidth]{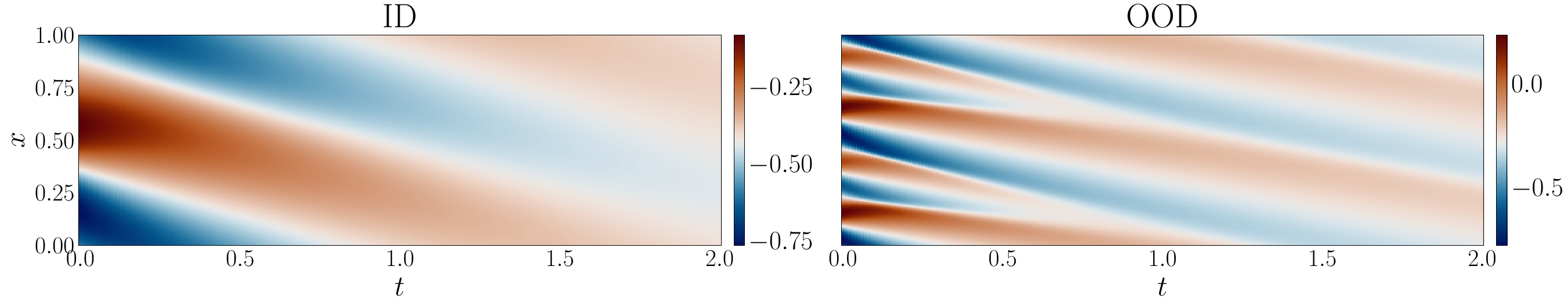} 
  \caption{Selected samples of the in-distribution ($\nu = 0.1$) and out-of-distribution dataset ($\nu = 0.01$) for the Burgers' equation.}
  \label{fig:burgers_data}
\end{figure}

\paragraph{1D PDE tasks}
To accommodate more complex tasks, we use two time-dependent one-dimensional partial differential equations (PDEs). While both PDEs could be analyzed in an autoregressive manner, we focus on a single-step prediction, where at each step the probabilistic model samples from a conditional distribution $~{u_s \sim f_\theta (\cdot \mid u_{s-1}, u_{s-2})}$,
predicting the dynamics $u_s - u_{s-1}$, given the last two timesteps. In particular, we consider the following two dynamical systems:

\emph{Burgers' equation.}
The Burgers' equation is given as
\begin{equation*}
\begin{split}
    \partial_t u(t,x) + \partial_x (u^2(t,x) /2) &= \nu / \pi \partial_{xx}u(t,x), \quad x \in (0,1), \ t\in (0,2]\\
    u(0,x) &= u_0(x), \quad x \in (0,1)
\end{split}    
\end{equation*}
where $u \in C([0,T]; H_{\mathrm{per}}^r((0,1); \Ree))$ for any $r>0$, $u_0 \in L_{\mathrm{per}}^2((0,1); \Ree)$ is the initial condition and $\nu \in \Ree_+$ is the diffusion coefficient\footnote{$H^r_{\text{per}}(\mathcal{D}; \mathbb{R}), L^2_{\text{per}}(\mathcal{D}; \mathbb{R})$ denote the periodic Sobolev and $L^2$ spaces, respectively.}. We utilize data from the PDEBENCH repository \citep{NEURIPS2022_0a974713}, which assumes a constant diffusion coefficient, which we choose as $\nu = 0.1$. The data is generated with a periodic boundary condition from a superposition of sinusoidal waves with the temporally and spatially 2nd-order upwind difference scheme for the advection term, and the central difference scheme for the diffusion term. 
As an out-of-distribution dataset, we use the same equation but with diffusion coefficient $\nu = 0.01$, which leads to the corresponding solutions being less smooth and having steeper gradients. \autoref{fig:burgers_data} shows a sample rollout of the two different systems.

\begin{figure}[ht]
  \includegraphics[width = \linewidth]{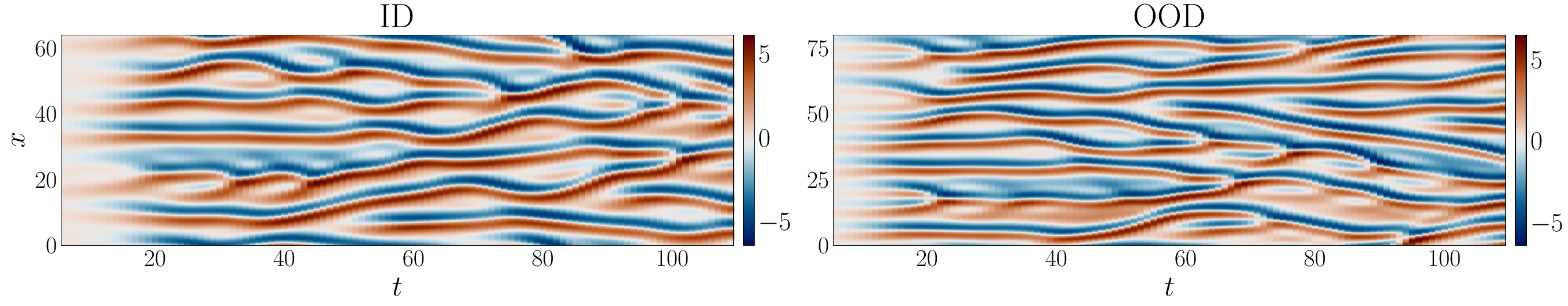} 
  \caption{Selected samples of the in-distribution ($L=64)$ and out-of-distribution dataset ($L=80$) for the Kuramoto-Sivashinsky equation.}
  \label{fig:ks_data}
\end{figure}

\emph{Kuramoto--Sivashinsky equation.}
The Kuramoto--Sivashinsky (KS-) equation in one spatial dimension is given as:
\begin{alignat*}{2}  
\partial_t u(x,t)+ u \partial_x u(x,t) + \partial_x^2 u(x,t) + \partial_x^4 u(x,t) &= 0, \qquad &&x \in \mathcal{D}, t \in (0,T]\\
u(x,0) &= u_0(x), \qquad &&x \in \mathcal{D}
\end{alignat*}
where $\mathcal{D} \subseteq \mathbb{R}$, $u \in C([0,T]; H_{\mathrm{per}}^4(\mathcal{D}; \mathbb{R}))$, and $u_0 \in L_{\mathrm{per}}^2(\mathcal{D};\mathbb{R})$. We follow the setup in \citet{bultepno} and simulate the KS-equation from random uniform noise $\mathcal{U}(-1,1)$ on a periodic domain $\mathcal{D} = [0,L], \ L=64$ using the py-pde package \citep{zwicker_py-pde_2020}. We generate 10000 samples with a resolution of $256 \times 50$ and $\Delta t = 2$.\\
As an out-of-distribution dataset, we again simulate from the KS-equation, but with an adjusted length scale of $L=80$, which leads to an increasing number of unstable spatial modes. \autoref{fig:ks_data} shows a sample rollout of the two different systems.

\begin{figure}[ht]
  \includegraphics[width = \linewidth]{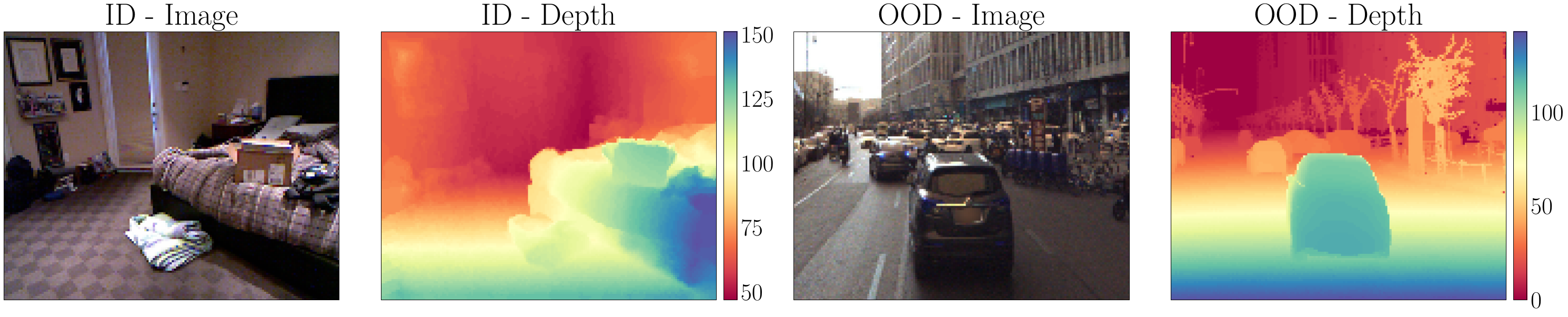}
  \caption{Selected samples of the NYU (in-distribution) and ApolloScape (out-of-distribution) dataset with corresponding ground truth image and depth target.}
  \label{fig:depth_data}
\end{figure}

\paragraph{2D tasks}
Finally, we use the following two high-dimensional imaging tasks:

\emph{Depth regression.}
As a typical vision task, we utilize the NYU Depth v2 dataset \citep{nyu}, similar to \cite{aminiDeepEvidentialRegression2020}, which consists of image-depth pairs of indoor scenes. As an out-of-distribution dataset, we utilize ApolloScape \citep{apollo}, a dataset of outdoor driving scenes. \autoref{fig:depth_data} shows the depth-image pairs for the two different datasets.

\begin{figure}[ht]
\centering
  \includegraphics[width = 0.75\linewidth]{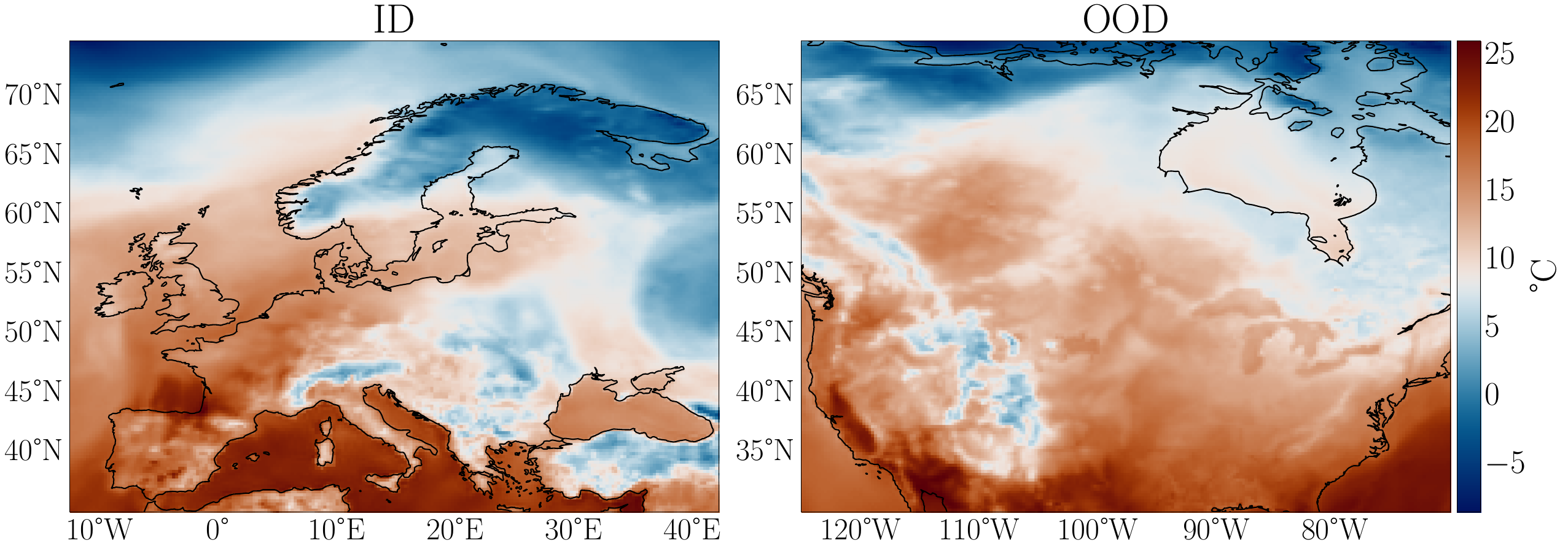}
  \caption{Selected sample of the two-meter surface temperature (28 October 2022, 00:00 UTC) for the in-distribution and out-of-distribution domain.}
  \label{fig:t2m_data}
\end{figure}

\emph{Surface temperature prediction (T2M).}
Finally, we use a grid-based surface temperature prediction task, where we utilize the ERA5 dataset \citep{hersbachERA5GlobalReanalysis2020} provided via the WeatherBench2 benchmark \citep{rasp2024weatherbench2benchmarkgeneration}.
We fix a 6-hour forecast horizon and use training data from 2011 to 2018, validation data from 2019 to 2020, and test data from 2022.
Similar to \citet{bultepno}, we use data with a spatial resolution of $0.25^\circ \times 0.25^\circ$ and a time resolution of $6h$. For computational reasons, we restrict the data to a European domain, covering an area from 35°N – 75°N and 12.5°W – 42.5°E with selected user-relevant weather variables (u-component and v-component of 10-m wind speed (U10 and V10), temperature at 2m and 850 hPa (T2M and T850), geopotential height at 500 hPa (Z500), as well as land-sea mask and orography) that serve as input to the model, while the prediction target is only T2M. The total number of input channels is 12. 
As an out-of-distribution dataset, we use a completely different domain that is still roughly similar to the in-distribution data regarding topography and climate regime, namely a domain over North America with similar latitudes.  In particular, we use a domain of the same size and spatial resolution, but ranging from 30°N - 69.75°N and 125°W - 70°E. 
Due to the size of the data and corresponding computational restraints, the resolution of the dataset is halved for the evaluation of the uncertainty measures in the selective prediction and OOD experiments.
\autoref{fig:t2m_data} shows a sample of the two-meter surface temperature for the two different domains.

\subsection{Backbone models}
\label{app:backbone_models}
For all models, we use the Adam optimizer \citep{kingma2017adammethodstochasticoptimization} with early stopping after 10 epochs and a learning rate schedule that halves the learning rate if no improvement in validation loss has been recorded for more than 5 epochs. For all datasets we use a 10\% train-test split and an additional 10\% split into training and validation.

\paragraph{Univariate regression}
As a model backbone, we use a MLP with two hidden layers with 50 neurons each and GELU activation function, where the final activation depends on the chosen uncertainty representation method. Here we use a learning rate of 1e-4, an early stopping patience of 25, and train for a maximum number of 5000 epochs. The batch size is chosen as 32, 64, or 128, depending on the size of the dataset.

\paragraph{1D PDE tasks}
For the two PDE tasks, we use a Fourier neural operator \citep{li2021fourierneuraloperatorparametric}, which is a neural network architecture that directly learns solutions in the corresponding function space and has shown great success in modeling and solving partial differential equations. Our implementation is based on the publicly available \emph{neuraloperator} library\footnote{\url{https://github.com/neuraloperator/neuraloperator}}. The models for both tasks are specified identically, namely with 20 Fourier modes, 64 hidden channels, and 256 projection and lifting channels. In total, both networks have roughly 463k parameters. The neural operators are trained with a learning rate of 1e-3, an early stopping patience of 10, and a batch size of 128 for a maximum of 500 epochs.

\paragraph{2D tasks}
For both two-dimensional tasks, we use a ConvNeXt architecture \citep{liu2022convnet}, which is based on a simple ResNet, but adapted to be more similar to the architecture of a Vision Transformer. It has shown comparable performance to different Transformer versions across a variety of tasks, while maintaining the simplicity and efficiency of regular convolutional neural networks. In particular, we can make use of the pre-trained version, available via PyTorch\footnote{\url{https://docs.pytorch.org/vision/main/models/convnext.html}}, where we use the tiny variant for the depth regression and the small variant for the temperature prediction task.
The models are pre-trained on ImageNet1K, and we only adapt the very first and last layer to accommodate for the different number of input and output channels. In total, the models have 34M and 56M parameters for the depth regression and temperature prediction tasks, respectively.
We use a learning rate of 1e-3, an early stopping patience of 10, and a batch size of 64 trained for a maximum of 250 epochs.

\subsection{Uncertainty representation methods}
\label{app:uncertainty_representation}
In this section, we describe the different uncertainty representation methods used in our experiments. Except for deep evidential regression, which directly learns a second-order distribution $Q$, all methods are first-order predictors and the corresponding second-order distribution is created via ensembling \citep{lakshminarayananSimpleScalablePredictive2017} with $M=10$ ensembles. The different uncertainty representation methods can be applied to any architecture; they just require an adjustment in the final layer processing. An exception is the neural operator, since the corresponding output is in function space, where probability densities are not properly defined; only the sampling-based approach is theoretically valid \citep{bultepno}. For the remaining methods, we share the model-specific parameters across the different backbones. For the multivariate tasks, the univariate methods (Deep evidential regression, natural Gaussian, and mixture density network) are interpreted as pointwise predictions.

\begin{figure}[ht]
    \centering
    \begin{subfigure}{0.8\textwidth}
        \centering
        \caption{NYU}
        \includegraphics[width = \linewidth]{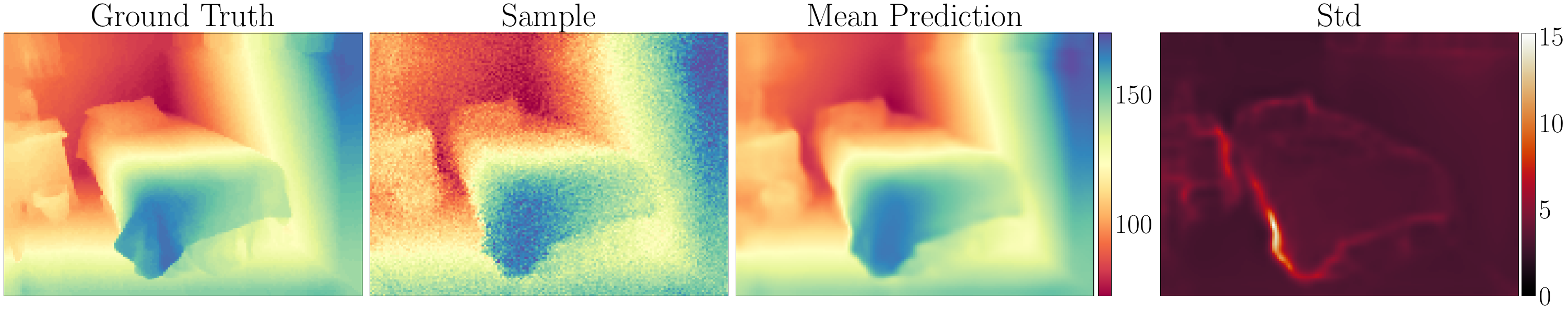}
    \end{subfigure}
    \begin{subfigure}{0.8\textwidth}
        \centering
        \caption{T2M}
        \includegraphics[width = \linewidth]{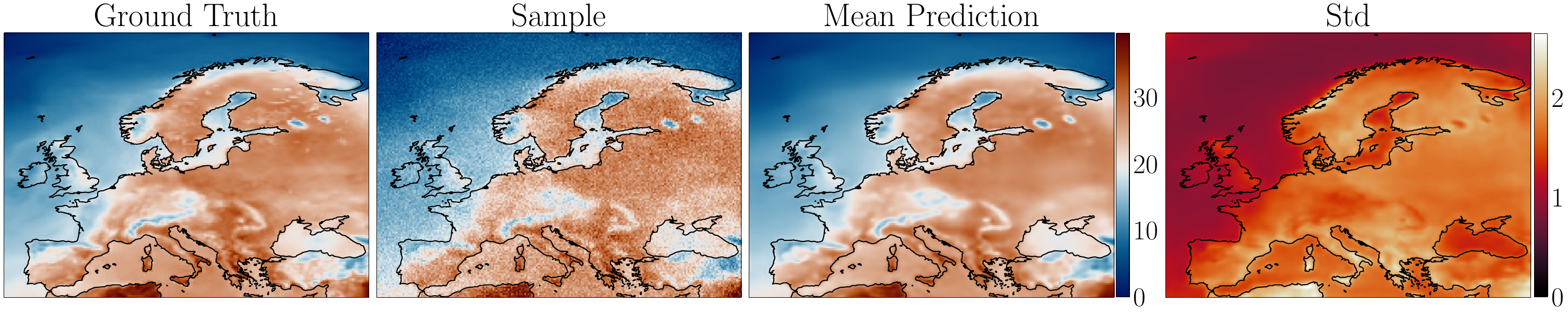}
    \end{subfigure}    
    \caption{Visualizations of the predictions of the natural Gaussian method for the depth regression and surface temperature prediction tasks.}
    \label{fig:ur_normal}
\end{figure}

\paragraph{Natural Gaussian}
Using a predictive normal distribution $p(\cdot \mid \vtheta(\boldsymbol{x})) = \mathcal{N}(\mu(\bx), \sigma^2(\bx))$ to approximate the probability distribution over $\by$ given $\bx$ for a given model has been common practice in many machine learning tasks \citep{374138,lakshminarayananSimpleScalablePredictive2017}. However, when training with the log-likelihood, direct optimization of $\mu$ and $\sigma^2$ usually leads to training instabilities. To prevent this, we follow the approach of \cite{immer2023effective}, which use the natural parametrization of the normal distribution, given by $\eta_1 = \mu / \sigma^2, \ \eta_2 = -1/2\sigma^2$ with $\eta_2 <0$, to obtain more stable optimization. The corresponding log-likelihood loss can be expressed in a closed form; for more details, compare \cite{immer2023effective}.
To fulfill the parameter restrictions, we use a softplus activation function on the parameter $\eta_2$ and reverse the sign. \autoref{fig:ur_normal} shows a generated sample, the mean prediction, and the corresponding standard deviation of this method for different datasets.

\begin{figure}[ht]
    \centering
    \begin{subfigure}{0.8\textwidth}
        \centering
        \caption{NYU}
        \includegraphics[width = \linewidth]{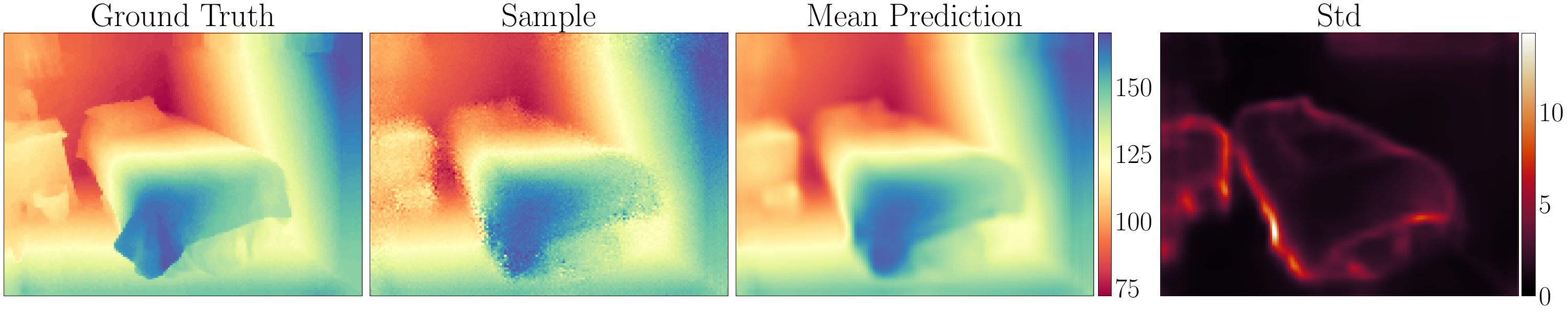}
    \end{subfigure}
    \begin{subfigure}{0.8\textwidth}
        \centering
        \caption{T2M}
        \includegraphics[width = \linewidth]{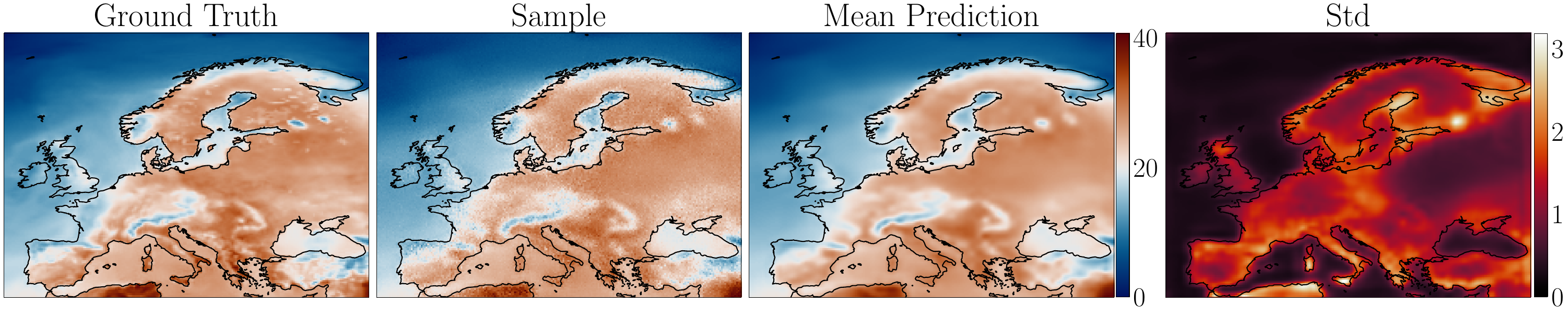}
    \end{subfigure}    
    \caption{Visualizations of the predictions of the deep evidential regression method for the depth regression and surface temperature prediction tasks.}
    \label{fig:ur_der}
\end{figure}

\paragraph{Deep evidential regression}
Moving beyond the first-order Gaussian, \cite{aminiDeepEvidentialRegression2020} propose to directly model second-order uncertainty by predicting the parameters of a Normal Inverse-Gamma (NIG) distribution, which is the conjugate prior of a Gaussian. In particular, one obtains $p(\cdot \mid \vtheta(\boldsymbol{x})) = \mathcal{N}(\mu, \sigma^2)$, with $\mu \sim \mathcal{N}(\gamma(\bx), \sigma^2\upsilon(\bx)^{-1})$ and $\sigma^2 \sim \Gamma^{-1}(\alpha(\bx), \beta(\bx))$. Here, the outputs of our neural networks are the four parameters $\vtheta = (\gamma, \upsilon, \alpha, \beta)^\top$, with $\gamma \in \Ree, \upsilon>0, \alpha >1, \beta > 0$, where the individual constraints are realized using a softplus activation. We use the log-likelihood loss function specified in \cite{aminiDeepEvidentialRegression2020} with a regularization factor $\lambda = 0.01$, as specified by the authors. \autoref{fig:ur_der} shows a generated sample, the mean prediction, and the corresponding standard deviation of this method for different datasets.

\begin{figure}[ht]
    \centering
    \begin{subfigure}{0.8\textwidth}
        \centering
        \caption{NYU}
        \includegraphics[width = \linewidth]{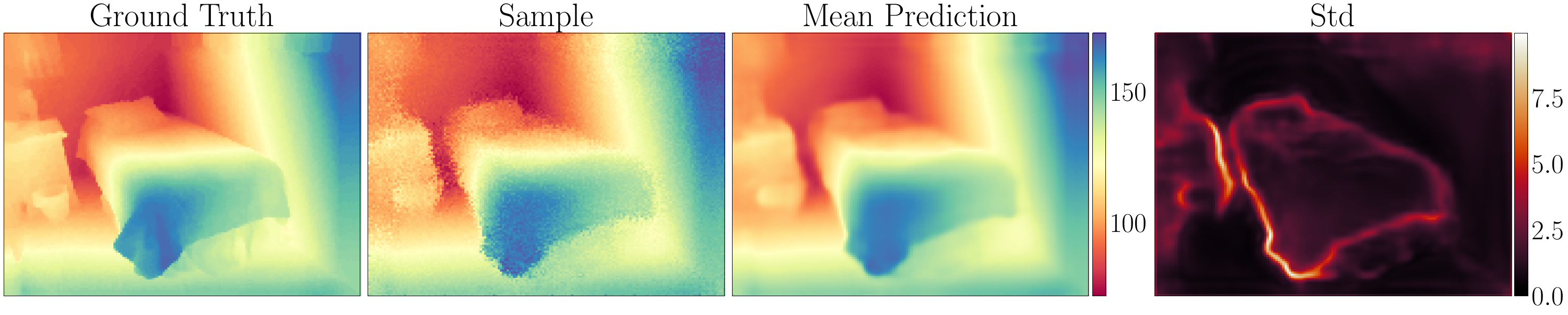}
    \end{subfigure}
    \begin{subfigure}{0.8\textwidth}
        \centering
        \caption{T2M}
        \includegraphics[width = \linewidth]{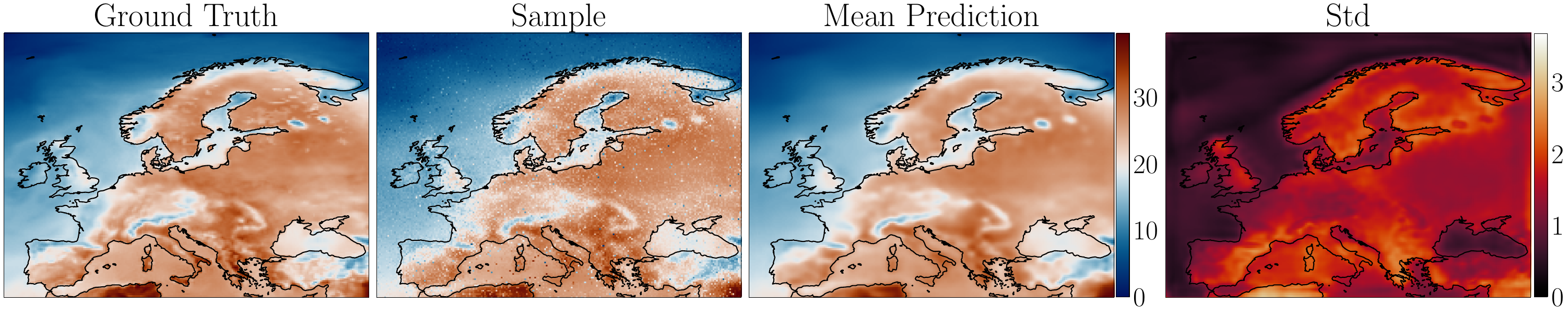}
    \end{subfigure}    
    \caption{Visualizations of the predictions of the mixture density network for the depth regression and surface temperature prediction tasks.}
    \label{fig:ur_mdn}
\end{figure}

\paragraph{Mixture density network}
To accommodate a potential (univariate) multimodal data distribution, we also employ a mixture density network, where the predictive density is specified as $p(\cdot \mid \vtheta(\boldsymbol{x})) = \sum_{k=1}^Kw_k\mathcal{N}(\mu_k(\bx), \sigma_k^2(\bx))$, which is a weighted mixture of several Gaussian distributions. Mixture density networks, as proposed by \cite{bishop} have been employed in machine learning for a long time, but recent work has focused on optimization and performance improvements \citep{Makansi_2019_CVPR, kelen2025distributionfree}, showing competitive performance across several benchmark tasks \citep{kelen2025distributionfree}. To stabilize training, we move beyond the typical log-likelihood loss and train the model using the continuous ranked probability score (CRPS), as proposed by \cite{refId0}. We use a softplus activation for the variance and a softmax for the weights in order to satisfy the corresponding constraints. We choose $K=10$ across all experiments to allow for a higher level of multimodality in the distribution, but omit detailed hyperparameter tuning, as this is not the focus of these experiments. \autoref{fig:ur_mdn} shows a generated sample, the mean prediction, and the corresponding standard deviation of this method for different datasets.

\begin{figure}[ht]
    \centering
    \begin{subfigure}{0.8\textwidth}
        \centering
        \caption{NYU}
        \includegraphics[width = \linewidth]{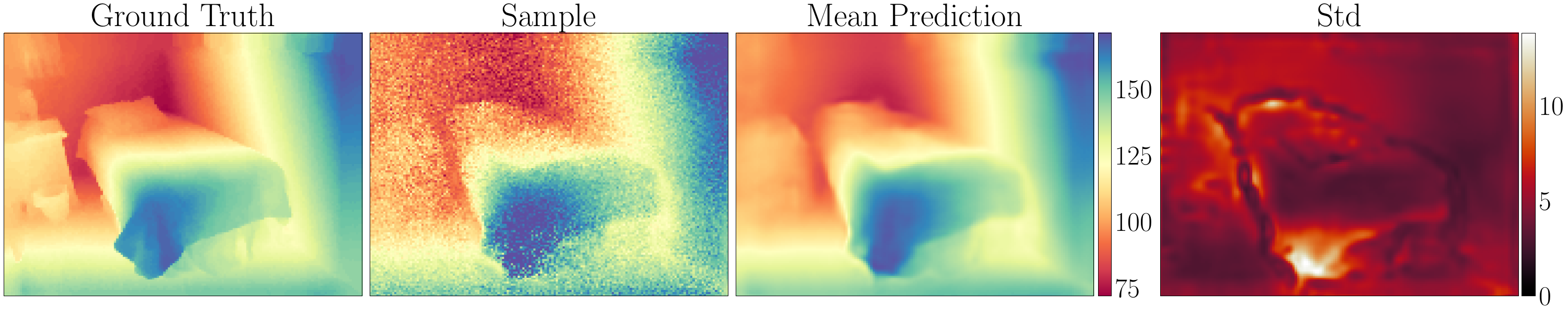}
    \end{subfigure}
    \begin{subfigure}{0.8\textwidth}
        \centering
        \caption{T2M}
        \includegraphics[width = \linewidth]{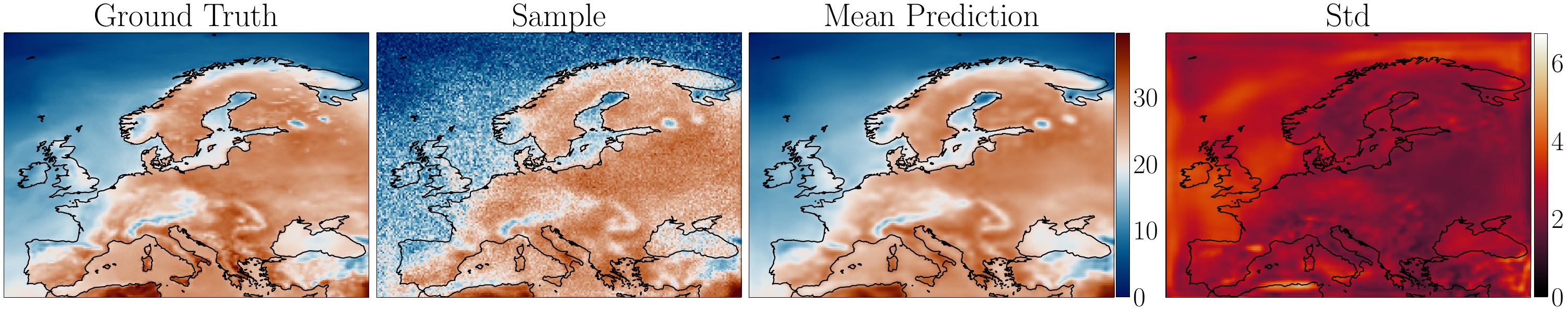}
    \end{subfigure}    
    \caption{Visualizations of the predictions of the low-rank multivariate normal method for the depth regression and surface temperature prediction tasks.}
    \label{fig:ur_lora}
\end{figure}

\paragraph{Low-rank multivariate normal}
Now, we move to a multivariate method, by again specifying a predictive Gaussian, but this time in a multivariate setting, via $p(\cdot \mid \vtheta(\boldsymbol{x})) = \mathcal{N}(\bm\mu(\bx), \Sigma(\bx))$, where $\bm \mu \in \Ree^d$ is the mean vector and $\Sigma \in \Ree^{d \times d}$ is the (positive definite and symmetric) covariance matrix. While the covariance matrix can be modeled using its Cholesky decomposition, which also ensures positive definiteness and symmetry \cite{muschinskiCholeskybasedMultivariateGaussian2024}, for large covariance matrices, this method becomes numerically unstable. Instead, we use a low-rank + diagonal decomposition \citep{rezendeStochasticBackpropagationApproximate2014}, where the covariance matrix is approximated via $\Sigma =  U  U^\top + D$ with a low-rank matrix $U \in \Ree^{d \times r}, \ r \ll d$ and a positive diagonal matrix $D \in \Ree^{d \times d}$.
This allows for computationally efficient modeling of the covariance matrix, while still enabling the model to learn the underlying covariance structure. Throughout the experiments, we use the softplus activation for the diagonal, a rank of $r=10$, and train the model using the Gaussian kernel score, which admits a closed-form expression and offers more stable training as compared to the log-likelihood.
\autoref{fig:ur_lora} shows a generated sample, the mean prediction, and the corresponding standard deviation of this method for different datasets.

\begin{figure}[ht]
    \centering
    \begin{subfigure}{0.8\textwidth}
        \centering
        \caption{NYU}
        \includegraphics[width = \linewidth]{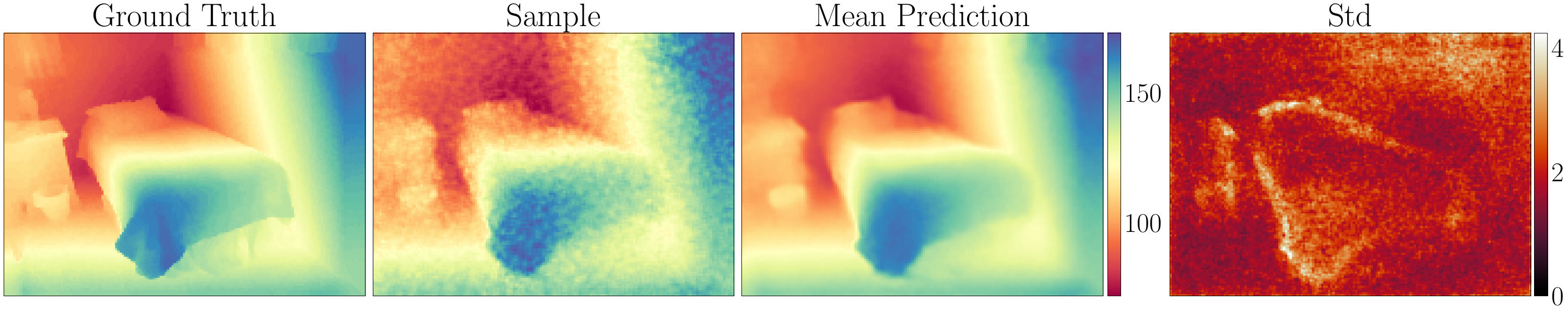}
    \end{subfigure}
    \begin{subfigure}{0.8\textwidth}
        \centering
        \caption{T2M}
        \includegraphics[width = \linewidth]{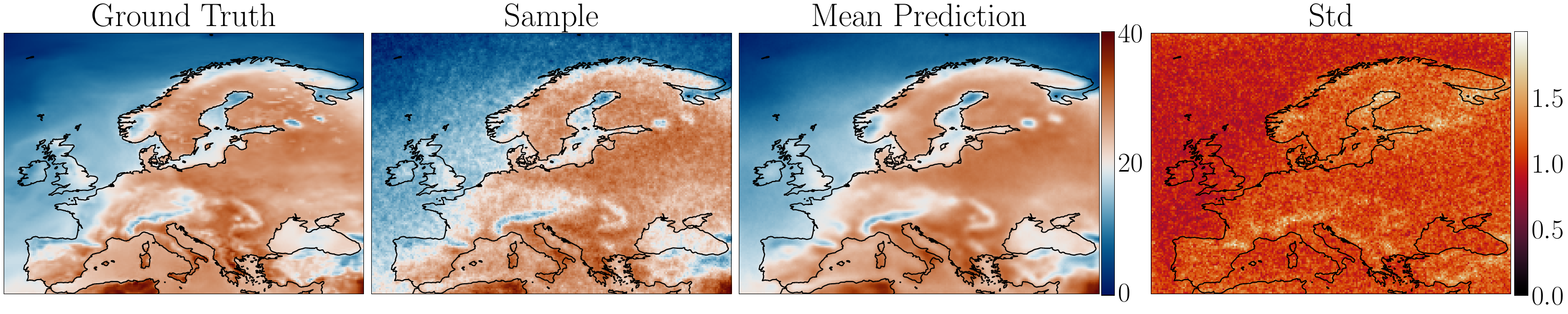}
    \end{subfigure}    
    \caption{Visualizations of the predictions of the sampling-based method for the depth regression and surface temperature prediction tasks.}
    \label{fig:ur_sampling}
\end{figure}

\paragraph{Sampling-based}
Finally, we also use a nonparametric, sampling-based method, which is based on the idea of incorporating noise into the neural network and training it with a (strictly) proper scoring rule. This generative method has recently gained popularity, both regarding theoretical analysis \citep{10.1093/jrsssb/qkae108, JMLR:v25:23-0038}, as well as applications \citep{chen_generative, alet2025skillfuljointprobabilisticweather, bultepno}. In particular, this method has been used both for the univariate setting, using the CRPS as a loss function\cite{kelen2025distributionfree}, as well as the multivariate setting, using the energy score \citep{chen_generative}. Furthermore, this method can also be employed together with neural operators, leading to an empirical distribution over the output function space \citep{bultepno}.
For the univariate tasks, we adopt the setup and loss from \cite{kelen2025distributionfree}, while the multivariate methods are similar to \cite{10.1093/jrsssb/qkae108}, where random noise is concatenated to the input channel of the model and the energy score is used as a loss function. \autoref{fig:ur_sampling} shows a generated sample, the mean prediction and the corresponding standard deviation of this method for different datasets. For this model, we use $N=50$ samples for the univariate, $N=10$ samples for the 1D PDE, and $N=5$ samples for the 2D tasks.

\FloatBarrier
\subsection{Robustness}
Here, we use a regular deep ensemble \citep{lakshminarayananSimpleScalablePredictive2017} on the concrete, energy, and yacht dataset from the UCI regression benchmark \citep{hernándezlobato2015probabilisticbackpropagationscalablelearning}. We train a base ensemble of $M=25$ and $M=5$ members and one additional member that is trained on a distorted target $\hat{y} = y + \mathcal{N}(0, \delta^2)$. This allows us to analyze the robustness of the different uncertainty measures with respect to an outlier in the ensemble prediction. 

First, we provide a theoretical analysis of the robustness in the case of this deep ensemble, which admits a first-order predictive Gaussian distribution $p(y \mid \vtheta) = \mathcal{N}(\mu, \sigma^2), \vtheta = (\mu, \sigma^2)^\top$. Assume that the second-order distribution fulfills $\|\bE_Q[H(P_{\vvtheta})]\| < \infty$, meaning that the aleatoric uncertainty of the sample distribution $Q$ is well defined, which always holds for a finite ensemble. In that case, we can analyze the influence function $ \mathrm{IF}(\vtheta_0; \au, Q)$ directly by analyzing the limit $\lim_{\vtheta_0 \to \infty} H(P_{\vtheta_0})$, since $\bE_Q[H(P_{\vvtheta})]$ is a finite constant. Table \ref{tab:theoretical_robustness} shows the closed-form expressions for $H(\vtheta_0)$, as well as the corresponding growth rates in the contamination $\vtheta_0$. While the Gaussian kernel score is the only scoring rule that is robust, since it admits a bounded influence function, the log-score and CRPS have a notably slower growth rate in $\vtheta_0$ as the variance-based measure, which grows linearly with $\sigma_0^2$.
\begin{table}[ht]
    \centering
    \caption{Limit and corresponding growth rates for the influence function $ \mathrm{IF}(\vtheta_0; \au, Q)$ in the limit $\vtheta_0 \to \infty$.}
    \label{tab:theoretical_robustness}
    \begin{tabular}{llcc}
        \midrule
        $S$        & $H(P_{\vtheta_0})$                                                           & $\lim_{\vtheta_0 \to \infty} H(P_{\vtheta_0})$ & Growth                             \\
        \midrule
        $S_\mathrm{log}$ & $\frac{1}{2} \log(2\pi e \sigma_0^2)$                                        & $\infty$                                       & $\mathcal{O}(\log (\sigma_0^2))$   \\
         $S_\mathrm{SE}$        & $\sigma_0^2$                                                                 & $\infty$                                       & $ \mathcal{O}(\sigma_0^2)$         \\        
         $S_\mathrm{ES}$                  & $ \frac{\sigma_0}{\sqrt{\pi}}$                                               & $\infty$                                       & $\mathcal{O}(\sqrt{\sigma_0^2})$   \\
         $S_{k_\gamma}$ & $ \frac{1}{2} \left(1- \frac{\gamma}{\sqrt{\gamma^2 + 4\sigma_0^2}}\right) $ & $0.5$                                            & $\mathcal{O}(1/\sqrt{\sigma_0^2})$ \\
        \midrule
    \end{tabular}
\end{table}

\begin{figure}[ht]
    \centering
    \begin{subfigure}{\textwidth}
    \caption{Concrete}
    \includegraphics[width = \textwidth]{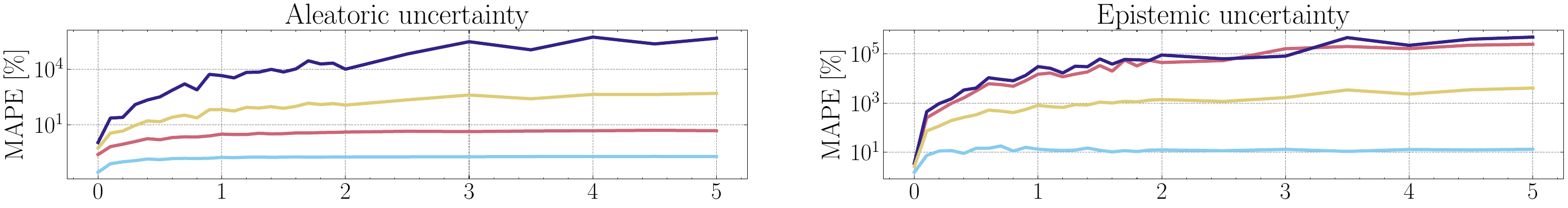}        
    \end{subfigure}
    \begin{subfigure}{\textwidth}
    \caption{Energy}
    \includegraphics[width = \textwidth]{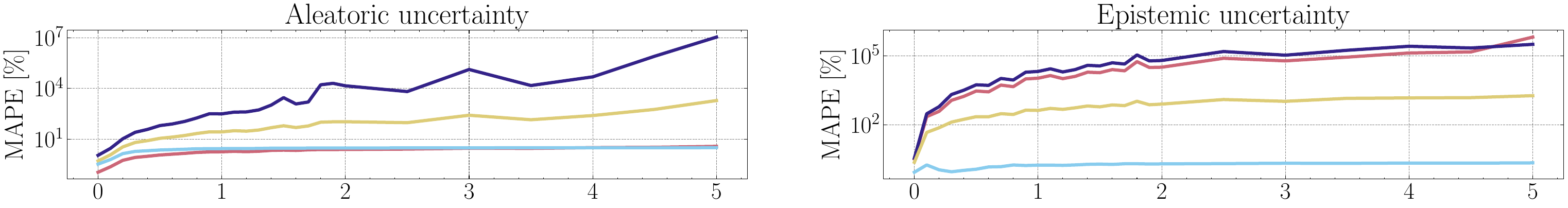}        
    \end{subfigure}
    \begin{subfigure}{\textwidth}
    \caption{Yacht}
    \includegraphics[width = \textwidth]{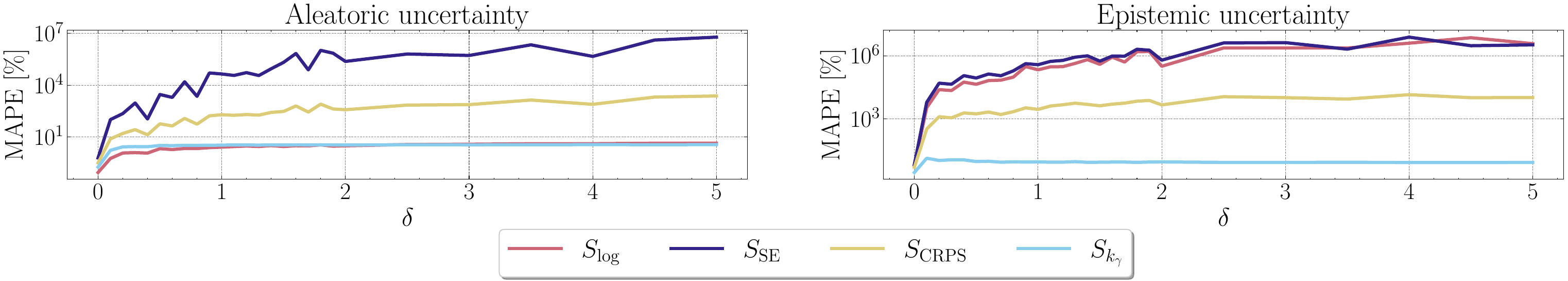}        
    \end{subfigure}%
    \caption{Effect of the added noise $\delta$ on the different uncertainty measures for an ensemble of size $M=25$ across all three datasets. The reported values are the mean absolute percentage error from the corresponding measure for the base ensemble.}
    \label{fig:robustness_full}
\end{figure}

This theoretical analysis is also supported by the numerical results on the UCI benchmark. \autoref{tab:robustness_full} shows the mean absolute percentage error of the epistemic and aleatoric uncertainty from the base ensemble for different values of $\delta$. \autoref{fig:robustness_full} shows corresponding visualizations of the MAPE vs $\delta$ for the different datasets.

\begin{table}[ht]
    \centering
    \caption{Effect of the added noise $\delta$ on the different epistemic and aleatoric
             uncertainty measures across all three datasets for $M=25$ ensemble members.
             The reported values are the mean absolute percentage error from the
             corresponding measure for the base ensemble.}
    \label{tab:robustness_full}
    \begin{tabular}{lllllllll}
    \toprule
        Experiment & Type & $S$ & 0.0 & 0.2 & 0.5 & 1.5 & 2.5 & 5.0 \\
        \midrule
        \multirow{8}{*}{\textbf{Concrete}}
        & \multirow{4}{*}{Aleatoric}
          & $S_\mathrm{log}$ & \num{0.25} & \num{0.90} & \num{1.56} & \num{3.34} & \num{4.47} & \num{4.82} \\
        & & $S_\mathrm{SE}$  & \num{1.11} & \num{2.53e1} & \num{3.24e2} & \num{7.21e3} & \num{6.77e4} & \num{4.78e5} \\
        & & $S_\mathrm{ES}$  & \num{0.55} & \num{4.65} & \num{1.47e1} & \num{7.83e1} & \num{2.24e2} & \num{5.04e2} \\
        & & $S_{k_\gamma}$   & \num{0.03} & \num{0.10} & \num{0.13} & \num{0.18} & \num{0.19} & \num{0.20} \\
        \cline{2-9}
        & \multirow{4}{*}{Epistemic}
          & $S_\mathrm{log}$ & \num{3.49} & \num{4.99e2} & \num{3.10e3} & \num{3.32e4} & \num{5.29e4} & \num{2.43e5} \\
        & & $S_\mathrm{SE}$  & \num{3.71} & \num{9.49e2} & \num{4.08e3} & \num{6.17e4} & \num{6.22e4} & \num{4.74e5} \\
        & & $S_\mathrm{ES}$  & \num{2.68} & \num{1.16e2} & \num{3.35e2} & \num{1.09e3} & \num{1.16e3} & \num{4.05e3} \\
        & & $S_{k_\gamma}$   & \num{1.57} & \num{1.13e1} & \num{1.47e1} & \num{1.21e1} & \num{1.17e1} & \num{1.33e1} \\
        \midrule
        \multirow{8}{*}{\textbf{Energy}}
        & \multirow{4}{*}{Aleatoric}
          & $S_\mathrm{log}$ & \num{0.11} & \num{0.57} & \num{1.17} & \num{2.28} & \num{2.62} & \num{3.79} \\
        & & $S_\mathrm{SE}$  & \num{1.12} & \num{1.07e1} & \num{6.41e1} & \num{2.86e3} & \num{6.57e3} & \num{1.09e7} \\
        & & $S_\mathrm{ES}$  & \num{0.52} & \num{3.52} & \num{1.13e1} & \num{6.24e1} & \num{9.52e1} & \num{1.93e3} \\
        & & $S_{k_\gamma}$   & \num{0.34} & \num{1.41} & \num{2.31} & \num{2.97} & \num{3.07} & \num{3.17} \\
        \cline{2-9}
        & \multirow{4}{*}{Epistemic}
          & $S_\mathrm{log}$ & \num{3.33} & \num{3.86e2} & \num{2.94e3} & \num{1.84e4} & \num{7.74e4} & \num{6.45e5} \\
        & & $S_\mathrm{SE}$  & \num{3.57} & \num{6.16e2} & \num{5.47e3} & \num{3.62e4} & \num{1.52e5} & \num{3.12e5} \\
        & & $S_\mathrm{ES}$  & \num{2.38} & \num{7.49e1} & \num{2.26e2} & \num{6.05e2} & \num{1.26e3} & \num{1.84e3} \\
        & & $S_{k_\gamma}$   & \num{0.88} & \num{1.12} & \num{1.19} & \num{1.98} & \num{2.08} & \num{2.24} \\
        \midrule
        \multirow{8}{*}{\textbf{Yacht}}
        & \multirow{4}{*}{Aleatoric}
          & $S_\mathrm{log}$ & \num{0.09} & \num{1.17} & \num{2.08} & \num{2.75} & \num{3.61} & \num{4.3} \\
        & & $S_\mathrm{SE}$  & \num{0.60} & \num{2.25e1} & \num{2.88e3} & \num{2.1e5} & \num{6.42e5} & \num{6.1e6} \\
        & & $S_\mathrm{ES}$  & \num{0.31} & \num{1.59e1} & \num{5.75e1} & \num{2.99e2} & \num{7e2} & \num{2.36e3} \\
        & & $S_{k_\gamma}$   & \num{0.18} & \num{2.65} & \num{3.18} & \num{3.36} & \num{3.52} & \num{3.54} \\
        \cline{2-9}
        & \multirow{4}{*}{Epistemic}
          & $S_\mathrm{log}$ & \num{6.04} & \num{2.43e4} & \num{4.39e4} & \num{3.89e5} & \num{2.34e6} & \num{3.77e6} \\
        & & $S_\mathrm{SE}$  & \num{6.58} & \num{4.95e4} & \num{8.69e4} & \num{5.53e5} & \num{4.06e6} & \num{3.31e6} \\
        & & $S_\mathrm{ES}$  & \num{4.87} & \num{1.25e3} & \num{1.71e3} & \num{4.13e3} & \num{1.12e4} & \num{1.02e4} \\
        & & $S_{k_\gamma}$   & \num{2.75} & \num{1.03e1} & \num{9.13} & \num{8.49} & \num{8.32} & \num{8.31} \\
        \bottomrule
    \end{tabular}
\end{table}

\FloatBarrier
\subsection{Selective prediction}
\label{app:selective_prediction}
\begin{figure}[ht]
    \centering
    \begin{subfigure}{\textwidth}
    \caption{Energy}
    \includegraphics[width = \textwidth]{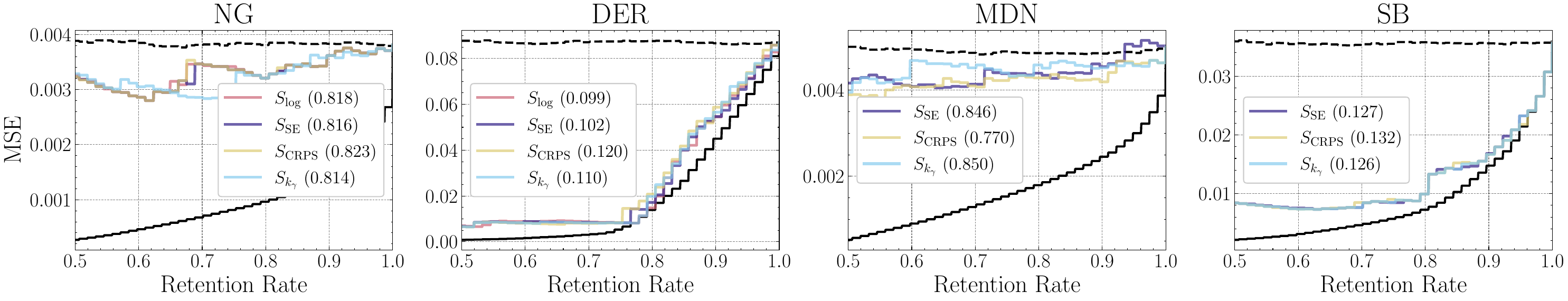}        
    \end{subfigure}
    \begin{subfigure}{\textwidth}
    \caption{Yacht}
    \includegraphics[width = \textwidth]{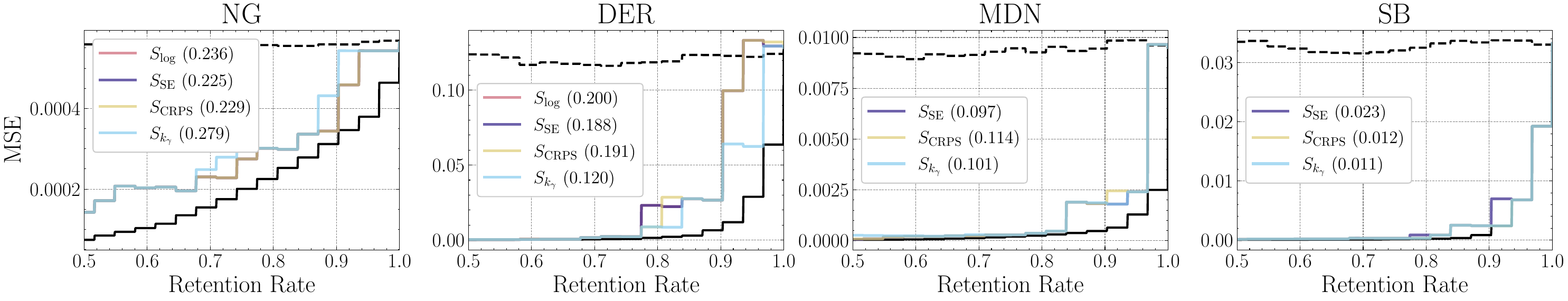}        
    \end{subfigure}
    \begin{subfigure}{\textwidth}
    \caption{Kin8nm}
    \includegraphics[width = \textwidth]{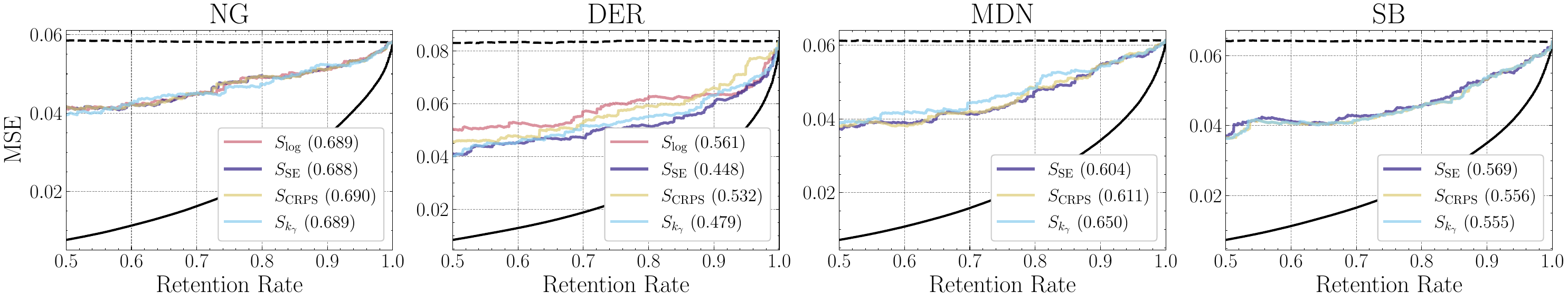}        
    \end{subfigure}
    \begin{subfigure}{\textwidth}
    \caption{Protein}
    \includegraphics[width = \textwidth]{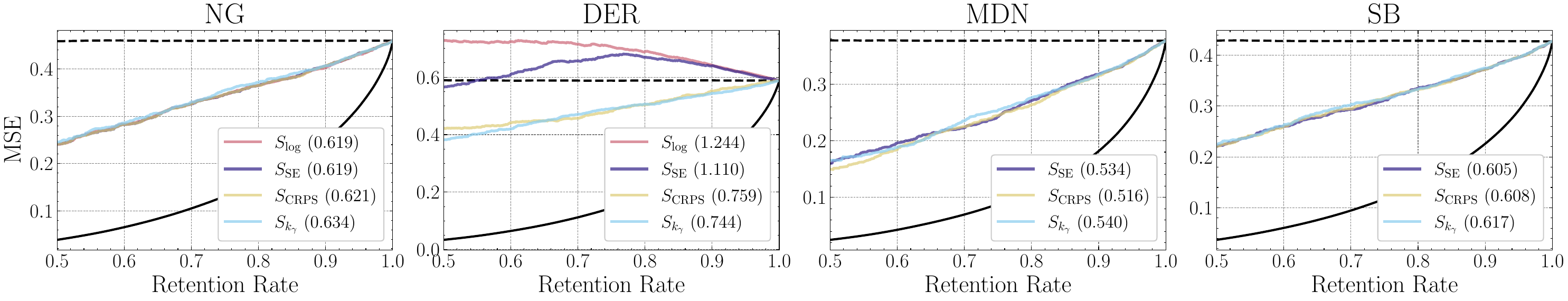}        
    \end{subfigure}
    \caption{Retention curves with the reported PRR per uncertainty measure for four of the different UCI datasets. The black solid line is the optimal retention, where the curve is sorted by the actual MSE per prediction. The black dashed line is the random baseline.}
    \label{fig:selective_prediction_uci}
\end{figure}

\begin{figure}[ht]
    \centering
    \includegraphics[width = \textwidth]{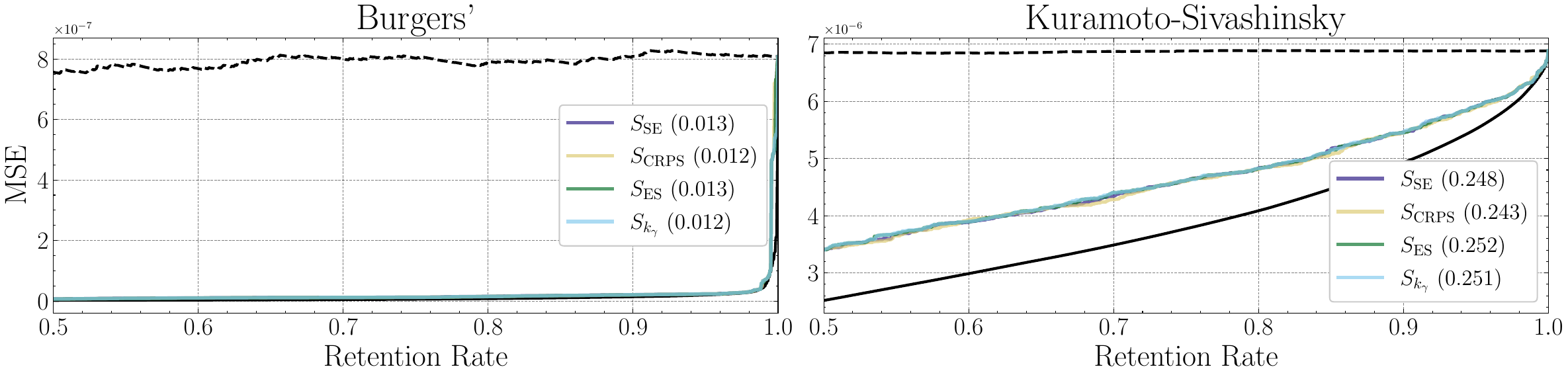}
    \caption{Retention curves with the reported PRR per uncertainty measure for the two one-dimensional PDE datasets. The black solid line is the optimal retention, where the curve is sorted by the actual MSE per prediction. The black dashed line is the random baseline.}
    \label{fig:selective_prediction_pde}
\end{figure}

\begin{figure}[ht]
    \centering
    \begin{subfigure}{\textwidth}
    \caption{Depth}
    \includegraphics[width = \textwidth]{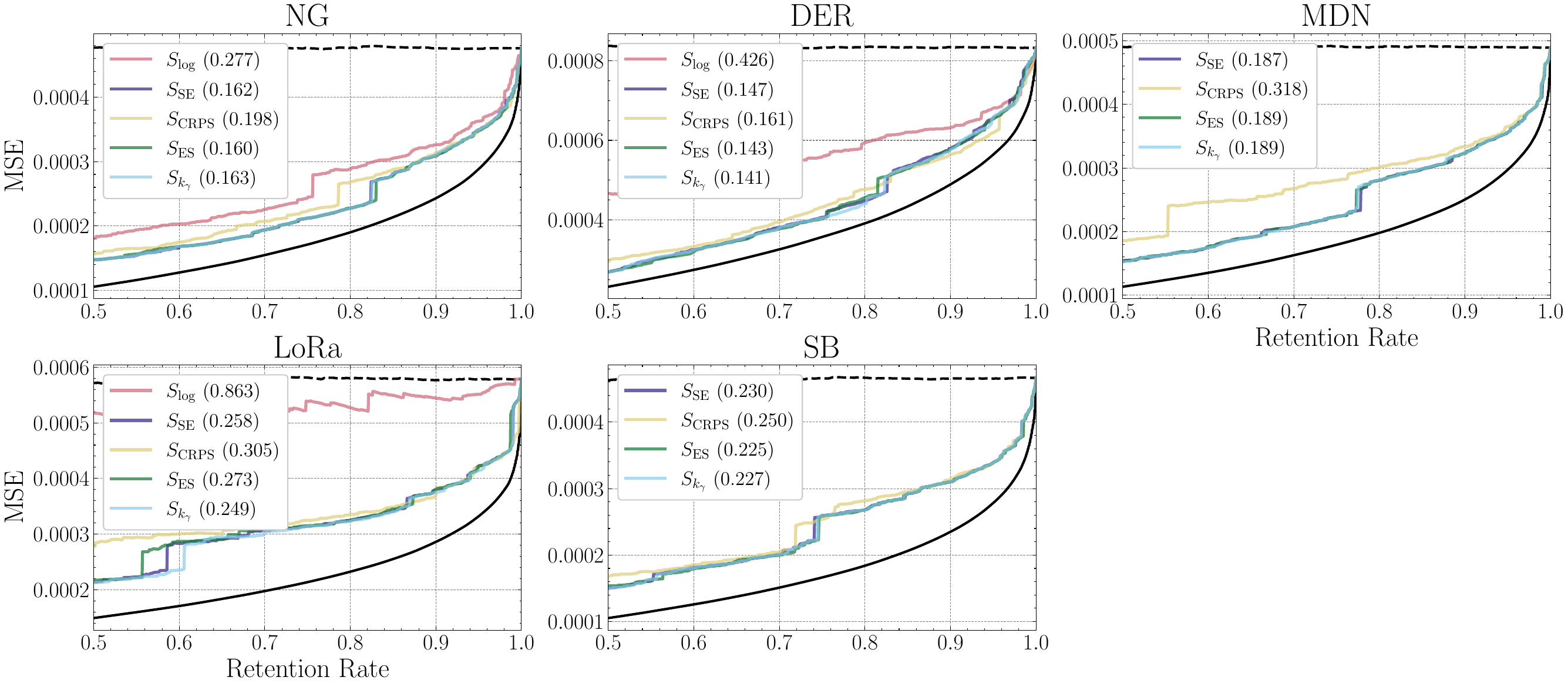}   
    \end{subfigure}
    \begin{subfigure}{\textwidth}
    \caption{T2M}
    \includegraphics[width = \textwidth]{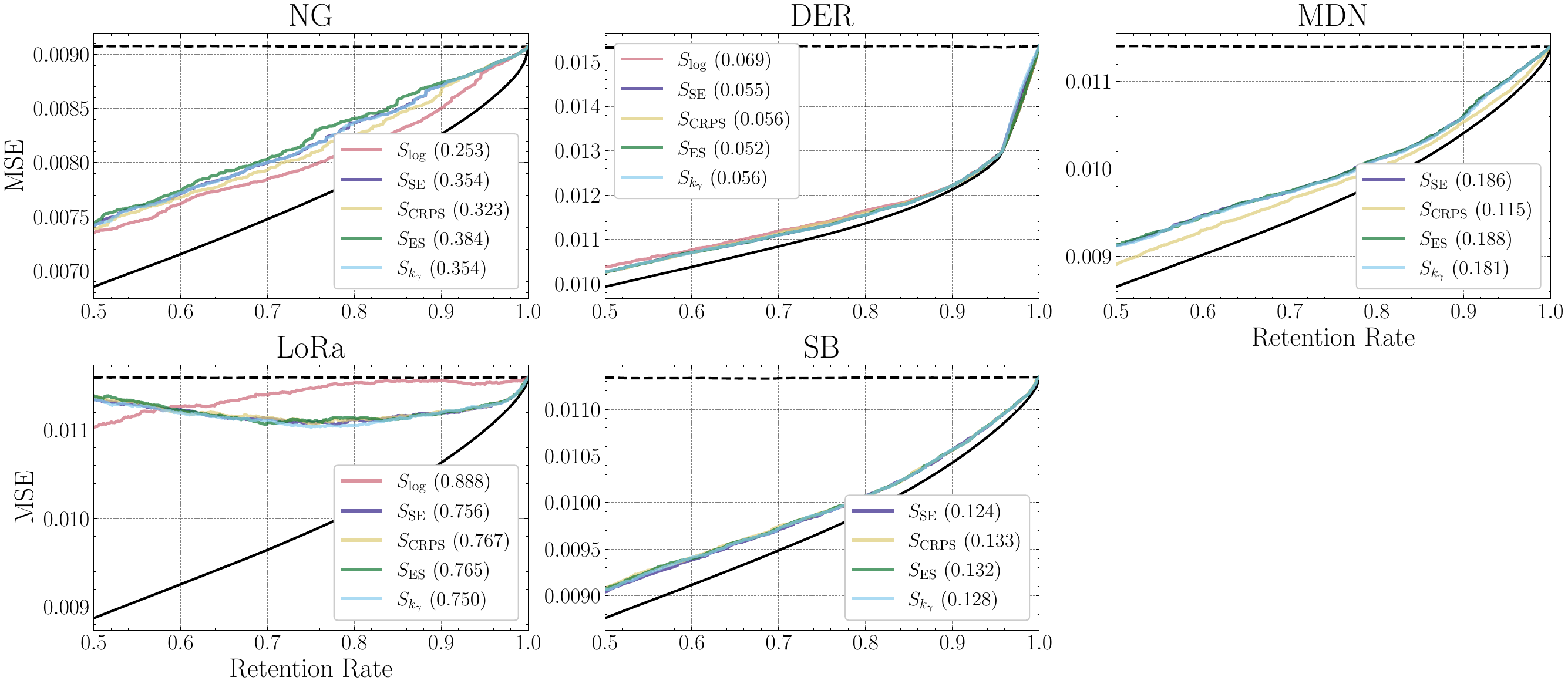}   
    \end{subfigure}
    \caption{Retention curves with the reported PRR per uncertainty measure for the two 2D tasks. The black solid line is the optimal retention, where the curve is sorted by the actual MSE per prediction. The black dashed line is the random baseline.}
    \label{fig:selective_prediction_2d}
\end{figure}

We evaluate selective prediction on all datasets using the corresponding and uncertainty representation methods described in this section. As an evaluation criterion, we evaluate the prediction-reject ratio \citep[PRR,][]{malinin2021uncertaintyestimationautoregressivestructured}, which can be extended to the regression setting using the mean squared error (MSE) as a performance metric \citep{fishkov2025uncertaintyquantificationregressionusing}. We evaluate using retention rates, i.e. $1-$rejection, from 0.5 to 1. The full results are available in \autoref{tab:selective_prediction_full} and corresponding visualizations of the retention curves for the different methods in Figures \ref{fig:selective_prediction_uci}, \ref{fig:selective_prediction_pde}, and \ref{fig:selective_prediction_2d}.

\begin{table}[ht]
    \centering
        \caption{Prediction-reject-ratios (PRR $\downarrow$) for all different datasets, uncertainty representation methods and uncertainty measures, where the best measure for each configuration is highlighted in bold. Note that for the univariate tasks, the $S_\mathrm{ES}$ coincides with $S_\mathrm{CRPS}$ and the values are omitted for readability.}
    \label{tab:selective_prediction_full}
\begin{tabular}{llrrrrr}
\toprule
Dataset & Method & $S_\mathrm{log}$ & $S_\mathrm{SE}$ & $S_\mathrm{CRPS}$ & $S_\mathrm{ES}$ & $S_{k_\gamma}$ \\
\midrule
Burgers' & SB & - & 0.013 & 0.013 & 0.013 & \textbf{0.012} \\
\cline{1-7}
KS & SB & - & 0.247 & \textbf{0.242} & 0.251 & 0.252 \\
\cline{1-7}
\multirow{5}{*}{Depth} 
 & NG & 0.278 & 0.162 & 0.197 & \textbf{0.161} & 0.162 \\
& DER & 0.426 & 0.147 & 0.160 & 0.143 & \textbf{0.142 }\\
 & MDN & - & \textbf{0.188} & 0.314 & 0.189 & 0.189 \\
 & LoRa & 0.859 & 0.262 & 0.310 & 0.272 & \textbf{0.251} \\
 & SB & - & 0.229 & 0.251 & \textbf{0.225 }& 0.227 \\
 \cline{1-7}
 \multirow{5}{*}{T2M}
 & NG & \textbf{0.253} & 0.356 & 0.322 & 0.385 & 0.356 \\
 & DER & 0.068 & 0.054 & 0.055 & \textbf{0.052 }& 0.056 \\
 & MDN & - & 0.186 & \textbf{0.116} & 0.188 & 0.181 \\
 & LoRa & 0.889 & 0.759 & 0.766 & 0.764 & \textbf{0.750} \\
 & SB & - & \textbf{0.124} & 0.134 & 0.132 & 0.129 \\
\cline{1-7}
\multirow{4}{*}{Concrete}
 & NG & 0.648 & \textbf{0.644} & \textbf{0.644 }& - & 0.742 \\
& DER & 0.591 & \textbf{0.566} & 0.606 & - & 0.709 \\
 & MDN & - & 0.503 &\textbf{ 0.497} & - & 0.531 \\
 & SB & - & 0.612 & 0.609 & - & \textbf{0.608 }\\
\cline{1-7}
\multirow{4}{*}{Energy}
 & NG & 0.824 & 0.818 & 0.823 & - & \textbf{0.798} \\
& DER & \textbf{0.101} & 0.103 & 0.121 & - & 0.110 \\
 & MDN & - & 0.820 & \textbf{0.770 }& - & 0.848 \\
 & SB & - & 0.125 & 0.125 & - & \textbf{0.124 }\\
\cline{1-7}
\multirow{4}{*}{Kin8nm}
 & NG & 0.690 & 0.687 & 0.688 & - & \textbf{0.682} \\
& DER & 0.560 & \textbf{0.448} & 0.535 & - & 0.479 \\
 & MDN & - & \textbf{0.602} & 0.610 & - & 0.646 \\
 & SB & - & 0.573 & \textbf{0.556} & - & 0.558 \\
\cline{1-7}
\multirow{4}{*}{Naval} & DER & 0.564 & \textbf{0.177} & 0.743 & - & 0.187 \\
 & MDN & - & 0.278 & \textbf{0.231} & - & 0.422 \\
 & NG & \textbf{0.530} & 0.551 & 0.545 & - & 0.608 \\
 & SB & - & 0.305 & 0.300 & - & \textbf{0.299} \\
\cline{1-7}
\multirow{4}{*}{Power}
 & NG & 0.936 & 0.936 & 0.936 & - & \textbf{0.891} \\
& DER & 0.931 & \textbf{0.851} & 0.891 & - & 0.958 \\
 & MDN & - & 0.745 & \textbf{0.735} & - & 0.765 \\
 & SB & - & 0.848 & 0.842 & - & \textbf{0.841} \\
\cline{1-7}
\multirow{4}{*}{Protein}
 & NG & \textbf{0.620 }& \textbf{0.620} & \textbf{0.620} & - & 0.633 \\
& DER & 1.243 & 1.110 & 0.759 & - & \textbf{0.743 }\\
 & MDN & - & 0.535 & \textbf{0.518} & - & 0.540 \\
 & SB & - & \textbf{0.607} & 0.609 & - & 0.616 \\
\cline{1-7}
\multirow{4}{*}{Yacht}
 & NG & \textbf{0.241 }& \textbf{0.241} & \textbf{0.241} & - & 0.296 \\
& DER & 0.201 & 0.201 & 0.198 & - & \textbf{0.124 }\\
 & MDN & - & \textbf{0.103} & 0.107 & - & 0.105 \\
 & SB & - & 0.024 & \textbf{0.012} & - & \textbf{0.012} \\
\bottomrule
\end{tabular}
\end{table}

\FloatBarrier
\subsection{Out-of-distribution detection}
\label{app:ood}
\begin{figure}[ht]
    \centering
    \begin{subfigure}{\textwidth}
    \caption{$S_\mathrm{SE}$ - ID}
    \includegraphics[width = \textwidth]{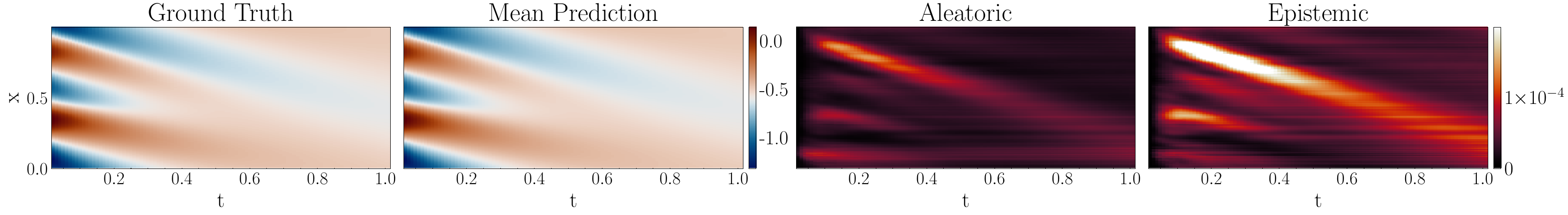}        
    \end{subfigure}
    \begin{subfigure}{\textwidth}
    \caption{$S_\mathrm{SE}$ - OOD}
    \includegraphics[width = \textwidth]{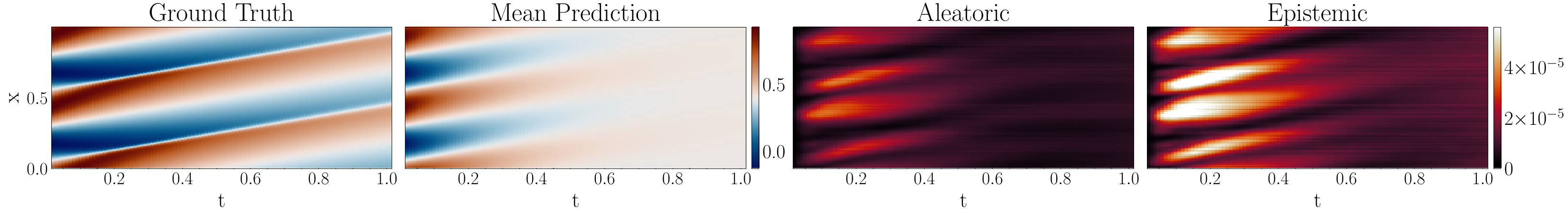}        
    \end{subfigure}
    \begin{subfigure}{\textwidth}
    \caption{$S_\mathrm{CRPS}$ - ID}
    \includegraphics[width = \textwidth]{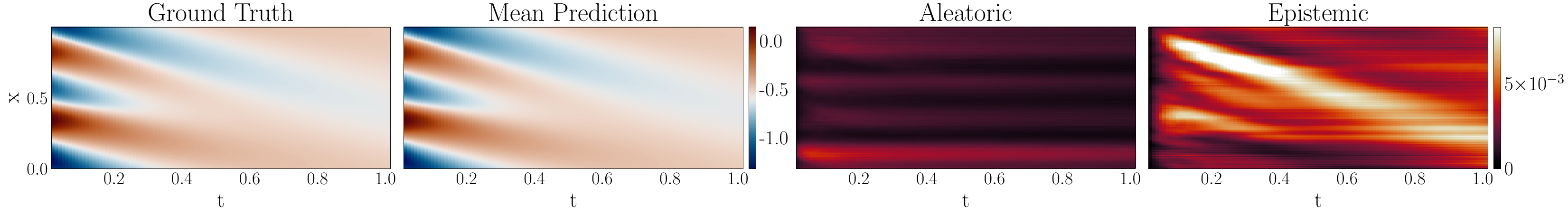}        
    \end{subfigure}
    \begin{subfigure}{\textwidth}
    \caption{$S_\mathrm{CRPS}$ - OOD}
    \includegraphics[width = \textwidth]{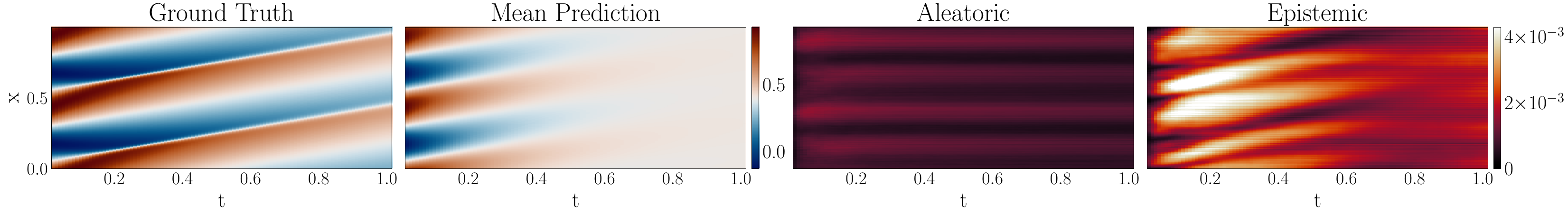}        
    \end{subfigure}
    \caption{Predictions and uncertainty estimates for a selected sample of the Burgers' equation. Shown are the in-distribution (ID) and out-of-distribution (OOD) predictions for the $S_\mathrm{SE}$ and $S_\mathrm{CRPS}$ measures, as those can be visualized pointwise.}
    \label{fig:ood_burgers}
\end{figure}

\begin{figure}[ht]
    \centering
    \begin{subfigure}{\textwidth}
    \caption{$S_\mathrm{SE}$ - ID}
    \includegraphics[width = \textwidth]{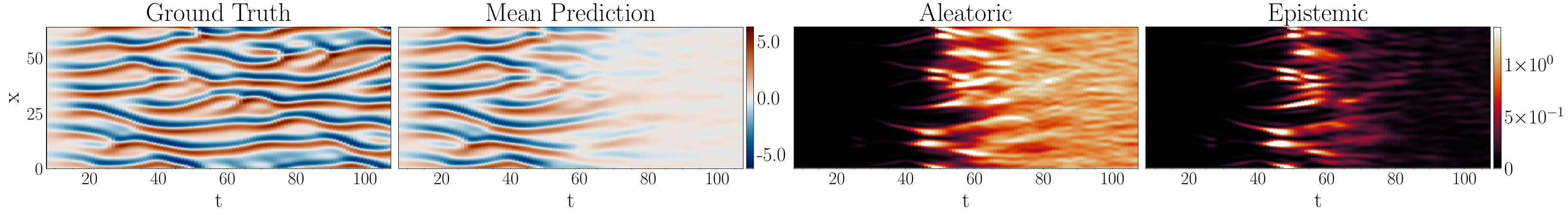}        
    \end{subfigure}
    \begin{subfigure}{\textwidth}
    \caption{$S_\mathrm{SE}$ - OOD}
    \includegraphics[width = \textwidth]{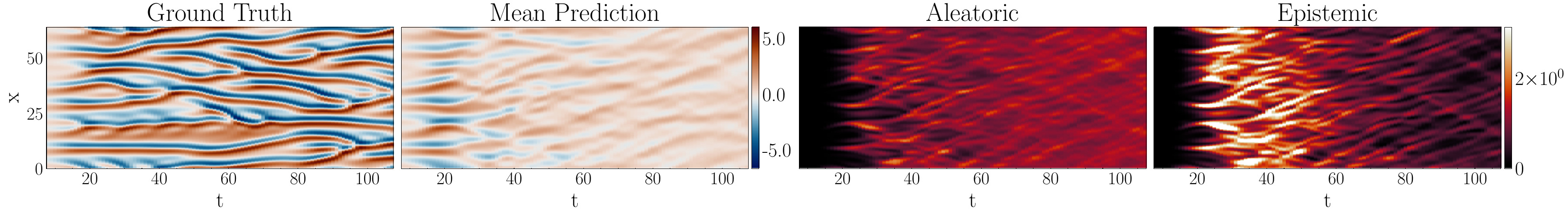}        
    \end{subfigure}
    \begin{subfigure}{\textwidth}
    \caption{$S_\mathrm{CRPS}$ - ID}
    \includegraphics[width = \textwidth]{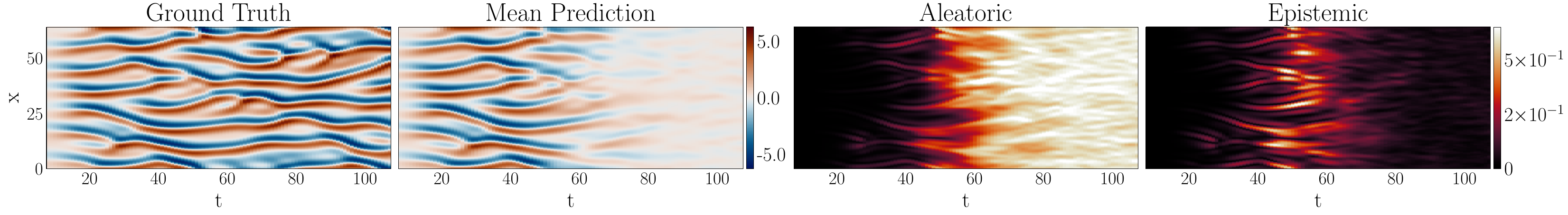}        
    \end{subfigure}
    \begin{subfigure}{\textwidth}
    \caption{$S_\mathrm{CRPS}$ - OOD}
    \includegraphics[width = \textwidth]{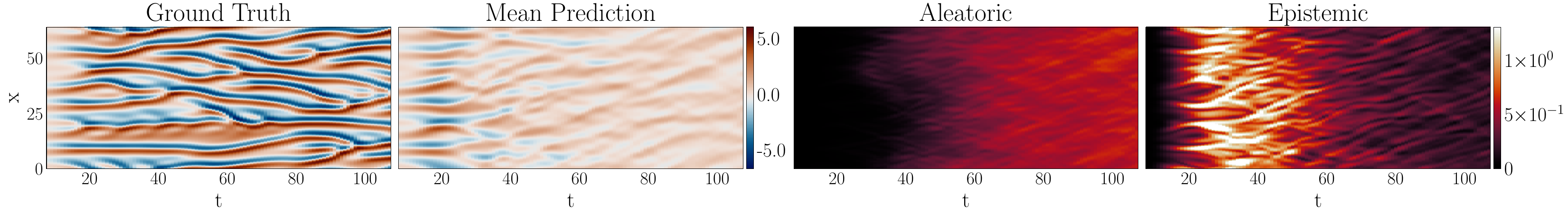}        
    \end{subfigure}
    \caption{Predictions and uncertainty estimates for a selected sample of the Kuramoto-Sivashinsky equation. Shown are the in-distribution (ID) and out-of-distribution (OOD) predictions for the $S_\mathrm{SE}$ and $S_\mathrm{CRPS}$ measures, as those can be visualized pointwise.}
    \label{fig:ood_ks}
\end{figure}

\begin{figure}[ht]
    \centering
    \begin{subfigure}{\textwidth}
    \caption{$S_\mathrm{SE}$ - ID}
    \includegraphics[width = \textwidth]{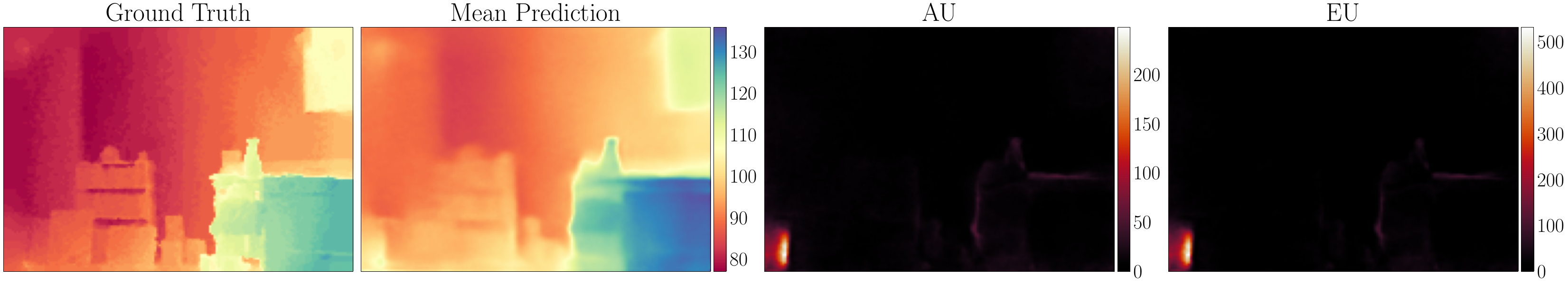}        
    \end{subfigure}
    \begin{subfigure}{\textwidth}
    \caption{$S_\mathrm{SE}$ - OOD}
    \includegraphics[width = \textwidth]{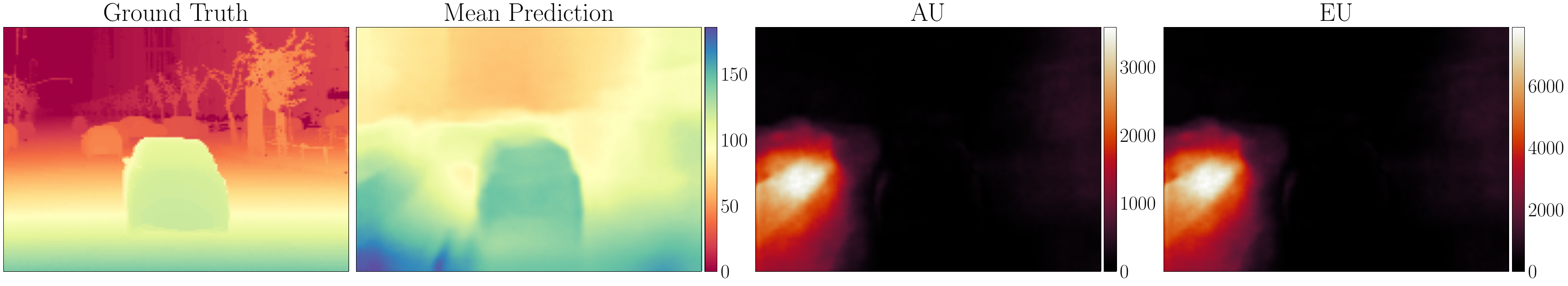}        
    \end{subfigure}
    \begin{subfigure}{\textwidth}
    \caption{$S_\mathrm{CRPS}$ - ID}
    \includegraphics[width = \textwidth]{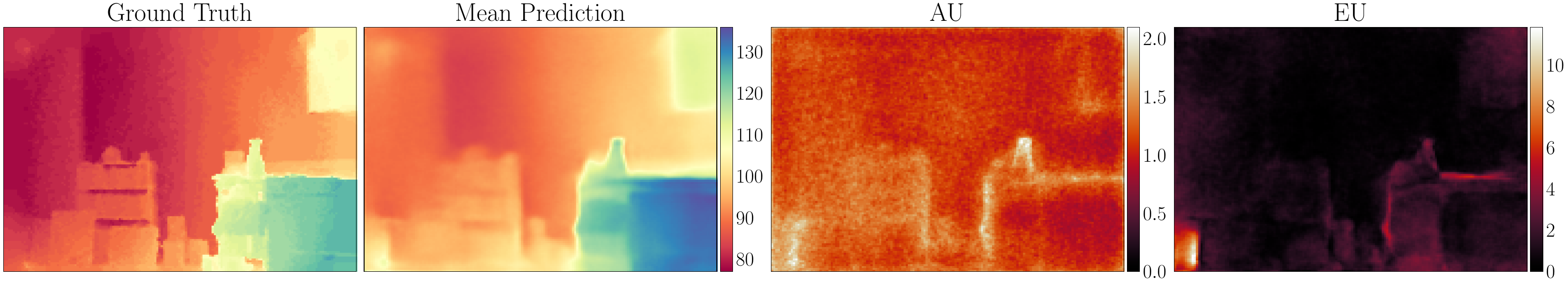}        
    \end{subfigure}
    \begin{subfigure}{\textwidth}
    \caption{$S_\mathrm{CRPS}$ - OOD}
    \includegraphics[width = \textwidth]{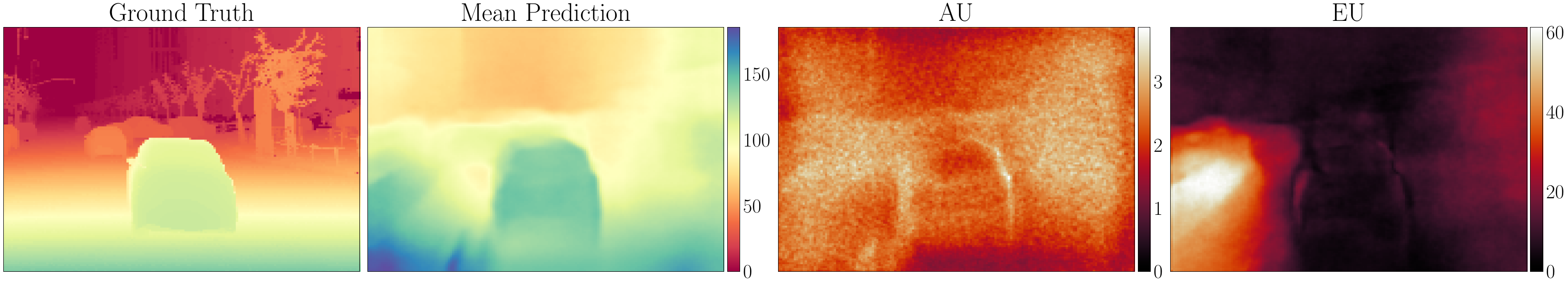}        
    \end{subfigure}
    \caption{Predictions and uncertainty estimates for a selected sample of the NYU (ID) and ApolloScape (OOD) datasets. Shown are the in-distribution (ID) and out-of-distribution (OOD) predictions for the $S_\mathrm{SE}$ and $S_\mathrm{CRPS}$ measures, as those can be visualized pointwise.}
    \label{fig:ood_depth}
\end{figure}

\begin{figure}[ht]
    \centering
    \begin{subfigure}{\textwidth}
    \caption{$S_\mathrm{SE}$ - ID}
    \includegraphics[width = \textwidth]{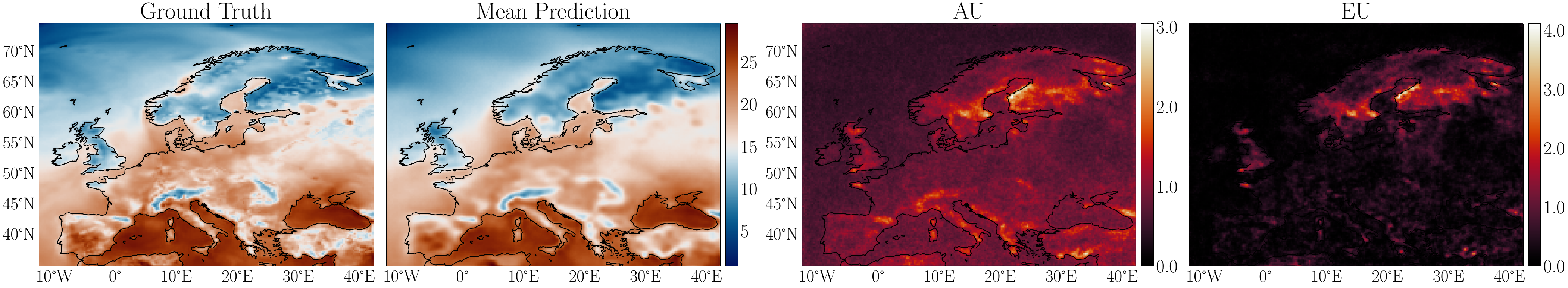}        
    \end{subfigure}
    \begin{subfigure}{\textwidth}
    \caption{$S_\mathrm{SE}$ - OOD}
    \includegraphics[width = \textwidth]{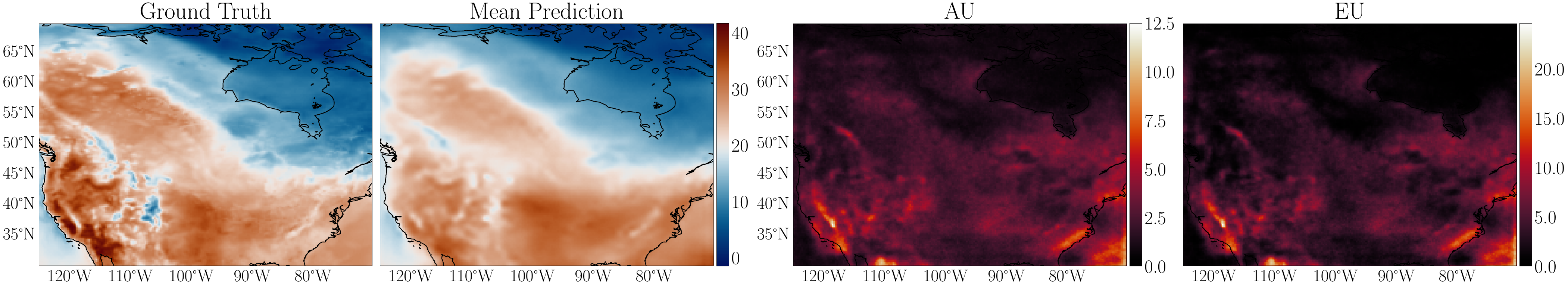}        
    \end{subfigure}
    \begin{subfigure}{\textwidth}
    \caption{$S_\mathrm{CRPS}$ - ID}
    \includegraphics[width = \textwidth]{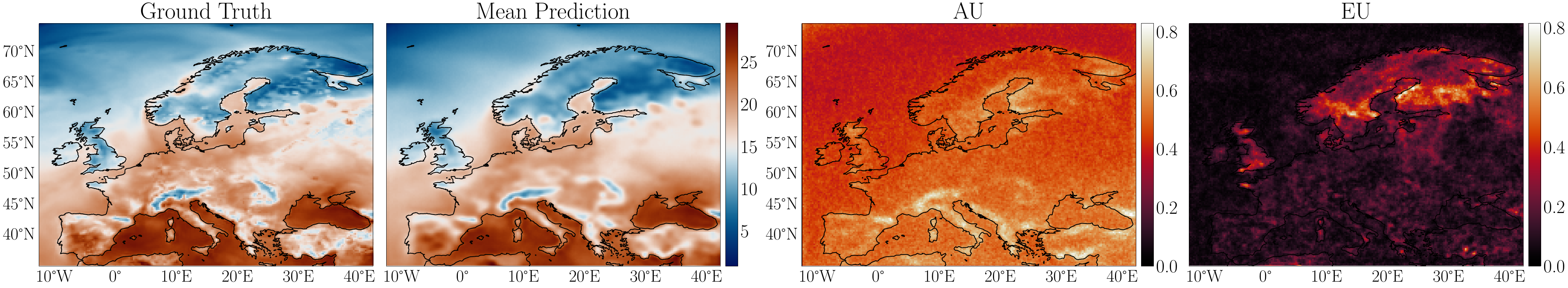}        
    \end{subfigure}
    \begin{subfigure}{\textwidth}
    \caption{$S_\mathrm{CRPS}$ - OOD}
    \includegraphics[width = \textwidth]{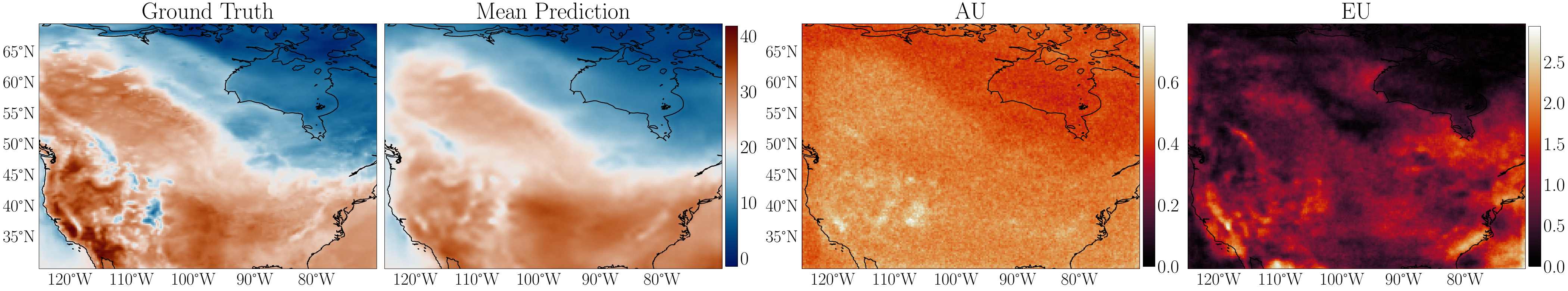}        
    \end{subfigure}
    \caption{Predictions and uncertainty estimates for a selected sample of surface temperature prediction task across the European (ID) and North American (OOD) domains. Shown are the in-distribution (ID) and out-of-distribution (OOD) predictions for the $S_\mathrm{SE}$ and $S_\mathrm{CRPS}$ measures, as those can be visualized pointwise.}
    \label{fig:ood_t2m}
\end{figure}

For evaluation, we use ID and OOD datasets of the same size, which we achieve by subsampling, if necessary, and compute the AUROC across all datasets and uncertainty representation methods. For the out-of-distribution generation procedure for each dataset, compare Appendix~\ref{app:datasets}. The full results are listed in \autoref{tab:ood_full}. Selected visualizations for the sampling-based method over the different datasets can be found in Figure~\ref{fig:ood_burgers}, \ref{fig:ood_ks}, \ref{fig:ood_depth}, and \ref{fig:ood_t2m}.

\begin{table}[ht]
    \centering
        \caption{Out-of-distribution detection (AUROC $\uparrow$) for all different datasets, uncertainty representation methods and uncertainty measures, where the best measure for each configuration is highlighted in bold.}
    \label{tab:ood_full}
\begin{tabular}{llrrrrr}
\toprule
Dataset & Method & $S_\mathrm{log}$ & $S_\mathrm{SE}$ & $S_\mathrm{CRPS}$ & $S_\mathrm{ES}$ & $S_{k_\gamma}$ \\
\midrule
Burgers' & SB & - & 0.9307 & 0.9371 & \textbf{0.9397} & 0.9327 \\
\cline{1-7}
KS & SB & - & \textbf{1.0000} & \textbf{1.0000} & \textbf{1.0000 }& \textbf{1.0000 }\\
\cline{1-7}
\multirow{5}{*}{Depth} 
 & NG & \textbf{0.9999} & 0.9995 & 0.9998 & 0.9998 & 0.9997 \\
& DER & \textbf{0.9923} & 0.9542 & 0.9886 & 0.9549 & 0.9590 \\
 & MDN & - & \textbf{1.0000} & \textbf{1.0000} & \textbf{1.0000} & \textbf{1.0000} \\
 & LoRa & 0.2951 & 0.9991 & \textbf{0.9997} & 0.9994 & 0.9993 \\
 & SB & - & 0.9994 & \textbf{0.9999} & \textbf{0.9999} & 0.9996 \\
\cline{1-7}
\multirow{5}{*}{T2M}
 & NG & \textbf{1.0000} & \textbf{1.0000} & \textbf{1.0000} & \textbf{1.0000} & \textbf{1.0000} \\
& DER & \textbf{1.0000} & 0.9986 & 0.9559 & 0.9547 & 0.9747 \\
 & MDN & - & \textbf{1.0000} & \textbf{1.0000} & \textbf{1.0000} & \textbf{1.0000} \\
  & LoRa & 0.8066 & \textbf{1.0000} & \textbf{1.0000} & \textbf{1.0000} & \textbf{1.0000} \\
 & SB & - & \textbf{1.0000} & \textbf{1.0000} & \textbf{1.0000} & \textbf{1.0000} \\
\bottomrule
\bottomrule
\end{tabular}
\end{table}

\FloatBarrier
\subsection{Active Learning}
\label{app:active_learning}
\begin{figure}[ht]
    \centering
    \begin{subfigure}{\textwidth}
    \caption{Energy}
    \includegraphics[width = \textwidth]{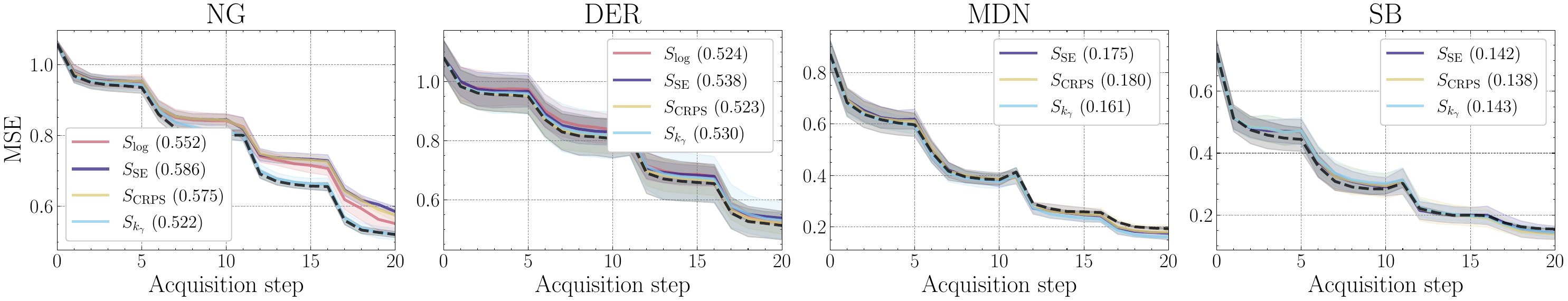}        
    \end{subfigure}
    \begin{subfigure}{\textwidth}
    \caption{Yacht}
    \includegraphics[width = \textwidth]{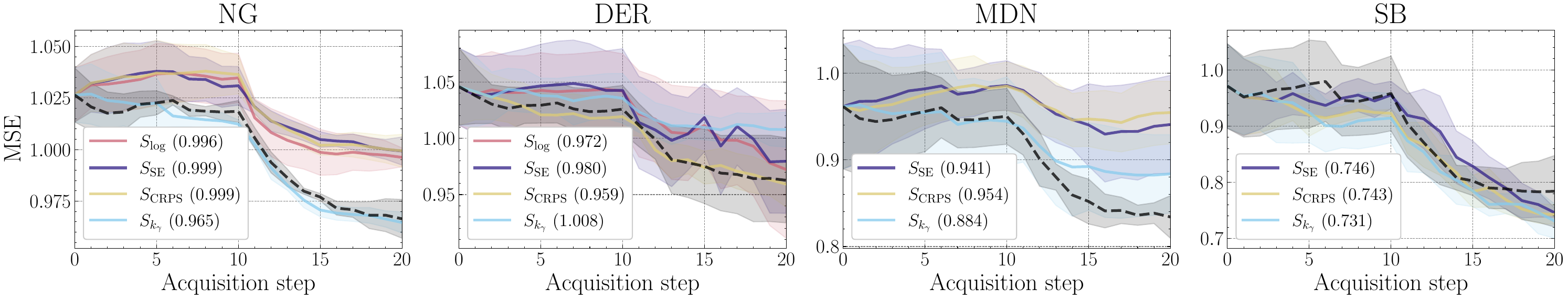}        
    \end{subfigure}
    \begin{subfigure}{\textwidth}
    \caption{Kin8nm}
    \includegraphics[width = \textwidth]{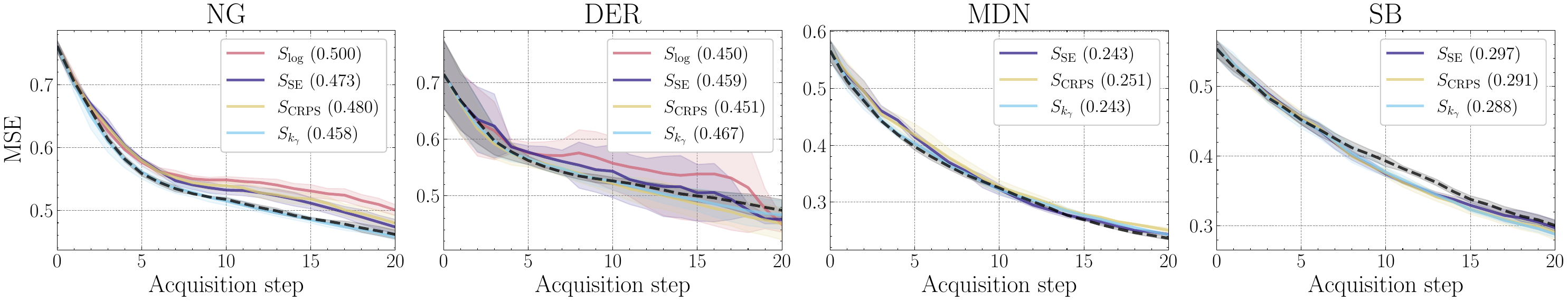}        
    \end{subfigure}
    \begin{subfigure}{\textwidth}
    \caption{Protein}
    \includegraphics[width = \textwidth]{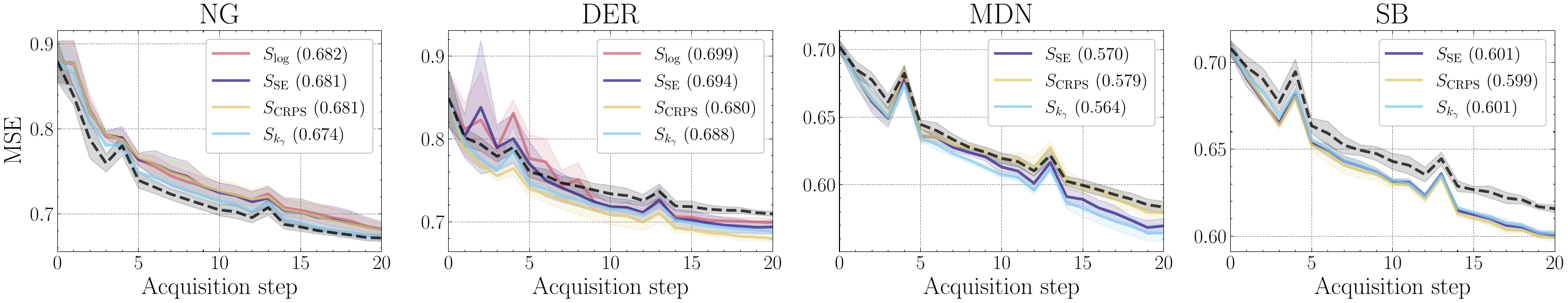}        
    \end{subfigure}
    \caption{Detailed results of the active learning runs for selected UCI datasets. Shown are the mean values and standard deviation (shaded regions) of the test MSE across the three different runs, with the final loss in brackets. The black dotted line is the random baseline.}
    \label{fig:active_learning_uci}
\end{figure}

\begin{figure}[ht]
    \centering
    \includegraphics[width = \textwidth]{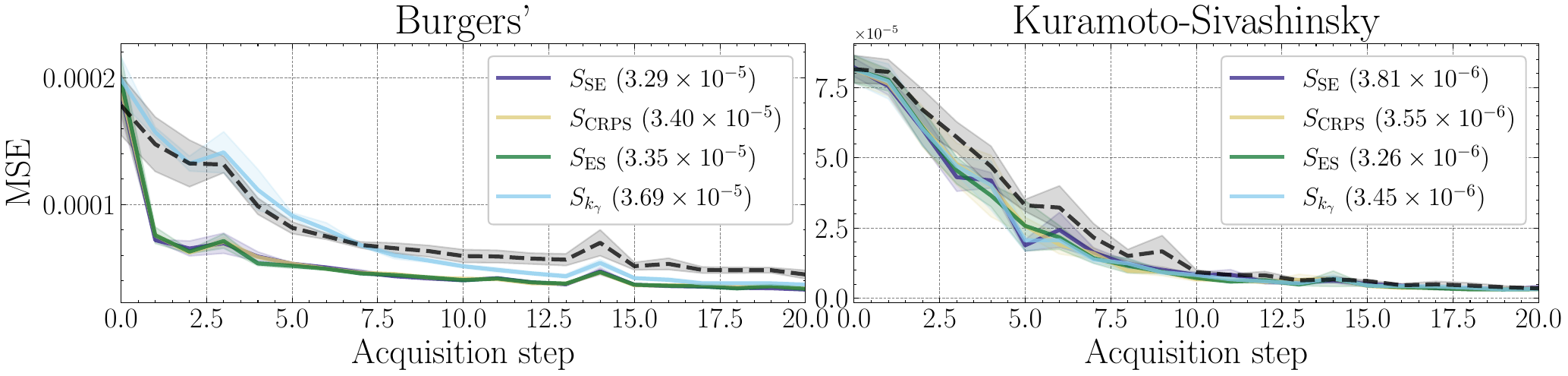}
    \caption{Detailed results of the active learning runs for PDE datasets. Shown are the mean values and standard deviation (shaded regions) of the test MSE across the three different runs, with the final loss in brackets. The black dotted line is the random baseline.}
    \label{fig:active_learning_pde}
\end{figure}

To make the active learning task consistent across the different data modalities, we base the configuration on the total size of the data. In particular, all datasets are split into 90/10 train and test datasets, and the former further into a 90/10 train validation split. We choose an initial fraction of 5\% of the training data size to randomly select the starting pool. The active learning loop is then run for 20 rounds, after which a total of 25\% of the data has been selected, meaning that the amount of new instances in each round is 1\% of the training data size. In each round, $M=5$ ensemble members are trained for a total of 50 epochs from scratch, and the whole process is repeated across three independent seeds. As a fundamental comparison, we also employ a random baseline, which selects the next samples using a uniform distribution.
The full results are listed in \autoref{tab:active_learning_full}. Selected visualizations of the MSE against the acquisition steps can be found in Figure~\ref{fig:active_learning_uci}, and \ref{fig:active_learning_pde}.

\begin{table}[ht]
    \centering
    \caption{Final test loss (MSE $\downarrow$) for the active learning task for all different datasets, uncertainty representation methods, and uncertainty measures. The best measure for each configuration is highlighted in bold, while the standard deviation across the different runs is given in brackets. Note that for the univariate tasks, the $S_\mathrm{ES}$ coincides with $S_\mathrm{CRPS}$ and the values are omitted for readability.}
    \label{tab:active_learning_full}
    \footnotesize
\begin{tabular}{llllllll}
\toprule
 Dataset & Method & $S_\mathrm{log}$ & $S_\mathrm{SE}$ & $S_\mathrm{CRPS}$ & $S_\mathrm{ES}$ & $S_{k_\gamma}$  & $\mathcal{U}$ \\
\midrule
Burgers' & SB & – & \makecell{$\mathbf{3.29 \times 10^{-5}}$\\  (\num{8.05e-07})} & \makecell{\num{3.40e-05} \\ (\num{9.12e-07})} & \makecell{\num{3.35e-05} \\  (\num{4.93e-07})} &\makecell{ \num{3.69e-05} \\ (\num{4.52e-07})} & \makecell{ \num{4.48e-05} \\ (\num{3.36e-06})} \\
\cline{1-8}
KS' & SB & – & \makecell{\num{3.81e-6}\\  (\num{9.25e-07})} & \makecell{\num{3.55e-06} \\ (\num{4.42e-07})} & \makecell{$\mathbf{3.26\times 10^{-6}}$ \\  (\num{3.19e-07})} &\makecell{ \num{3.45e-6} \\ (\num{2.63e-07})} & \makecell{ \num{3.58e-06} \\ (\num{2.62e-07})} \\
\cline{1-8}
\multirow{4}{*}{Concrete}
 & NG & 0.654  (0.041) & 0.657  (0.050) & 0.654  (0.049) & - & \textbf{0.635}  (0.009) & \textbf{0.635}  (0.010) \\
& DER & 0.685  (0.036) & 0.690  (0.018) & \textbf{0.640}  (0.020) & - & 0.653  (0.014) & 0.654  (0.016) \\
 & MDN & – & 0.391  (0.020) & \textbf{0.387}  (0.014) & - & 0.394  (0.018) & 0.402  (0.004) \\
 & SB & – & \textbf{0.346 } (0.025) & 0.360  (0.026) &- & 0.356  (0.023) & 0.373  (0.007) \\
\cline{1-8}
\multirow{4}{*}{Energy}
 & NG & 0.552  (0.034) & 0.586  (0.014) & 0.575  (0.018) & - & 0.522  (0.016) & \textbf{0.521}  (0.009) \\
& DER & 0.524  (0.009) & 0.538  (0.017) & 0.523  (0.039) & - & 0.530  (0.063) & \textbf{0.513}  (0.048) \\
 & MDN & – & 0.175  (0.025) & 0.180  (0.027) & - & \textbf{0.161}  (0.006) & 0.193  (0.002) \\
 & SB & – & 0.142  (0.021) & \textbf{0.138 } (0.018) & - & 0.143  (0.017) & 0.154  (0.001) \\
\cline{1-8}
\multirow{4}{*}{Kin8nm}
 & NG & 0.500  (0.009) & 0.473  (0.012) & 0.480  (0.013) & - & \textbf{0.458}  (0.005) & 0.461  (0.007) \\
& DER & \textbf{0.450 } (0.014) & 0.459  (0.008) & 0.451  (0.028) & - & 0.467  (0.025) & 0.475  (0.019) \\
 & MDN & – & 0.243  (0.002) & 0.251  (0.004) & - & 0.243  (0.002) & \textbf{0.236 } (0.003) \\
 & SB & – & 0.297  (0.010) & 0.291  (0.011) & - & \textbf{0.288 } (0.005) & 0.301  (0.008) \\
\cline{1-8}
\multirow{4}{*}{Naval}
 & NG & 1.143  (0.077) & 1.067  (0.028) & 1.148  (0.088) & - & 0.931  (0.010) & \textbf{0.924 } (0.012) \\
& DER & 1.141  (0.027) & 1.054  (0.182) & 0.918  (0.042) & - & \textbf{0.911} (0.062) & 0.913  (0.048) \\
 & MDN & – & 0.287  (0.049) & 0.309  (0.147) &- & \textbf{0.141} (0.039) & 0.167  (0.038) \\
 & SB & – & \textbf{0.284}  (0.037) & 0.295  (0.050) & - & 0.300  (0.058) & 0.309  (0.033) \\
\cline{1-8}
\multirow{4}{*}{Power}
 & NG & \textbf{0.060}  (0.001) & \textbf{0.060}  (0.001) & \textbf{0.060}  (0.001) & - & \textbf{0.060}  (0.000) & 0.061  (0.000) \\
& DER & 0.058  (0.001) & 0.058  (0.002) & \textbf{0.057}  (0.001) & - & \textbf{0.057}  (0.001) & \textbf{0.057}  (0.001) \\
 & MDN & – & \textbf{0.055}  (0.000) & 0.056  (0.000) & - & \textbf{0.055}  (0.000) & \textbf{0.055}  (0.000) \\
 & SB & – & \textbf{0.056 } (0.001) & \textbf{0.056}  (0.001) & - & \textbf{0.056 } (0.000)  &  0.057  (0.001) \\
\cline{1-8}
\multirow{4}{*}{Protein}
 & NG & 0.682  (0.007) & 0.681  (0.009) & 0.681  (0.007) & - & 0.674  (0.005) & \textbf{0.671}  (0.004) \\
& DER & 0.699  (0.005) & 0.694  (0.008) & \textbf{0.680}  (0.004) & - & 0.688  (0.002) & 0.709  (0.003) \\
 & MDN & – & 0.570  (0.005) & 0.579  (0.006) & - & \textbf{0.564}  (0.006) & 0.583  (0.004) \\
 & SB & – & 0.601  (0.002) & \textbf{0.599}  (0.002) & - & 0.601  (0.003) & 0.616  (0.002) \\
\cline{1-8}
\multirow{4}{*}{Yacht}
 & NG & 0.996  (0.005) & 0.999  (0.007) & 0.999  (0.008) & - & \textbf{0.965 } (0.005) & 0.966  (0.009) \\
& DER & 0.972  (0.061) & 0.980  (0.046) & \textbf{0.959}  (0.020) & \- & 1.008  (0.016) & 0.963  (0.038) \\
 & MDN & – & 0.941  (0.057) & 0.954  (0.036) & - & 0.884  (0.045) & \textbf{0.834}  (0.025) \\
 & SB & – & 0.746  (0.009) & 0.743  (0.018) & - & \textbf{0.731}  (0.031) & 0.784  (0.064) \\
\cline{1-8}
\bottomrule
\end{tabular}
\end{table}

\end{document}